\documentclass[accepted]{uai2025}

\usepackage{etoolbox}
\usepackage{makecell}
\usepackage{threeparttable}

\usepackage{stmaryrd}

\usepackage{hyperref}
\usepackage{tabularx}
\usepackage{comment}
\usepackage{xcolor}
\usepackage{mathrsfs}
\usepackage{floatflt}
\usepackage{thmtools}
\usepackage{thm-restate}
\usepackage{appendix}

\usepackage{chngcntr}
\usepackage{titlesec}

\usepackage{comment}
\usepackage{wrapfig}
\usepackage[utf8]{inputenc} % allow utf-8 input
\usepackage[T1]{fontenc}    % use 8-bit T1 fonts
\usepackage{hyperref}       % hyperlinks
\hypersetup{
    colorlinks=true, % AISTATS keeps links in black.
    linkcolor=blue,
    filecolor=blue,      
    urlcolor=blue,
    citecolor=blue % For biblatex.
    }
\usepackage{url}            % simple URL typesetting
\usepackage{booktabs}       % professional-quality tables
\usepackage{amsfonts}       % blackboard math symbols
\usepackage{nicefrac}       % compact symbols for 1/2, etc.
\usepackage{microtype}      % microtypography
\usepackage{xcolor}         % colors
\usepackage{physics}        % more math commands.
\usepackage{lipsum}         % spacing typesetting.
\usepackage{tikz}
\usetikzlibrary{positioning, shapes.geometric}
\usepackage{amsmath}
\usepackage{amssymb}
\usepackage{mathtools}
\usepackage{amsthm}
\usepackage{mathtools, enumitem}
\usepackage{thm-restate, thmtools}

\newtheorem{theorem}{Theorem}[section]
\newtheorem{property}{Property}
\newtheorem{definition}[theorem]{Definition}

\newtheorem{condition}[theorem]{Condition}

\usepackage{multicol}
\usepackage{graphicx}
\usepackage{subcaption}
\usepackage{appendix}
\usepackage{algorithm}
\usepackage[noend]{algpseudocode}
\usepackage{tikz}
\usetikzlibrary {arrows,patterns,decorations.pathreplacing,chains,positioning,shapes.geometric,calc} 
\usetikzlibrary{arrows}
\usepackage{pifont}% http://ctan.org/pkg/pifont % for checkmarks, etc
\usepackage[T1]{fontenc} %  enables bold small caps
\usepackage{pifont}  % For \ding{}
\usepackage{xcolor}  % For colored text
\usepackage{booktabs} % For \toprule, \midrule, \bottomrule

\definecolor{DarkGreen}{rgb}{0.0, 0.5, 0.0}
\definecolor{DarkRed}{rgb}{0.5, 0.0, 0.0}
\newcommand{\cmark}{\textcolor{DarkGreen}{\ding{51}}}%
\newcommand{\xmark}{\textcolor{DarkRed}{\ding{55}}}%

\AtBeginEnvironment{align}{\setcounter{equation}{0}}
\usepackage{amsmath}
\newcommand{\ind}{\perp\!\!\!\perp} 
\newcommand{\nind}{\not\!\perp\!\!\!\perp}

\usepackage{colortbl} % for colored table columns
\usepackage{array} % for multirows
\usepackage{adjustbox}  % for adjusting width of tables to text width
\usepackage{multirow} % vertical text in table
\usepackage{booktabs} %cmid rule under multicolumn

\usepackage{arydshln} % dashed and dotted hlines
\usepackage{hhline}   % for double cmidrules
\usepackage{setspace} % spacing on table of contents

\newcommand{\sujai}[1]{
{\color{blue} [Sujai: {#1}]}}

\usepackage[american]{babel}
\usepackage{natbib} % has a nice set of citation styles and commands
\title{LoSAM: Local Search in Additive Noise Models with Mixed Mechanisms and General Noise for Global Causal Discovery}

\author[1]{\href{mailto:<sh2583@cornell.edu>?Subject=Your UAI 2025 paper}{Sujai~Hiremath}}
\author[2]{{Promit~Ghosal}}
\author[1]{Kyra~Gan}
\affil[1]{%
    Cornell Tech\\
    New York, NY
}
\affil[2]{%
    University of Chicago\\
    Chicago, Illinois
}
\newcommand{\dtwodag}{\dag\!\dag}

\newcommand{\dddag}{\dag\!\dag\!\dag}

\begin{document}
\maketitle

\begin{abstract}
Inferring causal relationships from observational data is crucial when experiments are costly or infeasible. \emph{Additive noise model}s (ANMs) enable unique \emph{directed acyclic graph} (DAG) identification, but existing sample-efficient ANM methods often rely on restrictive assumptions on the data generating process, limiting their applicability to real-world settings.
We propose \emph{local search in additive noise models}, LoSAM, a topological ordering method for learning a unique DAG in ANMs with mixed causal mechanisms and general noise distributions. 
We introduce new causal substructures and criteria for identifying roots and leaves, enabling efficient top-down learning. We prove asymptotic consistency and polynomial runtime, ensuring scalability and sample efficiency. We test LoSAM on synthetic and real-world data,  demonstrating 
state-of-the-art performance
across all mixed mechanism settings.
\end{abstract}
  % We address the challenge of sample-efficient causal discovery in structural causal models with additive noise without imposing additional assumptions on the underlying data-generating process \promit{First sentence needs to be rephrased - could be something like - We consider the problem of sample-efficient causal discovery in structural causal models with additive noise, without imposing restrictive assumptions on the underlying data-generating process.}. 

% The ability to infer
% methods address this by 
% % aim to infer the underlying structure of the
% % data-generating process (DGP) by 
% restricting the functional form of causal relationships,
% making the structure learning problem well-posed \citep{zhang_identiability_2009}. %prior sentence needs work
% % FCM-based algorithms represent the inferred causal relationships between the
% % measured variables using causal graphical models, i.e., \emph{directed acyclic graphs} (DAGs), and
% This has fundamental applications in science, healthcare and economics \citep{runge_inferring_2019, lee_causal_2022, addo2021exploring}.
% making it
% 
% The additive noise model (ANM) is a popular modeling approach, where the causal parents of a variable are assumed to be statistically independent of its noise term \citep{peters_causal_2014}. 
% 
% 
% due to their sample efficiency and worst-case polynomial runtime .
\section{Introduction}\label{sec: intro}
Inferring causal relationships from observational data is crucial for answering interventional questions \citep{pearl_causal_2009}, especially when experiments are costly or infeasible \citep{causal_ground_faller_2024}. 
\emph{Functional causal model} (FCM)  methods address this challenge by
% ensure a well-posed structure learning problem by 
restricting the functional form of causal relationships, ensuring a well-posed structure learning problem that enables the identification of a unique DAG \citep{zhang_identiability_2009}.
FCM methods
% methods take an order-based approach  which 
usually decompose graph learning into two phases: 
% \citep{teyssier2012ordering}:
1) inferring a causal ordering of the variables (topological ordering), and 2) identifying edges consistent with the ordering (edge pruning). 

Among FCM approaches, those based on the ANM have gained significant traction \citep{hoyer_nonlinear_2008, peters_causal_2014, rolland_score_2022}.
The ANM assumes that each child variable in the DAG is a (potentially nonlinear) function of its parent variables plus an independent noise term. 
This framework is sample-efficient, and worst-case polynomial algorithms exist for structural identification (under additional assumptions) \citep{peters_causal_2014}.
These properties make ANM-based methods
theoretically appealing, and they have been applied in domains such as
% particularly valuable in applications such as
science, healthcare, and economics \citep{runge_inferring_2019, lee_causal_2022, addo2021exploring}.

\begin{table*}[t!]
    \centering
    \begin{threeparttable}
    \renewcommand{\arraystretch}{.8} % Adjust row height
    \setlength{\tabcolsep}{6pt} % Adjust column spacing
    \begin{tabular}{l c c c c}
        \toprule
        \textbf{Algorithm} 
        & \textbf{ANM Type} 
        & \makecell{\textbf{Ordering}\\\textbf{Approach}} 
        & \makecell{\textbf{Uses Leaf}\\\textbf{Properties}} 
        & \makecell{\textbf{Uses Root}\\\textbf{Properties}} \\
        \midrule
        DirectLiNGAM \citep{shimizu2011directlingam}
        & Linear Non-Gaussian ANM & Top-Down & \xmark & \cmark \\
        LISTEN \citep{ghoshal2018learning}
        & Linear ANM\textsuperscript{\dag} & Bottom-Up & \cmark & \xmark \\
        CAM \citep{buhlmann_cam_2014}
        & Nonlinear Additive Gaussian ANM & -- & -- & -- \\
        SCORE \citep{rolland_score_2022}
        & Nonlinear Gaussian ANM & Bottom-Up & \cmark & \xmark \\
        DAS \citep{montagna_scalable_2023}
        & Nonlinear Gaussian ANM & Bottom-Up & \cmark & \xmark \\
        DiffAN \citep{sanchez_diffusion_2023}
        & Nonlinear Gaussian ANM & Bottom-Up & \cmark & \xmark \\
        RESIT \citep{peters_causal_2014}
        & ANM & Bottom-Up & \cmark & \xmark \\
        NoGAM \citep{montagna_causal_2023}
        & Nonlinear ANM & Bottom-Up & \cmark & \xmark \\
        NHTS \citep{suj_2024}
        & Nonlinear ANM & Top-Down & \xmark & \cmark \\
        CaPS \citep{caps_xu2024ordering}
        & Gaussian ANM\textsuperscript{\dtwodag} & Bottom-Up & \cmark & \xmark \\
        Adascore \citep{montagna2025score}
        & ANM & --\textsuperscript{\dddag} & \cmark & -- \\
        \textbf{LoSAM (ours)}
        & ANM & Top-Down & \cmark & \cmark \\
        \bottomrule
    \end{tabular}
    \begin{tablenotes}
        \footnotesize
        \item[\dag] LISTEN requires an additional condition on the inverse of the covariance matrix for identifiability.
        \item[\dtwodag] CaPS requires additional conditions on marginal variances and the score function for identifiability.
        \item[\dddag] Unlike other methods, Adascore directly returns a DAG rather than a topological ordering.
    \end{tablenotes}
    \end{threeparttable}
    \caption{Comparison of FCM methods based on identifiability assumptions, search strategy (top-down or bottom-up), and whether specific substructures (leaf or root nodes) are leveraged.}
    \label{fig:fcm_assum}
\end{table*}

% Despite their general identifiability \citep{peters_causal_2014},
However, their practical utility is often constrained by restrictive assumptions on functional forms and noise distributions
% most prior work on ANMs imposes additional assumptions on functional forms and noise distributions to enable efficient solutions 
(see Table \ref{fig:fcm_assum}).
For example, most existing methods require 
% These methods typically restrict
% in general requiring 
all causal mechanisms to be either fully linear or nonlinear,  
limiting their applicability to real-world settings where mixed mechanisms might present. In biological networks, for instance, single-gene effects are assumed to be additive, while gene-gene interactions (epistasis) introduce nonlinear interactions \citep{kontio2020estimating, manicka2023nonlinearity, hadamard_bio}.

While sample-efficient polynomial-time algorithms have been proposed under these restrictive assumptions for identifying a unique DAG (Table \ref{fig:fcm_assum}), their existence remains unknown for the general mixed-mechanism setting,
where some relationships are linear and others are nonlinear.
Achieving stable performance across diverse ANM settings is particularly challenging, as existing methods often leverage special statistical properties of
causal substructures, like leaves and roots, that hold only under specific functional or distributional constraints. 
Recently, \cite{caps_xu2024ordering}
introduce CaPS, an efficient leaf-based topological ordering method that accommodates mixed causal mechanisms.
However, it assumes Gaussian noise and imposes unverifiable assumptions on noise variances (see Appendix \ref{appendix: caps_assum}), highlighting the need for more flexible solutions.

% developed a topological ordering method (CaPS) capable of accurate discovery in the prescence of linear, nonlinear, or mixed causal mechanisms. 

% Figure \ref{fig: main exp results} illustrates how the performance of baselines degrades when noise terms are non-Gaussian, and the proportion of linear and nonlinear mechanisms changes: performance in equally mixed setting (linear proportion = $0.5$) is particularly poor. This indicates that current FCM approaches may be unsuitable for real-world applications, as in practice we do not know a priori whether the causal relationships are linear or nonlinear, or what constraints the noise distribution may satisfy. Indeed, domain knowledge often indicates that real-world systems involve heterogeneous functional forms; for example, in biological networks, single-gene effects are assumed to be additive, while gene-gene interactions (epistasis) introduce nonlinear interactions \citep{kontio2020estimating, manicka2023nonlinearity, hadamard_bio}. Thus, there remains a need for a sample-efficient FCM method that flexibly handles mixed causal mechanisms and general noise distributions.

 %Adascore fails to outperform prior methods in synthetic and real experiments due to sample complexity issues resulting from high-dimensional regressions. 

\textbf{Contributions.}
In this paper, we propose LoSAM, a novel \emph{topological ordering} method for ANMs that efficiently recovers the unique DAG \emph{without} requiring additional assumptions. Our contributions are threefold:
\begin{itemize}[leftmargin=*, itemsep=0pt, parsep=0pt
% ,before=\vspace{-10pt}, after=\vspace{-8pt}
]
\item \textbf{Root and Leaf Identification:}
We introduce new local causal substructures: \emph{single root} \emph{descendants}, \emph{multi-root descendants}, \emph{v-patterns} (Section \ref{sec: root id}), \emph{valid leaf candidates}, \emph{nonlinear descendants}, and \emph{linear descendants} (Section \ref{sec: sort id}). We characterize their statistical properties in ANMs with mixed mechanisms, without relying on distributional constraints. Leveraging these structures, we establish novel criteria for identifying roots (Lemma \ref{lemma: root id step}) and leaves (Lemma \ref{lemma: ND prune}). Building on these, we introduce LoSAM (Algorithm \ref{algo: LoSAM}), a topological sorting algorithm that learns a DAG in a top-down manner by first identifying the roots, then using them to recursively identify leaves.
% uses the roots to learn DAG from the top-down, leveraging previously ordered nodes to reduce the size of conditioning sets used in subsequent nonparametric regressions.

\item \textbf{Theoretical Guarantees:}  
We prove that LoSAM is asymptotically consistent under identifiable ANMs (Theorem \ref{theorem: LoSAM correctness}), ensuring correct discovery in the limit. Further, we establish the worst-case polynomial runtime of LoSAM (Theorem \ref{theorem: LoSAM runtime}), enabling scaling to larger graphs. We show that LoSAM achieves theoretical gains in sample efficiency when compared to prior methods, in both the root-finding stage (Theorem \ref{prop: rootidcondset}) and overall procedure (Theorem \ref{theorem: losam_time_red}).

\item \textbf{Comprehensive Evaluation:} We extensively evaluate LoSAM on synthetic data, achieving state-of-the-art performance in linear and nonlinear settings, while outperforming baselines in mixed mechanism settings. We validate the real-world applicability of LoSAM on a popular biological dataset (Section \ref{sec: exp results}). The source code for LoSAM is publicly available at \url{https://github.com/Sujai1/local-search-discovery}.
% The source code for LoSAM is included in the supplementary material, and will be made public upon acceptance.
\end{itemize}

% \textbf{Outline.} In Section \ref{sec: related works} we review related work and introduce preliminaries.
% %in Section \ref{sec: prelim}.
% In Section~\ref{sec: observed model} we establish various local causal substructures,
% introducing a \emph{topological sorting} algorithm LoSAM. We test LoSAM in experiments with both real and synthetic data and conclude in Section \ref{sec: exp results}.

\subsection{Related Works}\label{sec: related works}
% \subsection{Related Works}
% dont discuss both in one sentence, sentence too long
% dont say traditional, thats controversial
% make it clear ur talking about global discovery algorithms.
% Causal discovery methods \citep{glymour_review_2019}
Causal discovery methods mostly fall into three categories: constraint-based \citep{peter_spirtes_causation_2000,spirtes_anytime_2001}, scoring-based \citep{chickering_learning_nodate, grasp_liam_2021}, and FCM-based \citep{glymour_review_2019}. 
The first two categories of methods return 
% % Constraint-based methods use conditional independence tests to infer causal edges, while scoring-based methods search for the causal model that best fits the data. Both approaches yield 
an equivalence class of models with the same conditional independence structure \citep{pearl_causal_2009}, rather than a unique DAG \citep{montagna_assumption_2023}.
% Constraint-based methods leverage conditional independence tests to identify causal relationships . Scoring-based methods search over the space of possible causal models, and select the one that maximizes a goodness-of-fit measure. Both approaches return a class of causal models that share the same conditional independence structure \citep{pearl_causal_2009}, rather than a unique DAG \citep{montagna_assumption_2023}. This ambiguity often limits their usefulness for downstream tasks that require a fully specified causal model \citep{shah2024front}.
% in contrast is supposed to be a bridge from one paragraph to next
% redefine MEC, its not been defined yet
% ddont repeat, just say directly what you wat to say (they return an ambiguous causal knowledge, insufficient for downstream tasks
% again, bringing up fcm in a paragraph about traiditonal
Moreover, these methods can exhibit poor sample efficiency \citep{maasch2024local}, and generally suffer exponential worst-case time complexity in the number of variables unless sparsity constraints are imposed \citep{chickering2004large, ganian2024revisiting, colombo2012learning, claassen_learning_2013}. 

%in contrast, what is the advantage of FCM (dont say MEC) - enables better downstream effect.
In contrast,
existing FCM methods
% under specific constraints, FCM methods 
return a unique DAG in polynomial time,
% with worst-case polynomial time complexity,
enabling point estimates for downstream causal effects (rather than bounds, e.g., \citealt{malinsky2016estimating}).
% downstream effect estimation \citep{hoyer_estimation_2008}. 
% just say here the LoSAM, in contrast, has the loosest set of assumptions. However, 
% Our work is most closely related to the recent stream of FCM methods, each of which is designed to handle a different set of parametric assumptions (see Table \ref{fig:fcm_assum}). 
%i.e., when a variable of interest is regressed against its hypothesized parent set, and an independent residual is recovered. 
The topological ordering step in FCM methods mostly falls into two categories: score-matching-based methods and regression-based methods.
As shown in Table \ref{fig:fcm_assum},
FCM methods typically leverage statistical properties of roots or leaves unique to different parametric models to recursively construct a topological ordering.
Score-matching-based methods (LISTEN, SCORE, DAS, DiffAN, NoGAM, CaPS, Adascore\footnote{Adascore leverages score-matching in mixed mechanism ANMs without explicitly constructing an ordering, but has empirical performance similar to NoGAM \citep{montagna2025score}.})
% first identify leaves.
% leverage properties of the score-function to identify leaves; 
% However, 
% They 
typically rely on large conditioning sets or Gaussianity assumptions
to estimate the score-function,
leading to low sample efficiency \citep{suj_2024, montagna_causal_2023}.

Regression-based methods exploit the independence of the noise term to identify causal relationships.
% through regressions and independence tests.
RESIT regresses
% exploits the fact that regressing
leaves onto unsorted vertices, yielding an independent residual in both linear and nonlinear ANMs. However, it relies on high-dimensional nonparametric regression, suffering in sample complexity \citep{peters_causal_2014}. DirectLiNGAM takes a top-down approach but heavily relies on the linearity assumption. 
% observes that
% % exploits the fact that 
% roots are independent of residuals when regressed on, but it relies heavily on linearity. 
Closest to our approach is NHTS, which first identifies roots in nonlinear models,
% via conditional independence conditions, 
and then leverages them to reduce the size of conditioning sets in subsequent steps. 
However, NHTS fundamentally fails at handling linear mechanisms, and can face sample efficiency issues when there are many roots, as illustrated in Appendix \ref{appendix: nonlinear anc-des pair}.
% Mixed causal mechanisms pose a challenge
% because the conditions required to recover an independent residual (and thereby identify a causal relationship) vary based on the underlying functional relationships.
In contrast,
LoSAM extends beyond prior topological ordering algorithms by handling mixed mechanisms, without imposing additional assumptions on the noise distribution. 

CAM does not fall into either category. It
% is a notable exception to the above paradigm, 
leverages MLE in nonlinear additive Gaussian models, rather than causal substructures, to construct a topological ordering. In our experiments, we include CAM as one of our benchmarks.

%  recursively identify roots or leaves of the causal graph. 
% % they condition on unsorted vertices, estimate score function (large conditioning set)
% regression + independence:
% RESIT,
% include ICA-LiNGAM, DirectLiNGAM, 
% % linear only, much lower conditioning set size, project out root, not going to work for nonlinear mechanism, pr
% and NHTS, whereas 
% % closest to our method, but fundamentally fails at handling linear mechanisms (which authors describe).

% Many FCM methods (RESIT, NHTS, DirectLiNGAM, etc.)  Mixed causal mechanisms pose a challenge
% because the conditions required to recover an independent residual (and thereby identify a causal relationship) vary based on the underlying functional relationships. \cite{suj_2024} show that in linear models specific ancestor-descendant pairs yield independent residuals, whereas the same test may yield dependent residuals when the mechanism is nonlinear (see Appendix \ref{appendix: nonlinear anc-des pair} for an example). Regression-based 

Once a topological ordering is obtained from an FCM method, it is straightforward to prune spurious edges via sparse regression (Lasso regression \citep{lasso_tibshirani}) additive hypothesis testing with generalized additive models (CAM-pruning \citep{buhlmann_cam_2014}), or conditional independence tests (Edge Discovery \citep{suj_2024}).

\section{Problem Setup}\label{sec: prelim}
An structural causal model (SCM) is represented by a DAG, denoted as $G=(V,E)$ on $|V|=d$ vertices, where $E$ represents directed edges. An edge $x_i\rightarrow x_j \in E$ 
\emph{iff} (if and only if) $x_i$ has a \emph{direct} causal influence on $x_j$. We define four pairwise relationships between vertices: 1)
\(\text{Ch}(x_i)\) denotes the \textit{children} of $x_i$ such that $x_j \in \text{Ch}(x_i)$ iff $ x_i \to x_j$, 2) \(\text{Pa}(x_i)\) denotes the \textit{parents} of $x_i$ such that $x_j \in \text{Pa}(x_i)$ iff $x_j \to x_i$, 3) \(\text{An}(x_i)\) denotes the \textit{ancestors} of $x_i$ such that \(x_j \in \text{An}(x_i)\) iff there exists a directed path $x_j \dashrightarrow x_i$, 4) \(\text{De}(x_i)\) denotes the \textit{descendants} of $x_i$ such that \(x_j \in \text{De}(x_i)\) iff there exists a directed path $x_i \dashrightarrow x_j$. 
% We partition vertices into 
We consider
four vertex categories based on the totality of their pairwise relationships: 1) $x_i$ is a \textit{root} iff \(\text{Pa}(x_i)=\emptyset\), 2) a \textit{leaf} iff \(\text{Ch}(x_i)=\emptyset\), 3) an \textit{isolated} vertex iff $x_i$ is both a root and a leaf, and 4) an \textit{intermediate} vertex otherwise.

We also classify vertices in terms of their triadic relationships: $x_i$ is a \textit{confounder} of $x_j$ and $x_k$ if $x_i \in \text{An}(x_j)\cap \text{An}(x_k)$, or a \textit{mediator} of $x_j$ to $x_k$ if $x_i \in \text{De}(x_j)\cap \text{An}(x_k)$.

\begin{definition}[Topological Ordering]\label{def: topo_order}
Consider a DAG denoted as $G = (V,E)$. We say that a mapping $\pi: V \rightarrow \{0,1,\ldots, |V|\}$ is a linear topological sort 
% (linear order)
iff $\;\forall x_j \in V$, whenever 
$x_i \in \text{Pa}(x_j)$, $x_i$ appears before $x_j$ in the sort $\pi$, i.e., $\pi(x_i) < \pi(x_j)$.
\end{definition}

\begin{definition}[ANMs,~\citealt{hoyer_bayesian_2009}]\label{def: anm}
Additive noise models are a specific class of SCMs with
\vspace{-5pt}
\begin{equation}\label{eq:ANM}
    x_i = f_i(\text{Pa}(x_i)) + \varepsilon_i
\end{equation}
$\forall x_i \in V$, where $f_i$'s are arbitrary functions and $\varepsilon_i$'s are independent arbitrary noise distributions.
\end{definition}
\textbf{Assumptions.\;}
Definition \ref{def: anm} implies the Causal Markov condition due to the joint
independence of noise terms \(\varepsilon_i\); we further assume that
all variables are observed, acyclicity, faithfulness \citep{spirtes_causal_2016}, as well as the
unique identifiability of the ANM.
% Definition \ref{def: anm} implicitly assumes the Causal Markov condition and acyclicity; we further assume that all variables are observed, faithfulness \citep{spirtes_causal_2016}, as well as the unique identifiability of the ANM. 
Intuitively, the identifiability condition rules out specific combinations of causal mechanisms and noise distributions, such as linear $f_i$ and Gaussian $\varepsilon_i$, but allows for a broad mix of linear and nonlinear functions with general noise distributions (see Appendix \ref{appendix: assum and method} for details).

\textbf{Outline.\;} LoSAM is a 
%two-step topological ordering that first identifies the root vertices, second leveraging 
topological ordering algorithm that follows a two-step procedure: 1) identifying the root vertices (Algorithm~\ref{algo: observed root id}, Section \ref{sec: root id}), and 2) leveraging the roots to recursively identify leaf vertices of the sorted nodes, yielding the rest of the topological sort (Algorithm~\ref{algo: observed sort id}, Section~\ref{sec: sort id}). 
In Section \ref{sec: LoSAM}, we build upon the above subroutines and introduce the full procedure of LoSAM (Algorithm~\ref{algo: LoSAM}). 

\section{Root Finding}\label{sec: root id}
To develop a subroutine 
for identifying
roots 
in ANMs with both linear and nonlinear causal mechanisms, we identify properties unique to roots.
We will establish a novel set of local causal structures where roots behave distinctly from non-roots (Lemmas \ref{lemma: MRDVP},  \ref{lemma: rootnoSRD}, and \ref{lemma: root id step}), regardless of
their
functional relationships.
Starting with the entire vertex set $V$,
we iteratively prune away non-roots 
by leveraging these asymmetric properties, yielding only the roots.

\begin{figure}[t!]
    \centering
    \includegraphics[width=0.15\textwidth]{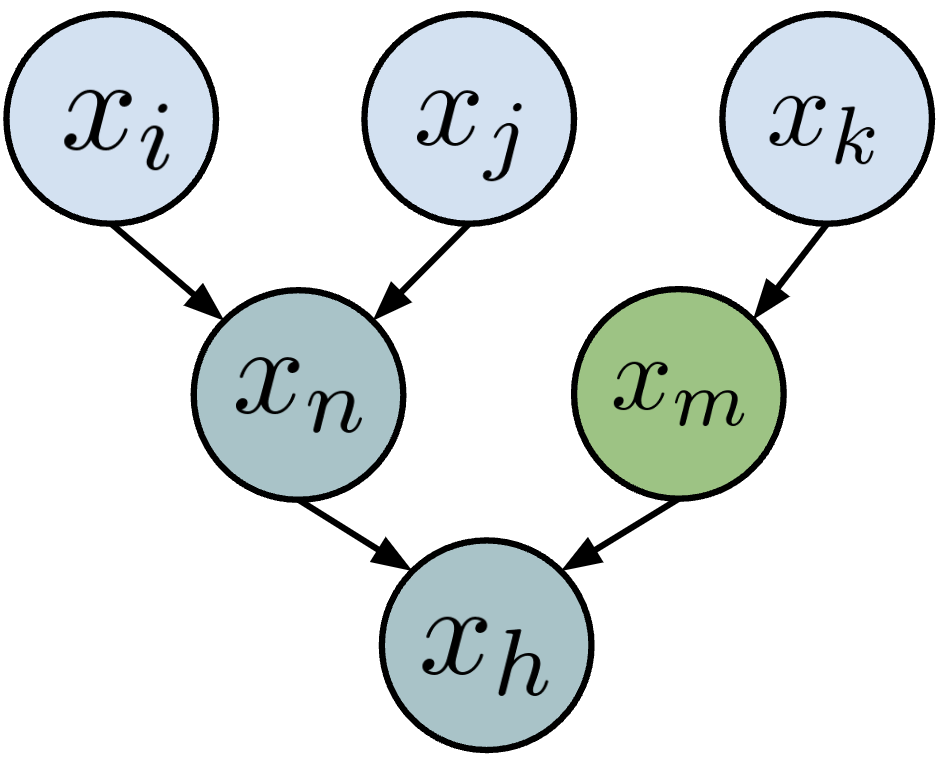}
    \caption{Example DAG where $x_i,x_j,x_k$ are roots, $x_m$ is an SRD, and $x_n, x_h$ are MRDs, inducing VPs between $x_i,x_j$. %that induces a v-pattern between $x_i,x_j$.
    }\label{figure: rootdescendents}
\end{figure}

We first note that non-roots can be partitioned into two distinct categories: those descending from a single root and those descending from multiple roots.
\begin{definition}[Single Root Descendant]\label{def: SRD}
    A vertex $x_i\in V$ is a \emph{single root descendant} (SRD) iff $\exists$ only one root $x_j$, such that  $x_j \in \text{An}(x_i)$.
\end{definition}
\begin{definition}[Multi-Root Descendant]\label{def: MRD}
    A vertex $x_i\in V$ is a \emph{multi-root descendant} (MRD) iff $\exists$ at least two roots $x_j, x_k$ such that $x_j, x_k \in \text{An}(x_i)$.
\end{definition}
Visually, we see this division of non-root vertices into SRDs and MRDs in Figure \ref{figure: rootdescendents}. For any MRD $x_i \in V$, we observe that $x_i$ forms a `v' with any two of its \emph{root ancestors}. We define this ancestor-descendant local substructure as a \emph{v-pattern} (VP),\footnote{Note that an MRD $x_i$ will form additional VPs with \emph{any} pair of independent ancestors.}
and give a characterization based on marginal dependence constraints:
\begin{definition}[VP]\label{def: VP}
        We say that a vertex $x_i$ induces a \emph{v-pattern} iff there exist 
        two vertices $x_j, x_k \in V$ such that $x_i \nind x_j, x_i \nind x_k,$
        $ x_j \ind x_k$.       
\end{definition}
Definition \ref{def: VP} is distinct from but related to the statistical constraints required by the notion of `v-structures' \citep{peter_spirtes_causation_2000}, a triplet of vertices where one vertex is a child of two parents that do not share a direct edge between them (see Appendix \ref{appendix: vpattern compar} for details).

\textbf{Prune MRDs.\;} Our root-finding procedure starts by pruning MRDs, leveraging VPs as described in Lemma \ref{lemma: MRDVP}.
% Next, in Lemma \ref{lemma: MRDVP}, we leverage VPs to distinguish MRDs from SRDs and roots.
\begin{restatable}[MRD Induces VP]{lemma}{MRD}\label{lemma: MRDVP}
        A vertex $x_i$ induces a VP
        % , i.e., $\exists$  $x_j,x_k$ such that $x_i$ induces a VP between $x_j,x_k$,
        iff $x_i$ is an MRD.
\end{restatable}
The proof of Lemma~\ref{lemma: MRDVP} (Appendix \ref{appendix: MRDVP}) relies on the fact that all vertices in $V$ are partitioned into roots, SRDs, and MRDs. We further observe that 1) an MRD induces a VP because roots are all independent of each other, and 2) an SRD or a root does not induce VP.

Lemma~\ref{lemma: MRDVP} implies that by checking whether a vertex 
induces a VP between any two vertices in $V$,
we can identify and prune MRDs from $V$, 
% resulting in a subset
leaving a
$V' \subset V$ 
containing only
% that is
% the union of
SRDs and roots. 
It remains to prune SRDs from $V'$ and identify roots. 

\textbf{Identify Isolated Roots.\;}
Next, 
in Lemma \ref{lemma: rootnoSRD} (proof in Appendix \ref{appendix: rootnoSRD}), we show that 
a root with no SRDs 
can be identified
by testing 
for independence from 
the other variables in
$V'$:
\begin{restatable}[Root with No SRDs]{lemma}{rootnoSRD}\label{lemma: rootnoSRD}
For any $x_i \in V'$, iff $x_i \ind x_j\; \forall x_j \in V'
\setminus \{x_i\}
$, $x_i$ is a root with no SRDs.
\end{restatable}

\textbf{Prune SRDs.\;\;}
Let $V''$ be the 
remaining 
vertices 
after identifying roots with no SRDs using
% that remain after leveraging
Lemma \ref{lemma: rootnoSRD}.  
This set consists of roots with at least one SRD, along with the SRDs themselves. To recover the roots, we follow a two-step procedure: first, we use pairwise nonparametric regression to recover a subset of $V''$ containing all roots, then we prune away the remaining non-roots using bivariate nonparametric regression.

% We follow a two-step procedure to first recover a subset of roots through univariate nonparametric regression, followed by pruning the remaining non-roots using bivariate nonparametric regression.

%To identify the children of the roots, we leverage 

%when a non-root is a child of the root and
% causal mechanisms are nonlinear.
% However, when the 
% % underlying
% mechanism is linear, this procedure can also identify non-roots that are descendants of the root. 
We first formally state the outcome of a nonparametric regression residual test below,
% in Definition~\ref{def: ancestordescendent}, 
considering both linear and nonlinear mechanisms.

\begin{definition}[Regression-Identification Test]\label{def: ancestordescendent}
 Let $x_i,x_j\in V''$,  and let $r_{ij}$ be the residual of $x_j$ nonparametrically regressed on $x_i$. 
 Then, $x_i$ is identified as $\in \text{An}(x_j)$
 % in Ancestor-Descendent (AD) relation to $x_j$ 
 iff: 1) $x_i \ind r_{ij}$, and 2) $x_j \nind r_{ji}$. 
\end{definition}

Definition \ref{def: ancestordescendent} defines the
  relation of "is identified as"
  between \(x_i\) and \(x_j\) as when the residuals from the regressions
  can be leveraged to identify that \(x_i \in An(x_j)\). Note that nonparametric regression yields an independent residual whenever the dependent variable is an additive function of the regressor. Therefore, roots are always identified as ancestors of SRDs that are their children. However, it is possible that some SRDs will pass the Regression-Identification Test; one such subcase occurs when an SRD has a child, and is its only parent. 

The Regression-Identification Test parallels a similar idea in \citealt{suj_2024} (Lemma 4.3), where their local-search approach leverages the result of regression and independence tests to find potential roots. However, their framework exploits local parent-child substructures (PP1, PP2, PP3, PP4) intrinsic to nonlinear models, while the Regression-Identification test captures general ancestor-descendant relationships that may occur in linear, nonlinear, and mixed mechanism models. Additionally, the Regression-Identifiation test is a univariate regression, rather than the multivariate regression employed in Lemma 4.3 of \citealt{suj_2024}.

Using the above Regression-Identification test,
Lemma \ref{lemma: root id step} establishes a subset $W$ of $V''$ that contains all the roots, and further shows that the roots can be distinguished from non-roots within $W$:

\begin{restatable}[Root ID]{lemma}{rootidstep}\label{lemma: root id step}
Let $W$ be a subset of $V''$ such that $\forall x_i \in W$, 1) $\exists x_j \in V$ such that
% $x_i$ passes ADT under $x_j$,
$x_i$ is identified as $\in \text{An}(x_j)$,
and 2) $\nexists x_k \in V$ such that
$x_k$ is identified as $\in \text{An}(x_i)$.
Then $W$ contains all roots in $V''$.

For $x_i,x_j,x_k\in V$, let $r_{ij}^{k}$ be the residual of $x_k$ nonparametrically regressed onto both $x_i$ and $x_j$. Then, vertex $x_i \in W$ is a root vertex if and only if for every other vertex $x_j \in W$ such that $x_i \nind x_j$, $\forall x_k$ such that $x_j$ is identified  $\in \text{An}(x_k)$, we have $x_j \ind r_{ij}^{k}$.
%Further, a vertex $x_i \in W$ is a root iff $\forall x_j \in W\setminus\{x_i\}$ with $x_j \nind x_i$, there exists $x_k \in V''$ such that 1) $x_i$ is identified as $\in \text{An}(x_k)$ and 2) $x_i$ does not d-separate $x_k$ and $x_j$, i.e.,  $x_k \nind x_j | x_i$.

\end{restatable}

% Note that $W$ will contain any SRDs that pass the Regression-Identification Test; one such subcase occurs when an SRD has a child, and is its only parent. To distinguish the roots from these SRDs, the proof of Lemma \ref{lemma: root id step} (Appendix \ref{appendix: root id step}) relies on the intuition that, for any root vertex $x_i$ and SRD $x_j \in \text{De}(x_i)$, there exists a child of $x_i, x_k\in \text{Ch}(x_i)$, such that 
% $x_k$ is a mediator of $x_i$ to $x_j$. This implies that $x_i$ satisfies both conditions of Lemma \ref{lemma: root id step}. Note that any child $x_k$ of a root $x_i$ cannot be included in $W$, as the Regression-Identification Test will always identify $x_i \in \text{An}(x_k)$.
% On the other hand, SRDs in $W$ 
% will
% fail to satisfy 
% % the lemma's 
% the second condition
% in Lemma~\ref{lemma: root id step}: 
% this asymmetry allows SRDs to be pruned.

To distinguish roots from SRDs that pass the Regression-Identification test, the proof of Lemma \ref{lemma: root id step} relies on the intuition that if $x_i \in W$ is a root (a nondescendant of any vertex in $W$), then adding it as a covariate should not change the independence results of any pairwise regression.
However, the reverse does not hold for any non-root SRD: the inclusion of these variables in the bivariate regression would yield at least one dependent residual when the regression involves its ancestor. We formally state this intuition mathematically in Appendix \ref{appendix: root id step}.

\paragraph{Root Finder.\;} Algorithm~\ref{algo: observed root id} outlines our root-finding procedure. 
% 
% We propose a subroutine, Algorithm~\ref{algo: observed root id}, that leverages the above results to identify the set of root vertices in $V$. 
In Stage 1, we run marginal independence tests between all vertices, leveraging Lemmas \ref{lemma: MRDVP} and \ref{lemma: rootnoSRD} to prune MRDs and roots with no SRDs, leaving SRDs and their root ancestors. In Stage 2, we 
% first
run pairwise nonparametric regressions and independence tests
on
% between 
regressors and residuals to obtain the root superset $W$.
% Finally,
We then 
apply
% leverage
Lemma \ref{lemma: root id step}, using conditional independence tests to 
% obtain
identify
the remaining roots.
We show 
the asymptotic correctness of Algorithm~\ref{algo: observed root id} in Proposition \ref{prop: rootidcorr} (proof in Appendix \ref{appendix: rootidcorr}):
\begin{algorithm}[t!]
    \begin{algorithmic}[1]
    \State \textbf{Input}: vertices $x_1,\ldots,x_d\in V$.
    \State \textbf{Initialize}: root set $RT$, root superset $W$.
    \State \textbf{Stage 1: Prune MRDs, Obtain Some Roots}
    \State Run pairwise $\ind$ tests between each pair of vertices $x_i,x_j \in V$ and remove all vertices that induce a VP (MRDs) from $V$ to obtain $V'$.
    \State Add any vertex $x_i \in V'$ to $RT$ if $x_i \ind x_j \forall x_j \in V'\setminus{\{x_i\}}$, and remove all $x_i$ from $V'$ to obtain $V''$.
    %Identify roots with no MRDs via Lemma \ref{lemma: rootnoSRD}, update $RT$, obtain $V''$.
    %\kyra{isolated roots is misleading: this somehow indicates that the roots have no descendants}
    \State \textbf{Stage 2: Prune SRDs, Obtain Leftover Roots}
    \State Run pairwise nonparametric regression between $x_i,x_j \in V''$; if $\exists x_l$ such that $x_k \in V''$ is identified as $\in\text{An}(x_j)$, and there does not exist $x_h$ such that $x_h$ is identified $\in\text{An}(x_k)$, add $x_k$ to $W$.
    %Identify root superset $W$ from $V''$ via pairwise nonparametric regression.
    \State For $x_i \in W$, if $\forall x_j \in W\setminus{\{x_i\}}$ such that $x_j \nind x_i$, $\exists x_k \in V''$ such that $x_i$ is identified as $\in \text{An}(x_k)$ and $x_k\nind x_j|x_i$, add $x_i$ to $RT$.
     \State For each $x_i,x_j \in W$ such that $x_i \nind x_j$, for all $x_k$ such that $x_j$ is identified $\in\text{An}(x_k)$ regress $x_k$ onto $x_i,x_j$ and collect the residual; add any $x_i$ that is always independent of the residual to $RT$.
 
    \State \textbf{return} $RT$.
    \end{algorithmic}
    
    \caption{Root Finder}
    \label{algo: observed root id}
\end{algorithm}
% 
 % 
% Note that high-dimensional nonparametric regressions have been noted as a bottleneck for sample and computational efficiency \citep{peters_causal_2014},
% hindering many FCM-based approaches from identifying the true topological sort from finite sample data despite asymptotic guarantees. 
% However, to our knowledge, our approach is the first to identify root vertices in ANMs without the need for \text{any} multivariate nonparametric regressions. 
% We show the correctness of Algorithm \ref{algo: observed root id} in Proposition \ref{prop: rootidcorr} (proof in Appendix \ref{appendix: rootidcorr}), and provide a formal analysis of the reduction in complexity when compared to NHTS in Theorem \ref{theorem: rootidcondset} (proof in Appendix \ref{appendix: rootidcondset}):
% Next, we show the correctness of our algorithm in Proposition \ref{prop: rootidcorr} (proof in Appendix \ref{appendix: rootidcorr}):
\begin{restatable}{proposition}{rootidcorr}\label{prop: rootidcorr}
    % Given the vertices of a DAG $G$ generated by an ANM, 
    % % (Eq. \ref{eq:ANM}), 
    % Algorithm \ref{algo: observed root id} asymptotically returns the correct set of root vertices. %\kyra{does n sample and asymptotic contradict with each other? statement needs to be revised; also, should you mention near equation 1 that we call these type of ANMs identifiable?} 
    Given the vertices of a DAG $G$ generated by an ANM, infinite data, a consistent nonparametric regression method, and a perfect independence test, Algorithm \ref{algo: observed root id} returns the correct set of root vertices.
\end{restatable}

We note that, under the assumption of linear mechanisms, root-based methods such as DirectLiNGAM are able to identify roots with only univariate regressions; in contrast, root-based nonlinear methods such as NHTS and leaf-based methods such as RESIT and NoGAM require multivariate nonparametric regression with potentially many covariates. Large covariate set size is a major bottleneck for sample and computational efficiency that limits many FCM-based methods from recovering the true sort from finite samples, despite asymptotic guarantees \citep{peters_causal_2014}. To our knowledge, our approach is the first to correctly identify root vertices in ANMs with mixed mechanisms and general noise with a bounded maximum covariate set size.
% using a maximum covariate set size of $2$.
% with a
% % bounded
% maximum size of 2 of the conditioning set.
% covariate set size.
In Theorem \ref{prop: rootidcondset} (proof in Appendix \ref{appendix: rootidcondset}), we formally show the reduction in the size of the conditioning sets used in Alg~\ref{algo: observed root id}: %compared to NHTS:\kyra{update the last sentence}
%out requiring multivariate nonparametric regressions.
% To our knowledge, our approach is the first to identify root vertices in ANMs with nonlinear mechanisms, without requiring \emph{any} multivariate nonparametric regressions. Linear methods such as DirectLiNGAM are able to identify roots with univariate linear regression

% In contrast, high-dimensional regressions are often a bottleneck for sample and computational efficiency \citep{peters_causal_2014}, which limits many nonlinear FCM-based methods from recovering the true topological sort from finite samples, despite asymptotic guarantees. In Proposition \ref{prop: rootidcondset} (proof in Appendix \ref{appendix: rootidcondset}), we formally show the reduction in the size of the conditioning sets used in Alg~\ref{algo: observed root id} compared to the root-based nonlinear method NHTS:
% \kyra{I would consider using $c^{Alg1}_{max}$ instead of rootID}
\begin{restatable}{theorem}{rootidcondset}\label{prop: rootidcondset}
    %Root ID requires no multivariate nonparametric regression, while the root identification step in NHTS requires multivariate regression whenever the graph contains MRDs.
    
    % Consider 
    Given a DAG $G=(V,E)$ under
    % generated by 
    an 
    % identifiable
    ANM,
    % (Eq.~\ref{eq:ANM}), 
    let $M \subseteq V$ and $R \subseteq V$ be the sets of MRDs and roots in $G$, respectively. Let 
    $c^\mathrm{Alg1}_\mathrm{max}$ 
    % $c^\mathrm{Root ID}_\mathrm{max}$ 
 be the max covariate set size
 % used
 in nonparametric regressions 
 % run by
 in
 Algorithm~\ref{algo: observed root id} required to recover roots,
 and similarly
 % Root Identification;
 % we similarly define
 $c^\mathrm{DirectLiNGAM}_\mathrm{max}$, 
 $c^\mathrm{NHTS}_\mathrm{max}$,  $c^\mathrm{NoGAM}_\mathrm{max}$,  $c^\mathrm{RESIT}_\mathrm{max}$.
 %for the root identification step in NHTS, 
 If the ANM is linear, then
 $c^\mathrm{Alg1}_\mathrm{max} = c^\mathrm{DirectLiNGAM}_\mathrm{max} = 1$. If the ANM is nonlinear,  Then,
 $c^\mathrm{Alg1}_\mathrm{max} = 2$,
 $ c^\mathrm{NHTS}_\mathrm{max}$ = $\max_{x_i \in M}(\text{An}(x_i)\cap R, 0)$, and 
 $c^\mathrm{RESIT}_\mathrm{max} = c^\mathrm{NoGAM}_\mathrm{max} = d-2$.
 
 This implies that 
 $c^\mathrm{Alg1}_\mathrm{max} < c^\mathrm{RESIT}_\mathrm{max} = c^\mathrm{NoGAM}_\mathrm{max}$, and when $M \neq \emptyset$, $c^\mathrm{Alg1}_\mathrm{max} 
 \leq c^\mathrm{NHTS}_\mathrm{max} $.
\end{restatable}
 
% We further compare the size of conditioning set sizes to DirectLiNGAM and 
% By leveraging the existence of VPs to prune MRDs, we limit conditioning sets used in nonparametric regressions in Algorithm~\ref{algo: observed root id} to a maximum size of two, 
% improving sample efficiency and achieving
% faster runtime.

Note, the results of Theorem \ref{prop: rootidcondset} hold under the assumptions that all independence tests and regression outcomes are correct. By leveraging the existence of VPs to prune MRDs, we limit the size of covariate sets used in nonparametric regressions in Algorithm \ref{algo: observed root id} to a maximum size of two. Although Algorithm \ref{algo: observed root id} may run potentially more pairwise regressions, avoiding multivariate regressions likely leads to improved sample complexity in many settings, which is confirmed by our experimental results in Section \ref{sec: exp results}.
% compared to
% the corresponding step in NHTS.
% \kyra{update the above}

\section{Sort-Finding}\label{sec: sort id}
% \sujai{We can reframe as leaffind, rename stuff to leaves %and that the quantity we minimize in this section is a property of the leaves...then our algorithm is combining root-finding and leaf finding...think about how to reframe as leaf properties
% }

After obtaining the set of roots, we develop Algorithm~\ref{algo: observed sort id}, which sorts the remaining non-roots into a topological ordering $\pi$. 
We start by adding roots to $\pi$ in any order (as they share no edges),
leaving the unsorted vertices $U = V\setminus{\pi}$.
We follow an iterative procedure by removing one vertex at a time from $U$ and adding it to $\pi$.

\begin{figure}[t!]
    \centering
    \includegraphics[width=0.2\textwidth]{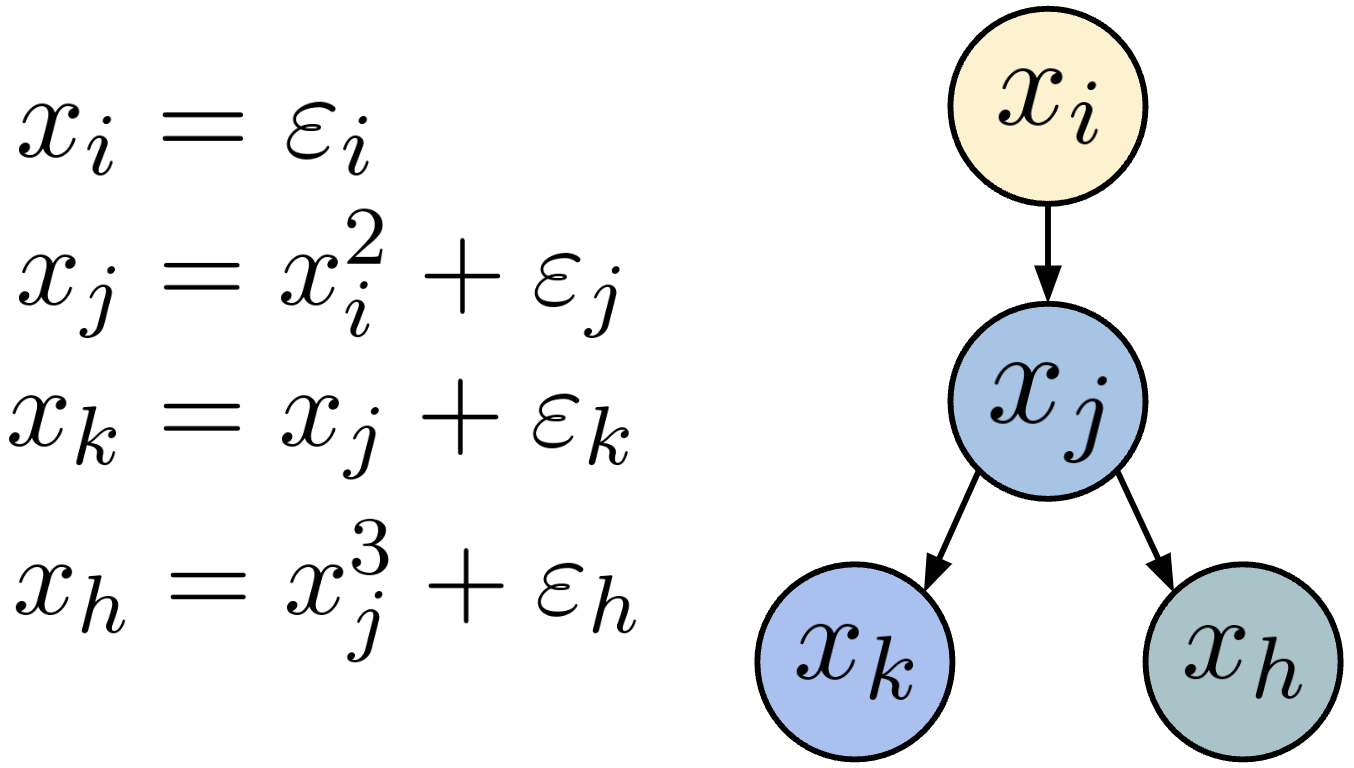}
    \caption{Exemplary DAG, where $x_i$ has been sorted into $\pi$, $x_j$ is a VLC, $x_k$ is a LD, $x_h$ is a ND.}\label{figure: sortpic}
\end{figure}

To ensure the resulting $\pi$ is valid (Definition \ref{def: topo_order}), this vertex must be a leaf of the subgraph induced by the sorted vertices. Therefore, in each iteration, we identify a \emph{valid leaf candidate} (VLC) as illustrated in Figure \ref{figure: sortpic}:
\begin{definition}[VLC]\label{def: VC}
      A vertex $x_i \in U$ is a \emph{valid leaf candidate} iff $\text{Pa}(x_i)\cap U = \emptyset$.
\end{definition}

We proceed by first partitioning non-VLC vertices in $U$ into different categories, relying on
novel local causal substructures (Definitions \ref{def:ND} and \ref{def:LD}).
We then leverage
asymmetric properties to prune non-VLCs from $U$, allowing the identification of a VLC at each iteration.

We first consider a non-VLC $x_i \in U$, such that
there exists a mediator $x_j \in \text{An}(x_i) \cap U$ 
between vertices in $\pi$ to $x_i$, and $x_i$ is a nonlinear function of $x_j$. In Definition~\ref{def:ND}, we characterize such vertices as
% We characterize vertices such as $x_i$ as
\emph{nonlinear descendants} (ND), illustrated in Figure \ref{figure: sortpic}:
\begin{definition}[ND]\label{def:ND}
A vertex $x_i \in U$ is a 
% \emph{nonlinear descendant} 
ND
iff $x_i$ is a nonlinear function of at least one $x_j \in U$.
% $x_j \in U$.
\end{definition}

We observe that
if
any mediator $x_j \in U$ between vertices in $\pi$ and an ND $x_i \in U$
%%%k: here we are already saying that x_i is a nonlinear descendent so we don't need to say again that it is nonlinearly related to x_i%%%
% that is nonlinearly related to $x_i$ 
is not included in the regression of $x_i$ onto vertices in $\pi$,
it will introduce
% ; this introduces 
omitted variable bias \citep{pearl_causal_2016}, 
resulting in
% which results in 
a dependent residual $e_i$. In other words, if $x_i$ is nonparametrically regressed onto the sorted vertices in $\pi$, 
producing a residual $e_i$, then $e_i$ will not be independent of at least one vertex in $\pi$. Accordingly, we identify NDs in Lemma
% We formalize this intuition in Lemma 
\ref{lemma: ND prune} (proof in Appendix \ref{appendix: ND prune}).

\begin{restatable}[ND Test]{lemma}{NDID}\label{lemma: ND prune}
    Vertex $x_i$ is an ND iff $e_i$ is dependent on at least one sorted vertex in $\pi$.
\end{restatable}

% % A direct implication of 
% Lemma~\ref{lemma: ND prune} 
% % is that
% implies that
% % a  
% % In contrast, we show in Lemma \ref{lemma: VC ind} (proof in Appendix \ref{appendix: VC ind lemma}) that VCs always have independent residuals:
% % \begin{restatable}{lemma}{vcind}\label{lemma: VC ind}
% if $x_i \in U$ is a VLC, then $e_i \in U_E$ is independent of all sorted vertices in $\pi$.
% % \end{restatable}

Let \(U_E\) be the set of residuals where \(e_i\in U_E\)
corresponds to the residual from regressing \(x_i\in U\) onto all
vertices in \(\pi\). Lemma \ref{lemma: ND prune} implies that if \(x_i \in U\) is a VLC,
then \(e_i\in U_E\) is independent of all sorted vertices in \(\pi\).

When all causal mechanisms are nonlinear, all non-VLCs in $U$ are NDs.
\begin{comment}

NHTS leverages this asymmetry between VCs and NDs in the sorting procedure
% in each iteration of the sorting procedure
to 
iteratively
recover $\pi$.
\end{comment}
% \emph{however}, this approach relies on the assumption of nonlinear functional causal relationships, i.e., that all non-VCs in $U$ are NDs. 
However, when linear mechanisms are allowed,
% If we allow linear causal mechanisms, then 
there may exist additional vertices $\in U$ that are neither VLCs nor NDs.
These are the \emph{linear descendants} (LD),
% we characterize these problematic vertices as `Linear-Descendants' (
as illustrated in Figure \ref{figure: sortpic}.
\begin{definition}[LD]\label{def:LD}
A vertex $x_i$ is a \emph{linear descendant} iff $x_i$ is a linear function of all ancestors in $U$, and $x_i$ is either a linear or nonlinear function of ancestors in $\pi$.
\end{definition}
We say that "$x_i$ is a linear function of all ancestors in U, and $x_i$ is either a linear or nonlinear function of ancestors in $\pi$" to mean that, if we decompose the ancestors of $x_i$, $An(x_i)$, into two disjoint subsets: 1) the ancestors in $U$ ($x_u \in An(x_i) \cap U$), and 2) the ancestors in $\pi$ ($x_p \in An(x_i) \cap \pi$), then for any representation of $x_i$ as a function of $x_u$ and other ancestors or noise ($x_i = f_i(x_u, An(x_i)\setminus{x_u})+\varepsilon_i$), $f_i$ cannot be nonlinear in $x_u$.
% We show 

Next, in Lemma \ref{lemma: LD ind}, we show that LDs and VLCs share the same independence conditions:
% that LDs satisfy the same independence 
% conditions
% % constraints 
% as VCs:
\begin{restatable}[LD independence]{lemma}{ldind}\label{lemma: LD ind}
    If $x_i \in U$ is an LD, 
    % then 
    $e_i \in U_E$ is independent of all sorted vertices in $\pi$.
\end{restatable}

Lemma~\ref{lemma: LD ind} (proof in Appendix \ref{appendix: LD ind}) explains why prior regression-based methods that leverage roots (DirectLiNGAM, NHTS) fail when both linear and nonlinear mechanisms are present: LDs cannot be distinguished from VLCs via residual independence tests alone. %(see Appendix \ref{appendix: nhts_dlingam compar} for details). 
When both mechanisms are present, to identify VLCs, it remains to prune LDs from $U$, obtaining a subset of vertices $U'$. Then, to improve the stability of the algorithm under a finite sample,\footnote{ If finite sample errors lead to even one false test result when checking the independence of a VLC's residual, the VLC would not be identified.} 
we select a vertex from $U'$ that minimizes a test statistic, rather than checking directly for residual independence.

\textbf{Pruning LDs.\;}
Lemma \ref{lemma: LD} establishes a subset $Q$ of $U$ that distinguishes LDs from VLCs, leveraging nonparametric regression:

%We define the following subset of $U$, leveraging nonparametric regression:
%\kyra{intuition is needed here before I can edit definition 3.10, and give it a name; maybe considering combining definition 3.10 and lemma 3.6: not sure why we need to keep them separate}
%\sujai{intuition added}
\begin{comment}
\begin{definition}\label{def: Q LD}
      For $e_i,e_j\in U_E$, let $q_{ij}$ be the residual of $e_j$ linearly regressed onto $e_i$, and $q_{ji}$ be the residual of $e_i$ linearly regressed onto $e_j$. Let $Q$ be the set of all $x_i \in U$ such that there exists $x_j \in U$ such that the following conditions hold: 1) $e_j \ind q_{ji}$, 2) $e_i \nind q_{ij}$. 
\end{definition}
\end{comment}

%We show in Lemma \ref{lemma: LD} (proof in \ref{appendix: LD lemma}) that LDs can be pruned from $U$ by leveraging $Q$:

\begin{restatable}{lemma}{LDID}\label{lemma: LD}
 For $e_i,e_j\in U_E$, let $q_{ij}$ be the residual of $e_j$ linearly regressed onto $e_i$, and $q_{ji}$ be the residual of $e_i$ linearly regressed onto $e_j$. Let $Q$ be the set of all $x_i \in U$ such that there exists $x_j \in U$ such that the following conditions hold: 1) $e_j \ind q_{ji}$, 2) $e_i \nind q_{ij}$. 
 
Then $Q \subseteq U$ contains all LDs $\in U$ and
% contains
no VLCs.
\end{restatable}

To distinguish LDs from NDs and VLCs, the proof of Lemma \ref{lemma: LD} relies on the intuition that, for any LD $x_i \in U$, there exists a VLC $x_j$ such that $x_j$ is an ancestor of $x_i$, and $x_i$ is a linear function of $x_j$. This allows us to decompose the residual of $x_i$, $e_i$, into a linear function of the residual of $x_j$, $e_j$, yielding an independent residual when $e_i$ is linearly regressed onto $e_j$, but a dependent residual in the reverse direction. This intuition is formalized mathematically in Appendix \ref{appendix: LD lemma}.
%To distinguish LDs from NDs and VCs, the proof of Lemma \ref{lemma: LD} (proof in Appendix \ref{appendix: LD lemma}) relies on the intuition that, for any LD $x_j \in U$, $\exists x_i \in U \cap \text{An}(x_j)$ such that $x_j$ is a linear function of $x_i$, and $x_i$ is a VC. Note, as $x_i$ is a VC we have $\text{Pa}(x_j)\subseteq\pi, \text{De}(x_j)\cap \pi = \emptyset$; this implies the residual $e_j $ equals the error term $\varepsilon_j$. Now, as $x_j$ is LD, it can be decomposed into a potentially nonlinear function of vertices in $\pi$ plus the sum of linear functions of error terms of vertices in $U$; therefore, the residual $e_j$ is simply the sum of those error terms, one of which is $\varepsilon_i$. This implies that the linear regression of $e_j$ onto $e_i$ yields a residual independent of $e_i$. In contrast, as $e_j$ is a function of $e_i$, when $e_i$ is linearly regressed onto the $e_j$, the resulting residual is dependent on $e_j$.
%\sujai{add intuition?}
% By removing $Q$ from $U$, 
% %%% k: let's avoid using the phrase data-driven, too vague. preferred phrase would something like testable or just say which test we use to eliminate what %%%
% we have a data-driven way to prune LDs from $U$, leaving us with the subset $U'$ containing only VC and ND vertices.
Lemma~\ref{lemma: LD} enables us to prune LDs from $U$, leaving the subset $U'$ containing only VLCs and NDs.

%\kyra{need to go through this section again to add paragraph titles to make the section more structured. This section is a bit all over the place now -> we talk about LD, then proposing another test statistic to increase the stability of the algorithm, they are almost orthogonal to each other}

\textbf{Improved Stability for ND Pruning.\;}
To determine whether a residual $e_i$ from regression is independent of its regressors, prior work \citep{peters_causal_2014,suj_2024} requires the independence of each residual from \emph{each} regressor. Accordingly, to prune NDs from $U'$, we might rely on testing the independence of each residual $e_i \in U_E$
to \emph{all} sorted vertices. 
While this procedure is asymptotically unbiased, 
in practice it requires
selecting a cutoff for the independence test given the finite samples.
If not done carefully, this can lead to erroneous VLC identification.\footnote{Bootstrap procedures may be used to compute an optimal cutoff for any given set of samples, but become computationally expensive when $n$ is large.} 
 
To improve the numerical stability of our procedure, we propose
the following test statistic, which captures the dependence between the residual $e_i$ 
and all sorted vertices in $\pi$ in a \emph{continuous} manner:
\begin{equation}
    t^*(e_i,\pi) = \sum_{x_j \in \pi}\widehat{MI}(x_j, e_i),
\end{equation}
where $\widehat{MI}(\cdot,\cdot)$ is a nonparametric estimator of the mutual information \citep{kraskov_estimating_2004}.
Instead of selecting a single cutoff to decide independence between the residual $e_i$ and all vertices in $\pi$, at each iteration, we choose the vertex in $U'$ with the lowest test statistic, $x_i = \arg\min_{x_j \in U'} t^*(e_j, \pi)$, to be a VLC, making our procedure more robust under finite samples.

Lemma \ref{lemma: test stat} (proof in Appendix \ref{appendix: test stat}) shows that the test statistics
$t^*(e_i,\pi)$ equals zero asymptotically only for
$x_i$ that are VLCs:

\begin{restatable}[Consistency]{lemma}{teststat}\label{lemma: test stat} Given a consistent estimator $\widehat{MI}(\cdot,\cdot)$, a fixed $x_i \in U'$ and corresponding residual $e_i$,
% the test statisticcan
$t^*(e_i,\pi)$ asymptotically approaches 0 as $n\rightarrow \infty$ iff $e_i$ is independent of all vertices in $\pi$, i.e., $x_i$ is a VLC.
\end{restatable}

\paragraph{Sort Finder.\;}
Leveraging Lemma~\ref{lemma: test stat}, 
we propose Algorithm~\ref{algo: observed sort id}, Sort Finder, 
which 
iteratively builds the topological sort $\pi$ in a two-stage procedure:
Stage 1
runs nonparametric regressions of each vertex in $U$ onto all vertices in $\pi$ to obtain the residual set $U_E$; we prune all LDs in $U$ using
Lemma \ref{lemma: LD},
obtaining the subset $U'$. 
In Stage 2, we identify a VLC by finding the vertex $x_i\in U'$ that minimizes the test statistic $t^*(e_i,\pi)$, according to Lemma \ref{lemma: test stat}. We repeat this procedure until
all vertices are sorted.
% the unsorted set $U$ is empty.
%%%K: we are writing a paragraph, not restating the code %%%
We provide asymptotic correctness
% show the correctness 
of Algorithm \ref{algo: observed sort id} in Proposition \ref{prop: sortidcorr} (proof in Appendix \ref{appendix: sortidcorr}).

\begin{restatable}[Correctness of Sort Finder]{proposition}{sortidcorr}\label{prop: sortidcorr}
     Given the vertices of a DAG $G$ generated by an ANM, the roots in $G$, infinite data, a consistent nonparametric regression method, and a perfect independence test, Algorithm \ref{algo: observed sort id} returns a valid sort $\pi$.

    % Given a DAG $G=(V,E)$ under
    % % the vertices of a graph $G$ generated by
    % an ANM
    % % (Eq. \ref{eq:ANM})
    % and the roots in $G$, Algorithm \ref{algo: observed sort id} asymptotically returns a valid sort $\pi$.
    
    % Given the vertices of a graph $G$ generated by an ANM (Eq. \ref{eq:ANM}) and the roots in $G$, Algorithm \ref{algo: observed sort id} asymptotically returns a valid sort $\pi$.
\end{restatable}

\begin{algorithm}[t!]
    \begin{algorithmic}[1]
    \State \textbf{Input}: vertices $x_1,\ldots,x_d\in V$, roots $RT$
    \State \textbf{Initialize}: add roots in $RT$ to $\pi$,
    unsorted vertices $U = V\setminus{\pi}$.
    \While{$U\neq \emptyset$}:
    \State \textbf{Stage 1: Prune U}
    \State Run a nonparametric regression of each vertex \hspace*{1.7em}in $U$ onto all vertices in $\pi$, and collect the \hspace*{1.7em}resulting residuals in the set $U_E$.
    % Obtain $U_E$ via nonparametric regression
    \State Run pairwise linear regression between each pair \hspace*{1.7em}of residuals $e_i,e_j\in U_E$ and collect residuals \hspace*{1.7em}$q_{ij},q_{ji}$; remove all $x_i \in U$ such that $e_j \ind \hspace*{1.7em}q_{ji}, e_i \nind q_{ij}$ from $U$ to obtain $U'$.
    %Prune $U$ via Lemma \ref{lemma: LD} to obtain $U'$.
    \State \textbf{Stage 2: Identify VLC}
    \State Identify $x_* =\arg\min_{x_j \in U'} t^*(e_j, \pi)$ as a VLC, 
    %via Lemmas \ref{lemma: test stat}.
    \hspace*{1.7em}add $x_*$ to $\pi$, remove $x_*$ from $U$.
    \EndWhile
    \State \textbf{return} $\pi$. 
    \end{algorithmic}
    
    \caption{Sort Finder}
    \label{algo: observed sort id}
\end{algorithm}

\begin{comment}
We show the correctness of Algorithm \ref{algo: observed sort id} in \ref{theorem: observed sort id correctness} (proof in Appendix \ref{}), and the worst case time complexity in Theorem \ref{theorem: observed sort id time} (proof in Appendix \ref{}). We provide a walk-through of Algorithm \ref{algo: observed sort id} in Appendix \ref{}.
\begin{theorem}\label{theorem: observed sort id correctness}
Given a graph $G$ and a set of roots $R$, Algorithm \ref{algo: observed sort id } asymptotically finds a correct topological sort of $G$.
\end{theorem}
\begin{theorem}\label{theorem: observed root id time}
Given $n$ samples of $d$ vertices generated by a identifiable nonlinear ANM, the worst case runtime complexity of Algorithm \ref{algo: observed root id} is upper bounded by \sujai{fix this time complexity} $O(d^3n^3)$.
\end{theorem}
\end{comment}
\subsection{Theoretical Guarantees}\label{sec: LoSAM}
By combining Algorithms \ref{algo: observed root id} and \ref{algo: observed sort id}, we describe our overall topological sort algorithm, \emph{local search in additive noise models}, LoSAM, in Algorithm~\ref{algo: LoSAM}. LoSAM extends prior top-down regression-based FCM methods
to ANMs with both linear and nonlinear 
mechanisms. 
It reduces the maximum
size of conditioning sets in the root identification phase to two and improves
algorithm stability under finite samples
by selecting the vertex with the smallest test statistic at each iteration of the sorting procedure.

\begin{algorithm}[t!]
    \begin{algorithmic}[1]
    \State \textbf{Input}: vertices $x_1,\ldots,x_d\in V$
    \State Run \emph{Root Finder} (Alg \ref{algo: observed root id}), obtain $RT$.
    \State Run \emph{Sort Finder} (Alg \ref{algo: observed sort id}) using $RT$, obtain sort $\pi$.
    \State \textbf{return} $\pi$. 
    \end{algorithmic}
    
    \caption{LoSAM}
    \label{algo: LoSAM}
\end{algorithm}

% \begin{figure}[t!]%{0.8\textwidth}
% \centering
%      \begin{subfigure}[t]{\columnwidth}  % 4th row, 1st column
%         \includegraphics[width=\columnwidth]{figures/unifnew2.png}
%     \end{subfigure}
    
%      \begin{subfigure}[t]{\columnwidth}  % 4th row, 1st column
%         \includegraphics[width=\columnwidth]{figures/laplanew2.png}
%     \end{subfigure}
%     \caption{LoSAM performance on sparse graphs by linear mechanism proportion. Top row: uniform noise. Bottom row: Laplace noise.} 
%     \label{fig: sparse results brokendown}
% \end{figure}

We provide the correctness of Algorithm \ref{algo: LoSAM} in Theorem \ref{theorem: LoSAM correctness} (proof in Appendix \ref{appendix: LoSAM correctness}), and a step-by-step walk-through of the method in Appendix \ref{appendix: LoSAM walkthrough}.

%We provide a detailed example on
% For clarity, we provide a walk-through of 
%Algorithm \ref{algo: LoSAM} in Appendix \ref{appendix: LoSAM walkthrough}.
\begin{restatable}[LoSAM Correctness]{theorem}{rootidalgo}\label{theorem: LoSAM correctness} 
 Given the vertices of a DAG $G$ generated by an ANM, infinite data, a consistent nonparametric regression method, and a perfect independence test, Algorithm \ref{algo: LoSAM} returns a valid topological sort $\pi$.
% Given a DAG $G$ under
% % the vertices of a graph $G$ generated by 
% an ANM constrained by Definition \ref{def: anm}, Algorithm \ref{algo: LoSAM} asymptotically returns a valid topological sort $\pi$.
\end{restatable}

Next, we establish the worst-case time complexity in Theorem \ref{theorem: LoSAM runtime} (proof in Appendix \ref{appendix: LoSAM correctness}),

\begin{restatable}[LoSAM Runtime]{theorem}{rootidruntime}\label{theorem: LoSAM runtime}
    Given $n$ samples generated from a $d$-dimensional DAG $G$ under an ANM, 
    the worst-case time complexity of Algorithm \ref{algo: LoSAM} is $O(d^3n^2)$.
    % $n$ samples of $d$ vertices generated by an ANM (Eq. \ref{eq:ANM}), the worst-case time complexity of Algorithm \ref{algo: LoSAM} is upper bounded by $O(d^3n^3)$.
\end{restatable}
% Improved runtime:
%  Handling nonlinear mechanisms and arbritrary noise requires nonparametric regression techniques - ordering methods that handle this include NoGAM, RESIT, NHTS. It has equal complexity to say RESIT, but is faster than nhts and nogam, due to reduced complexity of the nonparametric regression method we pick
 The worst case runtime of LoSAM in dimensionality $O(d^3)$ is slightly higher than the $O(d^2)$ complexity of methods that learn the ordering from the leaves to the roots (bottom-up) and do not require Gaussianity, such as RESIT  and NoGAM. However, it matches the complexity of NHTS ($O(d^3)$), a top-down algorithm that also learns the ordering from the roots to the leaves. Correspondingly, LoSAM enjoys improvements in sample efficiency common to root-based methods.
 
\textbf{Efficiency.\;} The number of multivariate nonparametric regressions run in each step of LoSAM is actually inversely related to the size of the covariate sets (similar to the root-based method NHTS), while the number of regressions in each step of leaf-based methods such as RESIT and NoGAM are directly proportionate to the covariate set size. %make more clear
Therefore, LoSAM preserves the reduction in covariate set size inherent to root-based methods over leaf-based methods, which leads to better sample efficiency when estimating nonlinear relationships. We provide a formal analysis of the reduction in complexity for the worst case (i.e., in a fully connected DAG) in Theorem \ref{theorem: losam_time_red} (proof in Appendix \ref{appendix: LoSAM time red}):

 \begin{restatable}[LoSAM Efficiency]{theorem}{losam}\label{theorem: losam_time_red}
     Consider a fully connected DAG $G=(V,E)$
with ANM. Let $d:= |V|$. 
Let $n^\mathrm{LoSAM}_k$ be the number of multivariate nonparametric regressions with covariate set size $k\in[d-2]$ run by LoSAM when sorting $V$; we similarly define $n^\mathrm{NHTS}_k, n^\mathrm{RESIT}_k$ and $n^\mathrm{NoGAM}_k$ respectively.  Then, $n^\mathrm{LoSAM}_k =n^\mathrm{NHTS}_k = d - k$, and $ n^\mathrm{RESIT}_k = n^\mathrm{NoGAM}_k = k+1$. This implies that for all  $k > \frac{d}{2}$, $n^\mathrm{LoSAM}_k = n^\mathrm{NHTS}_k <n^\mathrm{RESIT}_k = n^\mathrm{NoGAM}_k$.
 \end{restatable}

% \begin{figure}[t!]%{0.8\textwidth}
% \centering
%      \begin{subfigure}[t]{\columnwidth}  % 4th row, 1st column
%         \includegraphics[width=\columnwidth]{figures/newunif1.png}
%     \end{subfigure}
%      \begin{subfigure}[t]{\columnwidth}  % 4th row, 1st column
%         \includegraphics[width=\columnwidth]{figures/newpla2.png}
%     \end{subfigure}
%     \caption{Performance of LoSAM on ER1 synthetic data with $50\%$ proportion of linear mechanisms. Top row: uniform noise. Bottom row: Laplace noise.
%     }
%     \label{fig: sparse results overall}
% \end{figure}
\begin{figure}[t!]%{0.8\textwidth}
\centering
     \begin{subfigure}[t]{\columnwidth}  % 4th row, 1st column
        \includegraphics[width=\columnwidth]{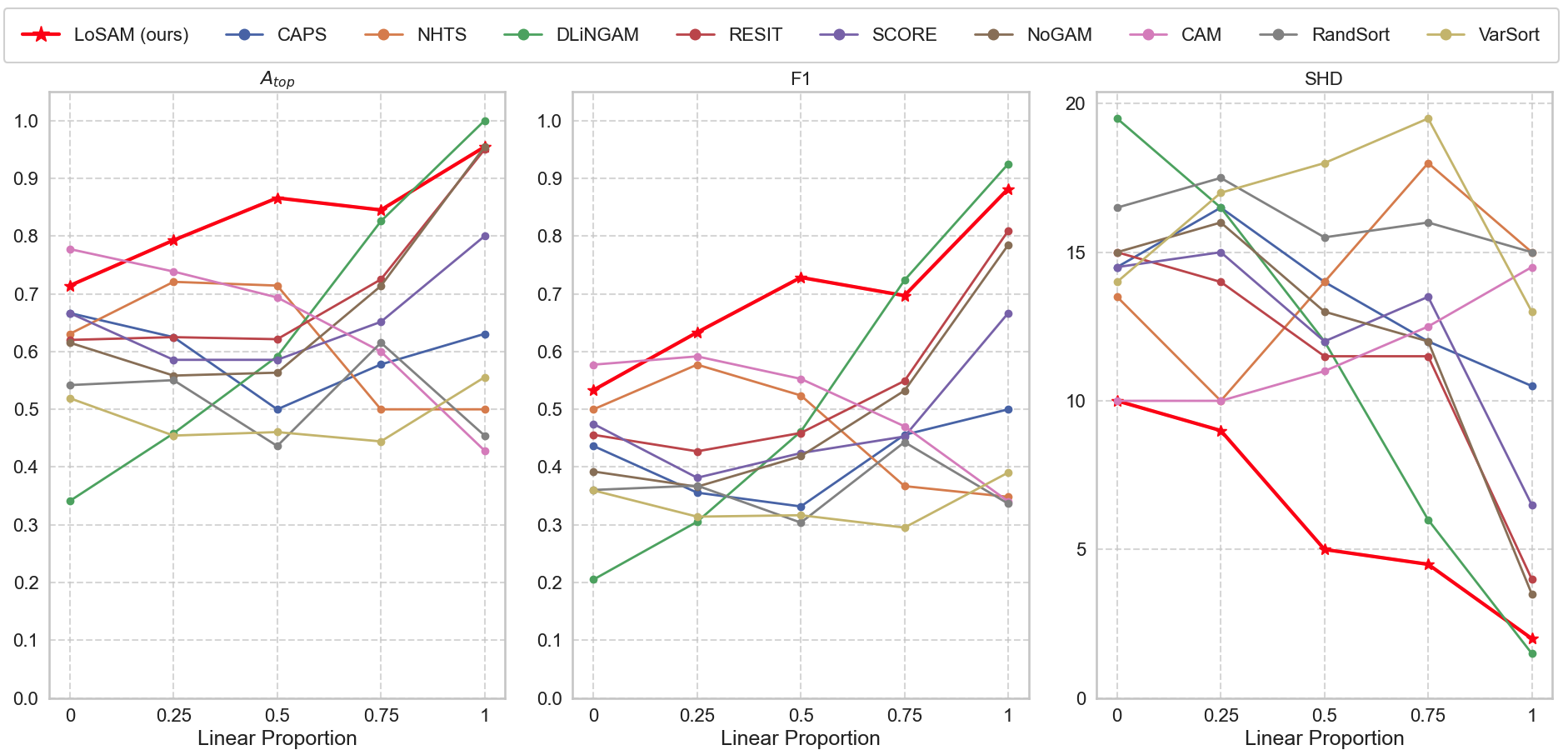}
    \end{subfigure}
    
     \begin{subfigure}[t]{\columnwidth}  % 4th row, 1st column
        \includegraphics[width=\columnwidth]{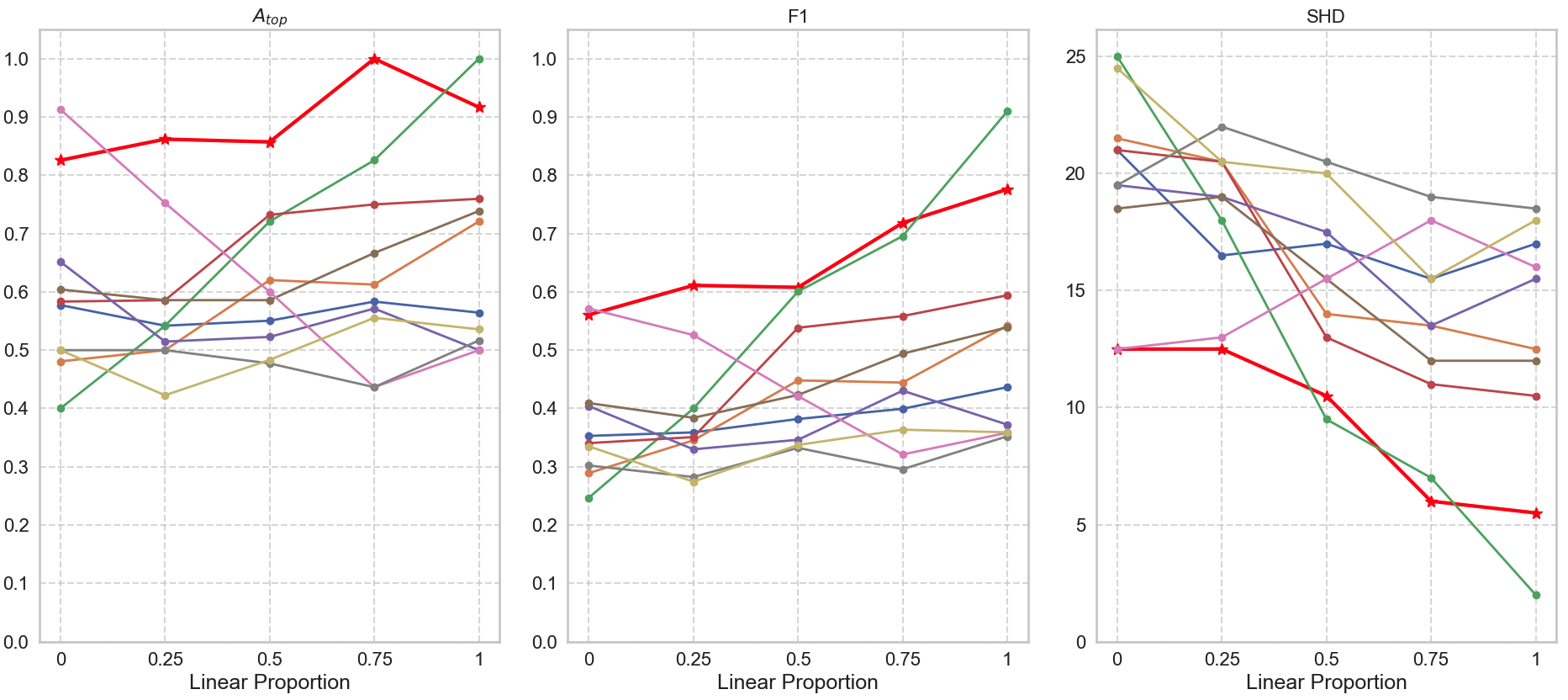}
    \end{subfigure}
    \caption{LoSAM performance on sparse graphs by linear mechanism proportion. Top row: uniform noise. Bottom row: Laplace noise.} 
    \label{fig: sparse results overall}
\end{figure}

%%%%%%%%%
%%%%%%%%%
\section{Experimental Results}\label{sec: exp results}
We evaluate LoSAM on synthetic datasets with mixed, purely linear, and purely nonlinear mechanisms, as well as a real-world protein expression dataset.\footnote{\url{https://github.com/Sujai1/local-search-discovery}.} LoSAM achieves state-of-the-art performance, outperforming existing algorithms in mixed mechanism settings.
% We numerically examine the performance of LoSAM across
% % provide empirical results for LoSAM in both real data and across 
%  a wide range of
% synthetic dataset with mixed, purely linear, and purely nonlinear mechanisms, and a real-world dataset on protein expression.  
% LoSAM achieves SOTA performance, outperforming existing algorithms  
% in mixed mechanisms settings.

% in the biological dataset considered, as well as in synthetic data generated by purely linear or nonlinear mechanisms. Notably, LoSAM particularly overperforms in mixed mechanisms settings, where baseline results are especially poor.

% \subsection{Setup}
\textbf{Synthetic Dataset Generation.\;}
%his dataset captures the expression levels of various proteins in human cells, with a ground-truth causal network established by genetic experiments
% To understand the performance of LoSAM across a wide selection of potential DGPs, 
We produce synthetic data with varying graph sparsity, exogenous error distribution, dimensionality, and causal mechanisms. DAGs are randomly generated with the Erdos-Renyi model \citep{erdos_renyi}; the average number of edges in each $d$-dimensional DAG is either $d$ (ER1) for sparse DAGs, or $2d$ (ER2) for dense DAGs. Uniform, Laplace or Gaussian noise is used as the exogenous error. We randomly sample the data ($d=10, n = 1000$) according to causal mechanisms with a different average proportion of linear mechanisms ($0\%, 25\%, 50\%, 75\%, 100\%$). We standardize and process the data to ensure that the simulated data is sufficiently challenging; methods are evaluated on 30 randomly generated seeds in each experimental setting. Further details
% about the synthetic data generation procedure 
can be found in Appendix \ref{appendix: exp}. 
% In the main text we focus on results on graphs with $10$ variables ($d=10$) and sparse causal mechanisms (ER1), and defer figures corresponding to additional settings to the appendix.  

% To understand the performance of LoSAM across a wide selection of potential DGPs, we produce synthetic data with varying graph sparsity, exogenous error distribution, and causal mechanisms. DAGs are randomly generated with the Erdos-Renyi model \citep{erdos_renyi}; 
% in the main text we focus on results for sparse graphs, where the average number of edges in each $d$-dimensional DAG is $d$ (ER1). Uniform or Laplace noise is used as the exogenous error. We randomly sample the data ($d=10, n = 1000$) according to causal mechanisms with a different average proportion of linear mechanisms (0\%, 25\%, 50\%, 75\%, 100\%). We standardize and further process the data to ensure that the simulated data is sufficiently challenging; further details about the synthetic data generation procedure can be found in Appendix \ref{appendix: data generation}.
% Methods are evaluated on 30 randomly generated seeds in each experimental setting. 

%the average number of edges in each $d$-dimensional DAG is either $d$ (ER1) for sparse DAGs, or $2d$ (ER2) for dense DAGs. In the main text we comment on sparse DAGs, where Uniform or Laplace noise is used as the exogenous error.

\textbf{Real-World Data.\;}
To confirm the real-world applicability of our approach, we test LoSAM on the Sachs dataset \citep{sachs2005causal}, a widely used real-world biological benchmark for causal discovery 
(see Appendix \ref{appendix: data processing} for details). The Sachs data captures the expression levels of various proteins in human cells, with a ground-truth causal network established by genetic experiments. Notably, prior work \citep{caps_xu2024ordering} has estimated that the causal relationships in the Sach's network have mixed mechanisms, making it a promising test case for assessing LoSAM.

\textbf{Baselines.}
We benchmark LoSAM against a mix of classical and SOTA topological ordering baselines: DirectLiNGAM, CAM, RESIT, SCORE, NoGAM, NHTS, and CaPS. We include the heuristic algorithm Var-Sort \citep{reisach_beware_2021} and a randomly generated sort (Rand-Sort) to measure gameability, as some FCM methods are prone to exploiting artifacts common to simulated ANMs \citep{reisach_beware_2021}.
% with citations included
%We benchmark LoSAM against a mix of classical and state-of-the-art topological ordering baselines: DirectLiNGAM \citep{shimizu2011directlingam},  CAM \citep{buhlmann_cam_2014}, RESIT \citep{peters_causal_2014}, SCORE \citep{rolland_score_2022}, NoGAM \citep{montagna_causal_2023}, NHTS \citep{suj_2024}, and CaPS \citep{caps_xu2024ordering}. We include the heuristic algorithm Var-Sort \citep{reisach_beware_2021} and a randomly-generated sort (Rand-Sort) to measure gameability, as some FCM methods are prone to exploiting artifacts common to simulated ANMs \citep{reisach_beware_2021}.

%$R^2$-Sort \citep{reisach_scale-invariant_2023} 

\textbf{Evaluation and Metrics.\;}
We first directly evaluate the topological orderings via the average topological divergence $A_{top}$ (higher $A_{top}$ is better), which is equal to the percentage of edges that can be recovered by the returned topological ordering (an edge cannot be recovered if a child is sorted before a parent) \citep{suj_2024}. We note that $A_{top}$ is a normalized version of the topological ordering divergence $D_{top}$ defined in \cite{rolland_score_2022} (see Appendix \ref{appendix: atop def} for details). To produce a predicted causal graph, we apply a standard edge pruning method (CAM-pruning, \citealt{buhlmann_cam_2014}) to the fully dense graph corresponding to each topological ordering. We adopt the Structural Hamming Distance (SHD) \citep{tsamardinos_max-min_2006} and F1 score for evaluation; the SHD is the sum of false positive, false negative and reversed edges (lower SHD is better), whereas the F1 score measures the balance between precision and recall of predicted edges (higher F1 is better). 

% \begin{figure}[t!]%{0.8\textwidth}
% \centering
%      \begin{subfigure}[t]{\columnwidth}  % 4th row, 1st column
%         \includegraphics[width=\columnwidth]{figures/unifnew2.png}
%     \end{subfigure}
    
%      \begin{subfigure}[t]{\columnwidth}  % 4th row, 1st column
%         \includegraphics[width=\columnwidth]{figures/laplanew2.png}
%     \end{subfigure}
%     \caption{LoSAM performance on sparse graphs by linear mechanism proportion. Top row: uniform noise. Bottom row: Laplace noise.} 
%     \label{fig: sparse results brokendown}
% \end{figure}
\begin{figure}[t!]%{0.8\textwidth}
\centering
     \begin{subfigure}[t]{\columnwidth}  % 4th row, 1st column
        \includegraphics[width=\columnwidth]{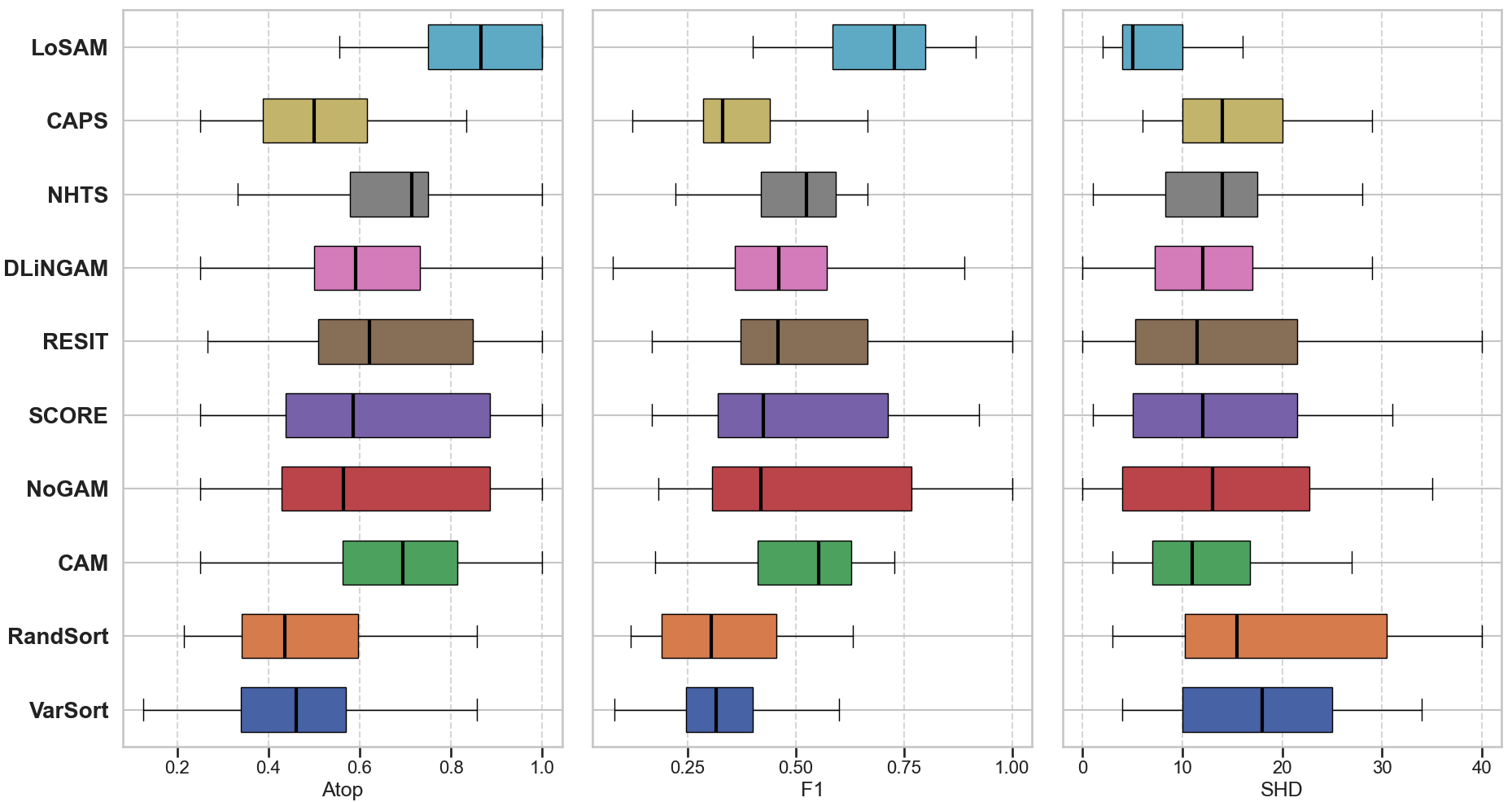}
    \end{subfigure}
     \begin{subfigure}[t]{\columnwidth}  % 4th row, 1st column
        \includegraphics[width=\columnwidth]{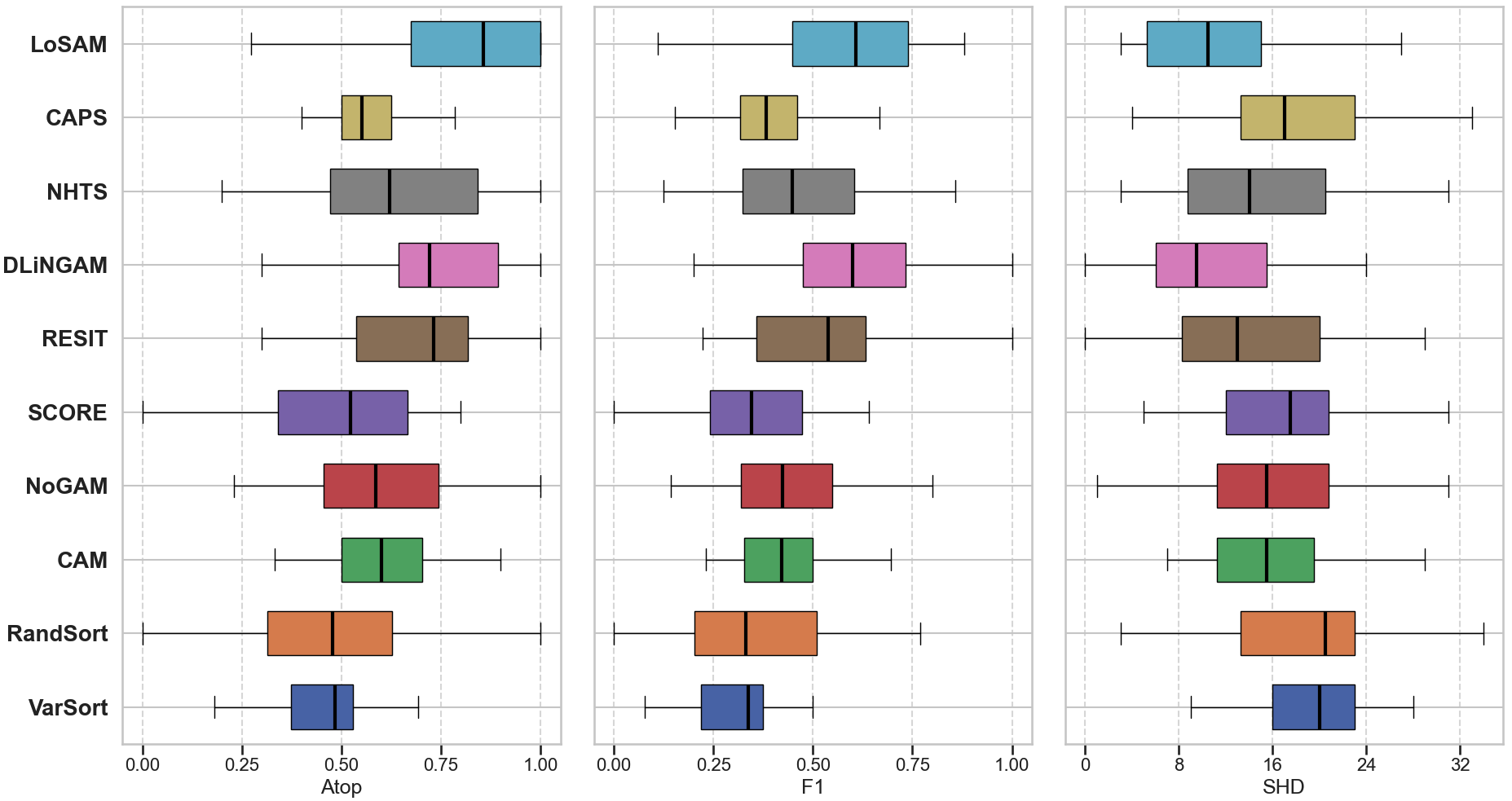}
    \end{subfigure}
    \caption{Performance of LoSAM on ER1 synthetic data with $50\%$ proportion of linear mechanisms. Top row: uniform noise. Bottom row: Laplace noise.
    }
    \label{fig: sparse results onespot}
\end{figure}
% SHD measures the number of edges that must be altered to make the predicted graph match the true causal graph 

% \subsection{Synthetic Data}\label{sec: syn data}
\textbf{Results on Synthetic Data.\;\;} Figure \ref{fig: sparse results overall} demonstrates the robustness of LoSAM as the proportion of linear and nonlinear mechanisms changes in sparse graphs (ER1). LoSAM achieves similar performance to SOTA methods in both the linear and nonlinear settings, under both noise distributions. In the mixed mechanism setting, LoSAM significantly overperforms all baselines in the uniform noise setting, and all baselines except DirectLiNGAM (which performs poorly in nonlinear ANM) in the laplacian setting; the degradation in baseline performance when assumptions are violated highlights the limited applicability of current FCM methods.

% \textbf{Overall Performance.\;}
Figure \ref{fig: sparse results onespot} demonstrates the overall performance of LoSAM in sparse graphs (ER1) where the proportion of linear mechanisms is fixed to $50\%$.
%We find that LoSAM outperforms all baselines in all experiments with nonlinear or mixed ANM, achieving the highest median $A_{top}$ in all trials.  The enhanced performance in the nonlinear setting is likely attributed to increased stability with finite samples, achieved through the use of novel test statistics. Further, we test the robustness of LoSAM under linear data-generating mechanisms (Figure~\ref{fig: main exp results} left column). We observe that 
LoSAM outperformed all baselines across all three metrics, especially for uniform noise, demonstrating the enhanced sample-efficiency of our root and leaf based procedure. We note that, as expected from Theorem \ref{prop: rootidcondset} and Theorem \ref{theorem: losam_time_red}, LoSAM had higher sample computational efficiency than nonlinear methods NHTS, RESIT and NoGAM (SCORE and CAM as well), with up to 2-5$\times$ faster runtime (see Appendix \ref{appendix: runtime}). 
%and achieved similar performance to DirectLiNGAM, a specialized method that leverages the linearity property. 
%\sujai{sample eff vs incorrectness distinction}

\begin{figure}[t!]%{0.8\textwidth}
\centering
     \begin{subfigure}[t]{\columnwidth}  % 4th row, 1st column
        \includegraphics[width=\columnwidth]{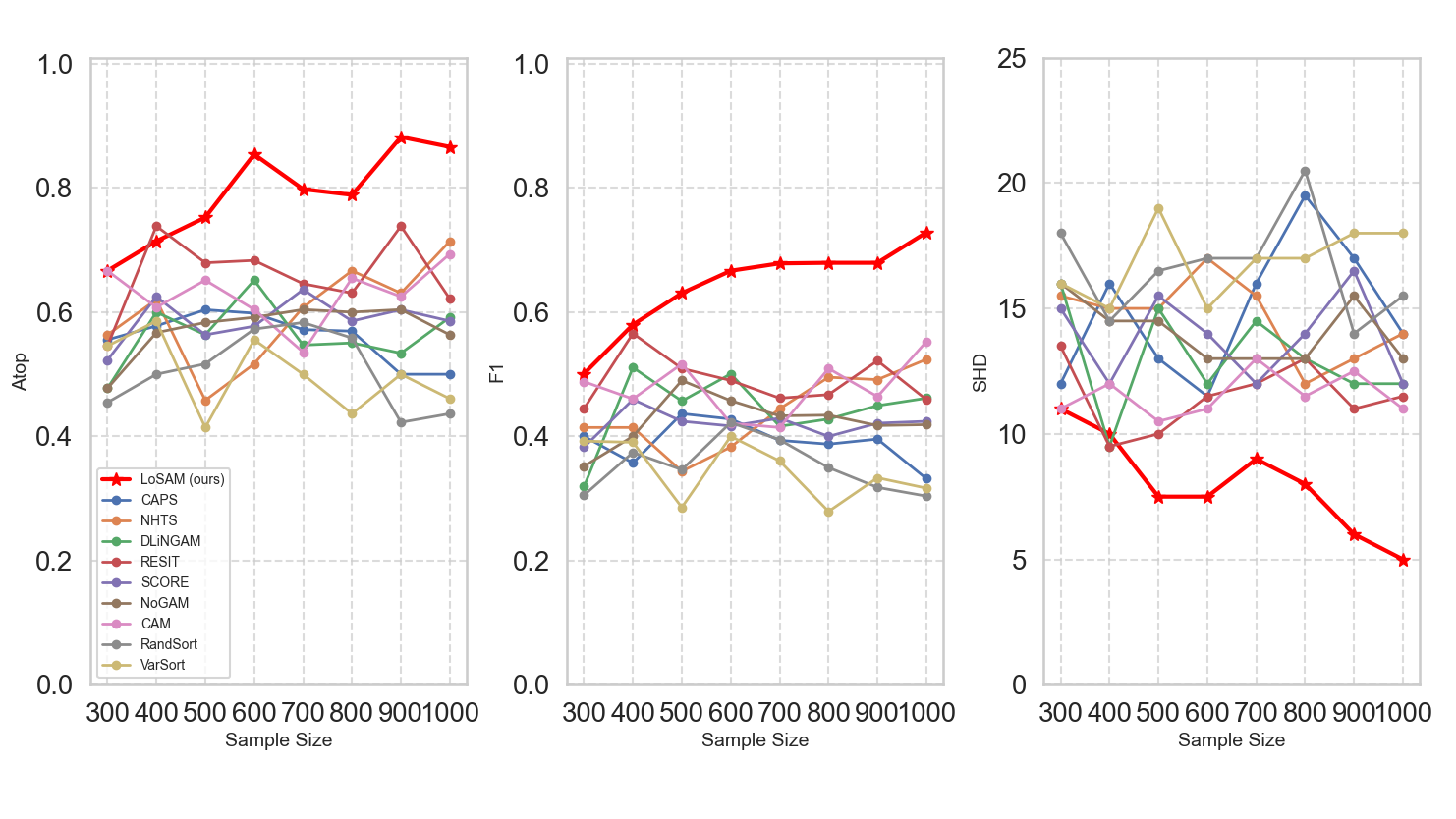}
    \end{subfigure}
    
    \caption{LoSAM performance on ER1 synthetic data with Uniform noise and $50\%$ proportion of linear mechanisms, with increasing sample size $n=300,400\ldots,1000$.} 
    \label{fig: sparse results onespotsample}
\end{figure}

Figure \ref{fig: sparse results onespotsample} shows how the performance of LoSAM changes as the sample size increases from $n=300$ to $n=1000$ (in sparse ER1 graphs with $d=10$, Uniform noise, $50\%$ linear mechanisms). The results demonstrate LoSAM's superior sample efficiency, as its performance improves fastest with increasing sample size and it consistently outperforms baselines. This aligns with Theorems \ref{prop: rootidcondset} and \ref{theorem: losam_time_red}, and shows that reduced conditioning set sizes do indeed enhance statistical efficiency.

% \textbf{Mechanism Specific Performance.\;}
\textbf{Additional Synthetic Experiments.\;}
To confirm the robustness of LoSAM, we present results in dense, high-dimensional, and Gaussian noise settings. Additionally, we test the sensitivity of LoSAM to estimation error. 

% \textbf{Dense Graphs.\;}
In Appendix \ref{appendix: dense graphs}, we examine the performance of LoSAM on denser graphs ($d=10,n=1000$).
% To further characterize the sample efficiency of LoSAM, we provide further evaluation on dense graphs (Appendix \ref{appendix: dense graphs}). 
We find that LoSAM still maintains a performance gap over baselines, likely due to the reduced covariate set size shown in Theorem \ref{theorem: losam_time_red}.

% \begin{table}[t!]
%     \centering
%     \renewcommand{\arraystretch}{1.2}
%     \setlength{\tabcolsep}{6pt}
%     \begin{tabular}{lccc}
%         \toprule
%         \textbf{Metrics} & $\mathbf{A_{top}}$\textbf{↑} & \textbf{F1↑} & \textbf{SHD↓} \\
%         \midrule
%         DirectLiNGAM & 0.56$\pm$0.13 & \underline{0.32$\pm$0.10} & \textbf{29.90$\pm$4.13} \\
%         CAM & \underline{0.61$\pm$0.12} & 0.30$\pm$0.07 & 38.60$\pm$4.87 \\
%         RESIT & 0.41$\pm$0.09 & 0.21$\pm$0.08 & 35.07$\pm$4.23 \\
%         SCORE & 0.28$\pm$0.02 & 0.17$\pm$0.03 & 37.67$\pm$3.10 \\
%         NoGAM & 0.28$\pm$0.03 & 0.17$\pm$0.03 & 38.50$\pm$4.05 \\
%         NHTS & 0.59$\pm$0.12 & 0.28$\pm$0.08 & 35.97$\pm$5.41 \\
%         CaPS & 0.29$\pm$0.03 & 0.18$\pm$0.02 & 37.03$\pm$2.52 \\
%         VarSort & 0.46$\pm$0.03 & 0.25$\pm$0.03 & 35.33$\pm$2.47 \\
%         RandSort & 0.47$\pm$0.13 & 0.24$\pm$0.08 & 38.33$\pm$5.54 \\
%         % \midrule
%         \textbf{LoSAM} & \textbf{0.62$\pm$0.09} & \textbf{0.33$\pm$0.05} & \underline{34.97$\pm$4.74} \\
%         \bottomrule
%     \end{tabular}
%     \caption{Results on real-world Sachs protein dataset.}
%     \label{fig: sachs_data}
% \end{table}

% \textbf{Higher Dimensional Experiments.\;} 
To understand the scalability of LoSAM we fix the proportion of linear mechanisms to $0.5$, sample size $n=2000$, noise to be uniform, and evaluate over increasing dimensionality ($d=10,20,\ldots,50$) in Appendix \ref{appendix: high dim graphs}.
We find that LoSAM maintains performance gains over baselines, demonstrating its efficacy in higher-dimensional settings.

% \textbf{Nonlinear Gaussian Experiments.\;}
We test the performance of LoSAM under the nonlinear Gaussian setting in Appendix \ref{appendix: nonlin gaussian graphs} ($d=10,n=1000$), a setting that is considered in prior benchmarks.
% ANMs with linear mechanisms and Gaussian noise are known to be unidentifiable without strong assumptions \citep{ParkKim2020}, so we test the efficiency of LoSAM in the nonlinear Gaussian setting (see Appendix \ref{appendix: nonlin gaussian graphs}). 
We find that LoSAM achieves similar performance to CAM,  which is designed to exploit Gaussian noise in its procedure, and significantly outperforms all other baselines (including other Gaussian-specific methods, e.g., SCORE and CaPS).

% Additionally, in Appendix \ref{appendix: estim error sensit} we investigate the sensitivity of LoSAM to estimation error ($d=10, n=300$, Uniform noise). We evaluate LoSAM variants with different regression estimators (Random Forest, Kernel Ridge Regression with different kernels, etc.). We find that LoSAM shows consistent accuracy across estimators; this confirms LoSAM's robustness to nonparametric estimation error, a crucial property for finite-sample applications where regression efficiency varies.
Finally, in Appendix \ref{appendix: estim error sensit}, we investigate LoSAM’s sensitivity to estimation error ($d=10, n=300$, Uniform noise) by evaluating LoSAM variants that use different regression estimators (Random Forest, Kernel Ridge Regression with various kernels, etc.). LoSAM maintains consistent accuracy across estimators, confirming its robustness—an essential property for finite-sample settings where regression efficiency may vary.

% \subsection{Real Data}\label{sec: real data}
\textbf{Results on Real-World Data.\;\;} The results of the Sachs dataset are presented in Table \ref{fig: sachs_data}: LoSAM achieves the best $A_{top}$ and F1 score, with SHD coming in second to DirectLiNGAM. We note that the SHD metric favors the prediction of sparse causal graphs \citep{caps_xu2024ordering}, as it double counts errors when edges are reversed, whereas F1 and $A_{top}$ penalize all errors equally. Therefore, taking all three metrics into account, we conclude that LoSAM outperforms baselines.

% \begin{figure}[t!]%{0.8\textwidth}
% \centering
%      \begin{subfigure}[t]{\columnwidth}  % 4th row, 1st column
%         \includegraphics[width=\columnwidth]{figures/sampleefficiencyplotfinal.png}
%     \end{subfigure}
    
%     \caption{LoSAM performance on ER1 synthetic data with Uniform noise and $50\%$ proportion of linear mechanisms, with increasing sample size $n=300,400\ldots,1000$.} 
%     \label{fig: sparse results brokendown}
% \end{figure}
\begin{table}[t!]
    \centering
    \renewcommand{\arraystretch}{1.2}
    \setlength{\tabcolsep}{6pt}
    \begin{tabular}{lccc}
        \toprule
        \textbf{Metrics} & $\mathbf{A_{top}}$\textbf{↑} & \textbf{F1↑} & \textbf{SHD↓} \\
        \midrule
        DirectLiNGAM & 0.56$\pm$0.13 & \underline{0.32$\pm$0.10} & \textbf{29.90$\pm$4.13} \\
        CAM & \underline{0.61$\pm$0.12} & 0.30$\pm$0.07 & 38.60$\pm$4.87 \\
        RESIT & 0.41$\pm$0.09 & 0.21$\pm$0.08 & 35.07$\pm$4.23 \\
        SCORE & 0.28$\pm$0.02 & 0.17$\pm$0.03 & 37.67$\pm$3.10 \\
        NoGAM & 0.28$\pm$0.03 & 0.17$\pm$0.03 & 38.50$\pm$4.05 \\
        NHTS & 0.59$\pm$0.12 & 0.28$\pm$0.08 & 35.97$\pm$5.41 \\
        CaPS & 0.29$\pm$0.03 & 0.18$\pm$0.02 & 37.03$\pm$2.52 \\
        VarSort & 0.46$\pm$0.03 & 0.25$\pm$0.03 & 35.33$\pm$2.47 \\
        RandSort & 0.47$\pm$0.13 & 0.24$\pm$0.08 & 38.33$\pm$5.54 \\
        % \midrule
        \textbf{LoSAM} & \textbf{0.62$\pm$0.09} & \textbf{0.33$\pm$0.05} & \underline{34.97$\pm$4.74} \\
        \bottomrule
    \end{tabular}
    \caption{Results on real-world Sachs protein dataset.}
    \label{fig: sachs_data}
\end{table}

\textbf{Discussion.\;}
% In this paper, we developed a novel topological ordering algorithm for use in global causal discovery. LoSAM (Alg \ref{algo: LoSAM}) improved on previous local search topological ordering methods by extending asymptotic guarantees from ANMs with purely linear or nonlinear mechanims to the full ANM setting, while improving sample efficiency.
% We tested LoSAM on a real biological dataset and across a range of robustly generated synthetic data, finding that our method achieved similar or better performance than baselines.
Future work includes extending LoSAM to handle latent confounding, and applying the local search approach to a time series setting. Additionally, we aim to build upon our theoretical guarantees and develop statistical sample complexity bounds for LoSAM, extending previously derived results \citep{sample_bound_2023} for FCM methods in nonlinear Gaussian ANMs.

% \paragraph{Root Recovery}
% see appendix, for comparison of root recovery for dlingam and nhts and losam.
% \paragraph{Comparison to Traditional Discovery Methods}
% SHD and F1 for skeleton recovery (only recover up to MEC). (compare to GRaSP, GES on sparse graphs) see appendix

\textbf{Acknowledgments}~\\
This paper is supported by an AWS Credits Grant from The Center for Data Science for Enterprise and Society at Cornell University.

\bibliography{references}

\onecolumn

\makeatletter
\let\AB@authors\@empty
\let\AB@affilnotes\@empty
\let\AB@affillist\@empty
\let\AB@addresses\@empty
\makeatother

\makeatletter
\patchcmd{\maketitle}{\AB@authors}{}{}{}
\patchcmd{\maketitle}{\AB@affillist}{}{}{}
\makeatother

\appendix
% Forcefully redefine the title to "APPENDIX", bc otherwise \maketitle reprints authors and affiliations
\makeatletter
\renewcommand{\maketitle}{%
    \begingroup
    \let\AB@authors\@empty
    \let\AB@affilnotes\@empty
    \let\AB@affillist\@empty
    \let\AB@addresses\@empty
    \let\@title\@empty
    \def\@title{APPENDIX} % Explicitly set the title here
    \renewcommand{\AB@maketitle}{%
        \hrule height4pt
        \vskip .25in
        \centering
        {\Large\bfseries \@title\par}%
        \vskip .25in
        \hrule height1pt
        \vskip .25in
    }%
    \AB@maketitle
    \endgroup
}
\makeatother

% Call \maketitle again for the appendix
\maketitle

%\input{main.bbl}

%\vspace{-0.1cm}
\vspace{0cm}
%\vspace{-20mm}
%\raggedbottom
\section{NOTATION}
\begin{description}[
  leftmargin=2.2cm,
  labelwidth=2cm,
  labelsep=.2cm,
  parsep=0mm,
  itemsep=2pt
  ]
\item[$G=(V,E)$] A DAG $G$ with $|V|=d$ vertices, where $E$ represents the set of directed edges between vertices.

\item[$\mathrm{Ch}(x_i)$] The set of child vertices of $x_i$. 

\item[$\mathrm{Pa}(x_i)$] The set of parent vertices of $x_i$. 

\item[$\mathrm{De}(x_i)$] The set of vertices that are descendants of $x_i$. 

\item[$\mathrm{An}(x_i)$] The set of vertices that are ancestors of $x_i$. 

\item[$\pi$] A topological ordering of vertices, i.e. a mapping $\pi: V \rightarrow \{0,1,\ldots,d\}$.

\item[$f_i$] An arbitrary function, used to generate vertex $x_i$.

\item[$\varepsilon_i$] An independent noise term sampled from an arbitrary distribution,  used to generate vertex $x_i$.

\item[$\mathrm{SRD}$] A single root descendant, a vertex with only one root ancestor.

\item[$\mathrm{MRD}$] A multi root descendant, a vertex with at least two root ancestors.

\item[$\mathrm{VP}$] A v-pattern, a causal substructure between three vertices.

\item[$V'$] A subset of $V$ that contains only and all roots and SRDs.

\item[$V''$] A subset of $V'$ that contains only and all SRDs and roots with at least one SRD.

\item[$r_{ij}$] The residual of the vertex $x_j$ nonparametrically regressed on the vertex $x_i$.

\item[$W$] A subset of $V''$ that contains only and all vertices identified by the Regression-Identification Test.

\item[$M$] The set of MRDs in $V$.

\item[$R$] The set of roots in $V$.

\item[$c^\mathrm{Alg1}_\mathrm{max}$] The maximum size of conditioning sets used in Algorithm \ref{algo: observed root id}.

\item[$c^\mathrm{DirectLiNGAM}_\mathrm{max}$] The maximum size of conditioning sets used in the root identification for DirectLiNGAM.

\item[$c^\mathrm{NHTS}_\mathrm{max}$] The maximum size of conditioning sets used in the root identification step of NHTS.

\item[$c^\mathrm{RESIT}_\mathrm{max}$] The maximum size of conditioning sets used to identify roots in RESIT.

\item[$c^\mathrm{NoGAM}_\mathrm{max}$] The maximum size of conditioning sets used to identify roots in NoGAM.

\item[$U$] The set of vertices left unsorted, i.e., not yet added to $\pi$.

\item[$\mathrm{VLC}$] A valid leaf candidate; a vertex in $U$ with no parents in $U$.

\item[$U_E$] The set of residuals produced from regressing each $x_i \in U$ onto vertices in $\pi$.

\item[$e_i$] A residual in $U_E$ corresponding to $x_i \in U$.

\item[$\mathrm{ND}$] A nonlinear descendant; a vertex in $U$ that is a nonlinear function of at least one vertex in $U$.

\item[$\mathrm{LD}$] A linear descendant; a vertex in $U$ that is a linear function of all vertices in $U$.

\item[$q_{ij}$] The residual produced by linearly regressing $e_j$ onto $e_i$.

\item[$Q$] A subset of $U$ that contains all LDs, but no VLCs.

\item[$U'$] A subset of $U$ that contains all VLCs and some NDs.

\item[$\widehat{MI}(x,y)$] A nonparametric estimator of mutual information between $x,y$. 

\item[$t^*(e_i,\pi)$] A test statistic corresponding to $x_i \in U'$.

\item[$r_{ij}^k$] The residual produced by nonparametrically regressing $x_k$ onto $x_i,x_j$.
\end{description}

\newpage
\section{GRAPH TERMINOLOGY}\label{appendix: graph term}
In this section we clarify the term `d-separation', a foundational concept for analyzing causal graphical models \citep{peter_spirtes_causation_2000}. First, we classify the types of paths that exist between vertices, then define d-separation in terms of those paths.  

We first note that a path between $x_i,x_k$ can either start and end with an edge out of $x_j,x_k$ ($x_k\dashrightarrow \cdots \dashleftarrow x_k $), start with an edge of out of $x_j$ and end with an edge into $x_k$ ($x_j \dashrightarrow \cdots \dashrightarrow x_k$) or start with an edge into $x_j$ and edge with an edge into $x_k$ ($x_j\dashleftarrow \cdots \dashrightarrow x_k $). Paths such as $x_k$ ($x_k\dashrightarrow \cdots \dashleftarrow x_k $) do not transmit causal information between $x_j,x_k$. Undirected paths that transmit causal information between two vertices $x_j,x_k$ can be differentiated into \textit{frontdoor} and \textit{backdoor paths} \citep{peter_spirtes_causation_2000}. A frontdoor path is a directed path $x_j \dashrightarrow \cdots \dashrightarrow x_k$, while a backdoor path is a path $x_j\dashleftarrow \cdots \dashrightarrow x_k $.

Paths between two vertices are further classified, relative to a vertex set \(\textbf{Z}\), as either \textit{active} or \textit{inactive} \citep{peter_spirtes_causation_2000}. A path between vertices $x_j, x_k$ is active relative to \(\textbf{Z}\) if every node on the path is active relative to \(\textbf{Z}\). Vertex $x_i \in V$ is active on path relative to $\textbf{Z}$ if one of the following holds: 1) $x_i \in \textbf{Z}$ and $x_i$ is a collider 2) $x_i \not \in \textbf{Z}$ and $x_i$ is not a collider 3) $x_i \not \in \textbf{Z}$, $x_i$ is a collider, but $\text{De}(x_i) \cap \textbf{Z} \neq \emptyset$. An inactive path is simply a path that is not active. Causally paths are typically described active or inactive with respect to $\textbf{Z} = \emptyset $ unless otherwise specified.

Vertices $x_i, x_j$ are said to be d-separated by a set $\textbf{Z}$ iff there is no active path between $x_i,x_j$ relative to $Z$.

\section{ASSUMPTIONS}\label{appendix: assum and method}

\subsection{Causal Markov}
The Causal Markov condition implies that a variable $x_i$ is independent of all non-descendants $x_j$, given its parents $\text{Pa}(x_i)$ \citep{spirtes_anytime_2001}. Equivalently, we say that an ANM $G=(V,E)$ satisfies the Causal Markov Condition if the joint distribution $p_V(V)$ over all $x_i \in V$ admits the following factorization:
\begin{equation}
    p_V(V) = \prod_i^d p_i(x_i|\text{Pa}(x_i)).
\end{equation}

\subsection{Acyclicity}
We say that a causal graph $G$ is acyclic if there does not exist any directed cycles in $G$ \citep{spirtes_anytime_2001}.

\subsection{Faithfulness}
Note that by assuming that a causal graph $G$ satisfies the Causal Markov assumption, we assume that data produced by the DGP of $G$ satisfies all independence relations implied by $G$ \citep{spirtes_anytime_2001}. However, this does not necessarily imply that all independence relations observed in the data are implied by $G$. Under the additional assumption of faithfulness, the independence relations implied by $G$ are the \emph{only} independence relations found in the data generated by $G$'s DGP \citep{spirtes_anytime_2001}.

\subsection{Identifiability of ANM}\label{appendix: anm_identify}
Following the style of \citep{montagna_assumption_2023}, we first observe that the following condition guarantees that the observed distribution of a pair of variables $x_i,x_j$ can only be generated by a unique ANM:
\begin{condition}[\citealt{hoyer_bayesian_2009}]\label{theorem:bivariate_anm}
Given a bivariate model $x_i = \varepsilon_i, x_j = f_j(x_i)+\varepsilon_j$ generated according to \eqref{eq:ANM}, we call the SEM an identifiable bivariate ANM  if the triple $(f_i,p_{\varepsilon_i},p_{\varepsilon_j})$ does not solve the differential equation $k'''=k''(-\frac{g'''f'}{g''}+\frac{f''}{f'})-2g''f''f'+g'f'''+\frac{g'g'''f''f'}{g''}-\frac{g'(f'')^2}{f'}$ for all $x_i,x_j$ such that $f'(x_i)g''(x_j-f_j(x_i))\neq0$, where $p_{\varepsilon_i},p_{\varepsilon_j}$ are the density of $\varepsilon_i,\varepsilon_j$, $f=f_j,k=\log p_{\varepsilon_i}, g=p_{\varepsilon_j}$. The arguments $x_j-f_j(x_i),x_i$ and $x_i$ of $g,k$ and $f$ respectively, are removed for readability.
\end{condition}

There is a generalization of this condition to the multivariate ANM proved by \citep{peters_causal_2014}:
\begin{theorem}\label{theorem: multivariate_anm}
(\citealt{peters_causal_2014}). An ANM corresponding to DAG $G$ is identifiable if $\forall x_j \in V, x_i \in \text{Pa}(x_j)$ and all sets $S \subseteq V$ with $\text{Pa}(x_j) \setminus \{i\} \subseteq S \subseteq \overline{\text{De}(j)}\setminus{\{x_i, x_j\}}$, $\exists$ $X_S$ with positive joint density such that the triple $\Big{(}f_j(\text{Pa}(j)\setminus{\{x_i\}},x_i), p_{x_i|X_s},p_{\varepsilon_j}\Big{)}$ satisfies Condition \ref{theorem:bivariate_anm}, and $f_j$ are non-constant in all arguments.
\end{theorem}

In this paper, we assume that all DAGs are generated by identifiable ANMs, as defined in Theorem \ref{theorem: multivariate_anm}.

Our work builds on the classical ANM identifiability assumptions from \citep{peters_causal_2014} (Theorem \ref{theorem: multivariate_anm}).
% Formally, we require the conditions in Theorem C.2, which restricts mechanism-noise combinations that solve the differential equation in Condition C.1.
Intuitively, these ensure each ANM generates a unique joint distribution, enabling unique identification.
This identifiability condition rules out specific mechanisms-noise pairs such as linear $f_i$ and Gaussian $\epsilon_i$.
Formally, for any
$x_i = f_i(Pa(x_i)) + \varepsilon_i$ where $f_i$ is linear in at least one of the parents in $Pa_i$, then $\varepsilon_i$ must be non-gaussian. Thus, 
LoSAM does not cover linear Gaussian ANMs (LiGAM) but applies to any linear, nonlinear, or mixed-mechanism setting satisfying Theorem \ref{theorem: multivariate_anm}.

While LiGAMs are generally non-identifiable, additional assumptions can enable identification: equal/known error variance \citep{peters2014identifiability}, heteroscedastic errors via variance/edge-weight conditions \citep{ghoshal2018learning, ParkKim2020}, conditional variance constraints \citep{park2020identifiability}, Varsortability/$R^2$-sortability \citep{reisach_beware_2021, reisach_scale-invariant_2023}, and score function restrictions \citep{caps_xu2024ordering}.
While extending LoSAM to LiGAM is an interesting research direction, the focus of this work is on providing a polynomial-time discovery algorithm for general ANMs with mixed mechanisms, without relying on strong functional, variance, or distributional assumptions. We leave extensions to LiGAM for future work.

\subsection{ASSUMPTIONS OF CAPS ON NOISE DISTRIBUTION}\label{appendix: caps_assum}

The causal discovery method CaPS \citep{caps_xu2024ordering} requires that the noise terms be Gaussian, and that at least one of the following conditions holds (see section 3.1 and 4.1 of their paper):

\textbf{CaPS Assumptions} \textit{(Sufficient conditions for identifiability).}

\begin{enumerate}
    \item \textit{Non-decreasing variance of noises.} For any two noises $\epsilon_i$ and $\epsilon_j$, $\sigma_j \geq \sigma_i$ if $\pi(i) < \pi(j)$.
    \item \textit{Non-weak causal effect.} For any non-leaf nodes $x_j$,
    \[
    \sum_{i \in Ch(j)} \frac{1}{\sigma_i^2} \mathbb{E} \left[ \left( \frac{\partial f_i}{\partial x_j} (pa_i(x)) \right)^2 \right] \geq \frac{1}{\sigma_{\min}^2} - \frac{1}{\sigma_j^2},
    \]
\end{enumerate}

where $\sigma_{\min}$ is the minimum variance for all noises. They comment that condition 1 is an extension of the equal variance assumption \cite{peters2014identifiability}, while condition 2 is a new sufficient condition that quantifies a lower bound of identifiable causal effects.
\\
~
\\
We note that, in contrast, LoSAM does not require or leverage either of the above conditions to recover a correct topological ordering, and allows for general noise distributions (beyond Gaussian noise).

\section{LEMMA PROOFS}\label{appendix: lemmas}

\subsection{Proof of Lemma \ref{lemma: MRDVP}}\label{appendix: MRDVP}
\MRD*
\begin{proof}
Suppose $x_i$ is an MRD. Then, $\exists x_j,x_k$ such that $x_j,x_k \in \text{An}(x_i)$, and $x_j,x_k$ are roots. Note that this implies that $x_j \ind x_k, x_i \nind x_j, x_i \nind x_k$, which is a VP between $x_j,x_k$ induced by $x_i$.

Suppose $x_i$ is not an MRD. We prove by contradiction that $x_i$ cannot induces a VP.

Suppose $x_i$ is an SRD. Let $x_p$ be any vertex such that $x_p \nind x_i$, which implies that there must exist an active causal path between $x_i, x_p$. Note, as $x_i$ is an SRD, there exists only one root vertex $x_j$ such that $x_j \in \text{An}(x_i)$. Suppose for contradiction that $x_p \not\in \text{De}(x_j)$. Suppose there is a frontdoor path from $x_i$ to $x_p$: then $x_p \in \text{De}(x_i)\implies x_p \in \text{De}(x_j)$, which contradicts our assumption that $x_p \not\in \text{De}(x_j)$. Suppose there is a frontdoor path from $x_p$ to $x_i$: as $x_p \not\in\text{De}(x_j)$, WLOG $\exists$ a root $x_m\neq x_j$ such that $x_p \in \text{De}(x_m)$. This implies that $x_i \in \text{De}(x_m)$ which contradicts the assumption that $x_i$ is an SRD. Suppose for contradiction that there exists a backdoor path between $x_p,x_i$: this implies that $\exists$ $x_c$ that is a confounder between $x_p,x_i$. Again, this implies that WLOG $\exists$ a root $x_m\neq x_j$ such that $x_c \in \text{De}(x_m) \implies x_i \in \text{De}(x_m)$, which contradicts the assumption that $x_i$ is an SRD. Therefore, it must be that, for any vertex $x_p \nind x_i$, $x_i, x_p\in \text{De}(x_j)$, where $x_j$ is a root. Note, the above implies that for any vertices $x_j, x_k$ such that $x_j \nind x_i, x_k \nind x_i$, we have $x_j, x_k \in \text{De}(x_j) \implies x_j \nind x_k$. This implies that $x_i$ cannot induce a VP between any two vertices $x_j,x_k$.

Suppose $x_i$ is a root. Let $x_j,x_k$ be any two vertices such that $x_j\nind x_i, x_k \nind x_i$. Note that as $x_i$ is a root, there can only exist frontdoor paths between $x_i$ and other vertices, so the dependence relations imply that $x_j, x_k \in \text{De}(x_i)$. This means that $x_i$ is a confounder of $x_j,x_k \implies x_j \nind x_k$. Therefore, $x_i$ cannot induce a VP between $x_j,x_k$.
\end{proof}
\subsection{Proof of Lemma \ref{lemma: rootnoSRD}}\label{appendix: rootnoSRD}
\rootnoSRD*
\begin{proof}
    Note that $V'$ is the union of vertices that are either SRDs or root vertices. We prove this by contradiction. Suppose that $x_i$ is an SRD: then $\exists x_k \in V'$ such that $x_k$ is a root and $x_i \in \text{De}(x_k)$ which implies $ x_i \nind x_k$. However, this contradicts our assumption that $x_i \ind x_j, \forall x_j \in V'$. Now we show that $x_i$ cannot a root vertex with SRDs.  Suppose that $x_i$ is a root vertex with SRDs: then $\exists x_k \in \text{De}(x_i) \implies x_i \nind x_k$. However, this contradicts our assumption that $x_i \ind x_j, \forall x_j \in V'$. Additionally, note that a root vertex $x_i$ with no SRDs satisfies $x_i \ind x_j, \forall x_j \in V'$. Therefore, $x_i$ must be a root vertex with no SRDs.

     Suppose $x_i$ is a root with no SRDs. Note that $V'$ contains only SRDs and roots. Then, as $x_i$ has no SRDs, it has no descendants. As $x_i$ is a root, it has no ancestors; therefore, $x_i \ind x_j$ $\forall x_j \in V'\setminus{\{x_i\}}$.
\end{proof}

\subsection{Proof of Lemma \ref{lemma: root id step}}\label{appendix: root id step}
\rootidstep*
\begin{proof}

% In this proof, we say that $x_i$ is in `AD relation' to $x_k$ iff $x_i$ is identified as $\in\text{An}(x_k)$ by the Ancestor Descendant test (Definition \ref{def: ancestordescendent}).

In this proof, we say that \(x_i\) is in  'AD relation' to \(x_k\) iff \(x_i\) is identified as \(\in An(x_k)\) by the Regression-Identification Test (Definition \ref{def: ancestordescendent})

 By definition, any root vertex $x_i \in V''$ has at least one SRD $x_j$, such that $x_j \in \text{Ch}(x_i)$. Note that as $x_j = f_{ij}(x_i) + \varepsilon_j$, $x_i$ is in AD relation to $x_j$. 
 %Note that as $x_i$ is a root vertex, all $x_j \in V''$ are either independent of $x_i$ or descendants of $x_i$. 
 Note that as $x_i$ is a root vertex, all $x_j \in V''$ are not ancestors of $x_i$.
 If $x_j \ind x_i$, then $x_j$ will not be in AD relation to $x_i$ as $x_i \ind r_{ij}$, violating condition 2) in the AD definition. If $x_j \nind x_i$, then $x_j \not\in \text{An}(x_i)$, and therefore by assumption of the restricted ANM \ref{theorem: multivariate_anm} we have that $x_j \nind r_{ji}$, implying that $x_j$ is not in AD relation to $x_i$. This implies that all root vertices in $V''$ satisfy condition 1) and 2) of the root superset $W$, implying that all root vertices in $V''$ are contained in $W$.

Let $x_i \in W$ be a root. Consider $x_j$ such that $ x_j \nind x_i$. Consider any $x_k$ such that $x_j$ is in AD relation to $x_k$. Note that $x_j \nind x_i$ and $x_i$ being a root implies that $x_i$ is not a descendant of either $x_j$ or $x_k$. Additionally, as $x_j$ is in AD relation to $x_k$, $x_i$ cannot be a confounder of $x_j,x_k$. This implies that $x_i \ind x_k | x_j$, which implies that $r_{ij}^k \ind x_i$. 

Let $x_i \in W$ be a nonroot. Then, $\exists x_j \nind x_i$ such that $x_j$ is the root ancestor of $x_i$. Note that $x_i$ is in AD relation to all of its children, $\text{Ch}(x_i)$. Note that $\exists x_k \in \text{Ch}(x_i)$ such that $x_k$ is an ancestor of $x_i$. As $x_i$ is a descendant of $x_k$, this implies that $x_i \nind r_{ij}^k$.

More intuitively: if $x_j$ is identified as $\in\text{An}(x_k)$, implying that $x_j$ is independent of residual of $x_k$ regressed onto $x_j$, $x_j$ should remain independent of residual produced by the bivariate regression of $x_k$ onto $x_i,x_j$. In contrast, for any non-root SRD $x_p$ included in $W$ and identified as in $\text{An}(x_h)$, there exists a root ancestor $x_l \in W$; therefore, if $x_h$ is regressed onto $x_l,x_p$ the resulting residual will be dependent on $x_p$.

\end{proof}

\subsection{Proof of Lemma \ref{lemma: ND prune}}\label{appendix: ND prune}
\rootidcorr*
\begin{proof}
    Suppose $x_i \in U$ is a ND. Then, by definition $\exists x_j \in U$ such that $x_j$ is a nonlinear function of its parents in $U$ ($\text{Pa}(x_j) \cap U$), and $x_j$ is an ancestor of $ND$ ($x_j \in \text{An}(x_i$)). Note that this implies that the regression of $x_i$ onto sorted vertices $\pi$ leads to omitted variable bias \citep{pearl_causal_2016}, which leads to $e_i$ being dependent on any sorted vertex $x_k \in \pi$ such that $x_k$ is an ancestor of $x_j$ ($x_k \in\text{An}(x_j)$). Therefore, $e_i$ is dependent on at least one sorted vertex in $\pi$.

    Suppose $e_i$ is dependent on at least on sorted vertex in $\pi$. Note that by assumption there are only nonlinear functions, $x_i$ is not an LD. Suppose for contradiction that $x_i$ is a VLC. Then, $x_i = f_i(\text{Pa}(x_i))+\varepsilon_i$. Note that as $\pi$ is a valid topological sort, it follows that $\text{Pa}(x_i)\subseteq \pi$ and $\text{De}(x_i)\cap \pi = \emptyset$. Therefore, $e_i  = \varepsilon_i$, which implies $e_i \ind x_j, \forall x_j \in \pi$. This implies that $e_i$ is independent of all vertices in $\pi$, contradicting our above assumption.

    %however, as shown in the proof of  Lemma \ref{lemma: LD}, we can rewrite the residual $e_i$ as just the independent error term:
    %\begin{equation}
    %    e_i = \varepsilon_i.
    %\end{equation}
    %This implies that $e_i$ is independent of all vertices in $\pi$, contradicting our above assumption. Therefore, $x_i$ must be an ND.
    %Let $x_i \in U$ be a VC. Then, $x_i = f_i(\text{Pa}(x_i))+\varepsilon_i$. Note that as $\pi$ is a valid topological sort, it follows that $\text{Pa}(x_i)\subseteq \pi$ and $\text{De}(x_i)\cap \pi = \emptyset$. Therefore, $r_i  = \varepsilon_i$, which implies $r_i \ind x_j, \forall x_j \in \pi$. 
\end{proof}
\begin{comment}
\subsection{Proof of Lemma \ref{lemma: VC ind}}\label{appendix: VC ind lemma}
\vcind*
\begin{proof}
    Let $x_i \in U$ be a VC. Then, $x_i = f_i(\text{Pa}(x_i))+\varepsilon_i$. Note that as $\pi$ is a valid topological sort, it follows that $\text{Pa}(x_i)\subseteq \pi$ and $\text{De}(x_i)\cap \pi = \emptyset$. Therefore, $r_i  = \varepsilon_i$, which implies $r_i \ind x_j, \forall x_j \in \pi$. 
\end{proof}
\end{comment}

\subsection{Proof of Lemma \ref{lemma: LD ind}}\label{appendix: LD ind}
\ldind*
\begin{proof}
    Let $x_i \in U$ be a LD. This implies that we can decompose $x_i$ into a potentially nonlinear function of its parents in $\pi$ and its the error terms of its ancestors in $U$:
    \begin{equation}
          x_i = f_i(\text{Pa}(x_i)\cap \pi) + \sum_{x_j \in \text{An}(x_i)\cap U} \alpha_{ij} \varepsilon_j + \varepsilon_i.
    \end{equation}
This implies that the residual of $x_i$ nonparametrically regressed onto $\pi$ equals
\begin{equation}
    e_i = \sum_{x_j \in \text{An}(x_i)\cap U} \alpha_{ij} \varepsilon_j + \varepsilon_i.
\end{equation}

As $\varepsilon$ are mutually marginally independent, we have that $e_i \ind x_j, \forall x_j \in \pi$.

\end{proof}

\subsection{Proof of Lemma \ref{lemma: LD}}\label{appendix: LD lemma}
\LDID*
\begin{proof}
    Suppose $x_i \in U$ is an LD. Let $x_j \in U$ be an ancestor of $x_i$ such that $x_j$ has no parents in $U$, i.e. $\text{Pa}(x_j)\cap U = \emptyset$. Such a vertex must always exist, as other $x_i$ would be a VLC.
    We can decompose $x_j$ into a function of parents in $\pi$, and an independent error term:
    \begin{equation}
        x_j = f_j(\text{Pa}(x_j)\cap \pi) + \varepsilon_j.
    \end{equation}
    We can decompose $x_i$ into a function of parents in $\pi$, a sum of linear functions of parents in $U$, and the independent error term $\varepsilon_i$:
    \begin{equation}
        x_i = f_i(\text{Pa}(x_i)\cap \pi) + \sum_{x_j \in \text{Pa}(x_i)\cap U} \alpha_{ij} x_j + \varepsilon_i.
    \end{equation}

Note that as $x_i$ is a LD, all of its ancestors in $U$ are linear functions of their parents in $U$. Therefore, we can further decompose the sum of ancestors in $U$ as a linear sum of error terms: 
    \begin{equation}
        x_i = f_i(\text{An}(x_i)\cap S) + \sum_{x_j \in \text{An}(x_i)\cap U} \beta_{ij} \varepsilon_j + \varepsilon_i.
    \end{equation}
Now, we consider the residuals $e_i,e_j \in U_E$, which we can write as:
\begin{equation}
   e_j = \varepsilon_j 
\end{equation}
and
\begin{equation}
    e_i = \sum_{x_j \in \text{An}(x_i)\cap U} \beta_{ij} \varepsilon_j + \varepsilon_i.
\end{equation}
As $e_i$ is a linear function of $e_j$, we have that $e_j \ind q_{ji}, e_i \nind q_{ij}$, implying that $x_i \in Q$. Therefore, $Q$ contains all LDs.

Let $x_i \in U$ be a VLC. Then, we can write $x_i$ as the sum of a function of parents (which are all contained in $\pi$) and independent error term:
\begin{equation}
    x_i = f_i(\text{Pa}(x_i)) + \varepsilon_i. 
\end{equation}
Note that $e_i \in U_E$ is then just the independent error term:
\begin{equation}
    e_i = \varepsilon_i.
\end{equation}
Note that $e_i$ has no parents in $U_E$: therefore, for any vertex $x_j$ with associated residual $e_j \in U_E$, either $ e_i \ind e_j \implies e_j \ind q_{ji}, e_i \ind q_{ij}$, or $e_i \nind e_j \implies e_j \nind q_{ji}, e_i \nind q_{ij}$. The last statement follows from the fact that if $e_i \nind e_j$, as $e_i$ is an independent error term, $e_j$ must be a function of $e_i$; then, the conclusion follows by assumption of ANM \ref{theorem: multivariate_anm}.
Therefore, $x_i \not\in Q$, and therefore no VLCs are included in $Q$.
\end{proof}

\begin{comment}
\subsection{Proof of Lemma \ref{lemma: VS top}}
\VSANC*
\begin{proof}
    Suppose for contradiction that $x_i$ is an ancestor of both $x_j,x_k$: then, $x_i$ would be a confounder of $x_j,x_k$, implying $x_j \nind x_k$. This contradicts our assumption that $x_j \ind x_k$.

    WLOG, suppose for contradiction that $x_i$ is an ancestor of $x_j$, but not an ancestor of $x_k$. As $x_k\nind x_i$, there must exist an active causal path between $x_k,x_i$. Suppose there exists a frontdoor path from $x_i$ to $x_k$: this contradicts our assumption that $x_i \not\in \text{Anc}(x_k)$. Suppose there exists a frontdoor path from $x_k$ to $x_i$: this contradicts our assumption that $x_j \ind x_k$. Suppose there exists a backdoor path between $x_i$ and $x_k$: this contradicts our assumption that $x_j \ind x_k$. Therefore, $x_i$ cannot be ancestor of $x_j$. WLOG, this implies that $x_i$ also cannot be ancestor of $x_k$.
\end{proof}
    
\end{comment}
\subsection{Proof of Lemma \ref{lemma: test stat}}\label{appendix: test stat}
\teststat*
\begin{proof}
    Note, $U'$ contains only ND and VLC vertices.

    Suppose test stastistic $t^*(e_i,S)$ asymptotically aproaches $0$ as $n\rightarrow \infty$. Suppose for contradiction that $e_i$ is not independent of $x_j \in \pi$. Note that this implies that the mutual information $\hat{MI}(x_j, e_i)\neq 0$, which contradicts our assumption that $t^*(e_i,S)\rightarrow 0$.

    Suppose $e_i$ is independent of all vertices in $S$. This implies that, for each $x_j \in \pi$ the mutual information approaches $0$: $\hat{MI}(x_j, e_i)\rightarrow 0$. This implies that the sum also approaches $0$: $t^*(e_i,S) = \sum_{x_j \in \pi}\hat{MI}(x_j, e_i) \rightarrow 0$.

\end{proof}

\section{PROPOSITION PROOFS}\label{appendix: propositions}
\subsection{Proof of Proposition \ref{prop: rootidcorr}}\label{appendix: rootidcorr}
\rootidcorr*
\begin{proof}
    
By Definition \ref{def: SRD}, Definition \ref{def: MRD}, Definition \ref{def: VP}, and Lemma \ref{lemma: MRDVP}, Lemma \ref{lemma: rootnoSRD}, we have Stage 1 of Algorithm \ref{algo: observed root id}, correctly identifies all VPs in graph $G$, eliminates all MRDs to obtain $V'$, then identifies all roots with no SRDs to obtain $V''$,  a set that contains only SRDs and roots.
%By Lemma \ref{lemma: rootnoSRD}, Stage 2 correctly identifies roots $\in V'$ with no SRDs, obtaining the set $V''$ that contains SRDs and their root ancestors.
%As Definition \ref{def: rootsuperset} explicitly defines the root superset $W\subseteq V''$ in terms of AD relations, which are in term defined explictly in terms of results from nonparametric regression in Definition \ref{def: ancestordescendent}, Stage 2 correctly identifies $W$ from $V''$.
By the Regression-Identification test (Definition \ref{def: ancestordescendent}), a superset of roots $W \subseteq V''$ is identified, and by
Lemma \ref{lemma: root id step} Stage 2 identifies all roots $\in V''$ by pruning nonroots from $W$.
Therefore, Root ID (Algorithm \ref{algo: observed root id}) correctly returns all roots in $G$.
\end{proof}

% \subsection{Proof of Proposition \ref{prop: rootidcondset}}\label{appendix: rootidcondset}
% \rootidcondset*
% \begin{proof}
% In Root ID, nonparametric regression is used in Stage 2 to identify the root superset $W$. These regressions are pairwise, requiring only univariate regressions. However, in Stage 2 of NHTS, regressions are run where $x_j$ is regressed on $x_i$ and all $x_k \in P_{ij} = \{x_k\in V: x_k\ind x_i, x_k\not\ind x_j\}$. We note that for $x_j$ that is an MRD, $P_{ij}$ is nonempty when $x_i$ is a root ancestor of $x_j$: in particular, $P_{ij}$ contains all other root ancestors of $x_j$. Therefore, NHTS requires multivariate regression whenever the graph contains MRDs; additionally, the size the largest multivariate regression is $P_{ij}$, which is bounded below by $\max_{x_i \in M}(\text{An}(x_i)\cap R, 0)$.
% \end{proof}

\subsection{Proof of Proposition \ref{prop: sortidcorr}}\label{appendix: sortidcorr}
\sortidcorr*
\begin{proof}
    Note that the roots of graph $G$ are provided as input to Sort ID. Therefore, $\pi$ is initialized with the roots.

We now induct on the length of $\pi$ to show that Sort ID recovers a correct topological sort of vertices in $G$.
\paragraph{Base Iterations ($1,2$)}
\begin{enumerate}
    \item Suppose there are $m$ roots identified by Root ID. Then, $m$ out of $d$ vertices have been correctly sorted into $\pi$.
    \item Note that in Stage 1, Sort Finder uses Lemma to prune the unsorted vertices $U$ to obtain $U'$, which contains only NDs and VLCs. Then, by Lemma \ref{lemma: test stat}, a VLC $x_*$ is selected using the test statistic $t^*$. Therefore, when $x_*$ is added to $\pi$, $\pi$ is still a correct topological sort.
\end{enumerate}
\paragraph{Iteration $k-1$, Inductive Assumption} We have correctly sorted $k-1$ vertices of $G$ into $\pi$.
\begin{enumerate}
\item Note that in Stage 1, Sort Finder uses Lemma to prune the unsorted vertices $U$ to obtain $U'$, which contains only NDs and VLCs. Then, by Lemma \ref{lemma: test stat}, a VLC $x_*$ is selected using the test statistic $t^*$. Therefore, when $x_*$ is added to $\pi$, $\pi$ is still a correct topological sort.
\end{enumerate}
Iteration $k-1$ Inductive Assumption is satisfied for iteration $k$, therefore we recover a valid topological sort $\pi$ from $|\pi| = m$ to $k$. Thus, for a DAG with $d$ vertices, Sort ID correctly recovers the full topological sort when provided the roots.
\end{proof}

\section{THEOREM PROOFS}\label{appendix: theorems}

\subsection{Proof of Theorem \ref{prop: rootidcondset}}\label{appendix: rootidcondset}
\rootidcondset*
\begin{proof}
Suppose the ANM is linear. Then, in Stage 2 of Root ID, after running pairwise univariate regression, roots are found to be in $An()$ relation to all vertices they are dependent on, and are immediately identified. No bivariate regressions take place. For DirectLiNGAM, only univariate regressions are used at any stage in the method, so the maximum conditioning set size is $1$.

In Root ID, nonparametric regression is used in Stage 2 to identify the root superset $W$. These regressions are either pairwise or bivariate; thus, the upper bound on the covariate set size is $2$.

Suppose the ANM is nonlinear. Then in Stage 2 of Root ID, bivariate regressions may be used, rendering the max covariate set size $2$. In Stage 2 of the root-finding procedure of NHTS, regressions are run where $x_j$ is regressed on $x_i$ and all $x_k \in P_{ij} = \{x_k\in V: x_k\ind x_i, x_k\not\ind x_j\}$. We note that for $x_j$ that is an MRD, $P_{ij}$ is nonempty when $x_i$ is a root ancestor of $x_j$: in particular, $P_{ij}$ contains all other root ancestors of $x_j$. Therefore, NHTS requires multivariate regression whenever the graph contains MRDs; additionally, the size the largest multivariate regression is $P_{ij}$, which is bounded below by $\max_{x_i \in M}(\text{An}(x_i)\cap R, 0)$.

The results for RESIT and NoGAM follow from Theorem \ref{theorem: losam_time_red}.

\end{proof}

\subsection{Proof of Theorem \ref{theorem: LoSAM correctness}}\label{appendix: LoSAM correctness}
\rootidalgo*
\begin{proof}
The correctness of Algorithm \ref{algo: observed root id} follows from Proposition \ref{prop: rootidcorr}, so the correct set of roots is returned. Then, it follows from Proposition \ref{prop: sortidcorr} that when Algorithm \ref{algo: observed sort id} is given the roots it correctly returns a valid topological ordering $\pi$.
\end{proof}

\subsection{Proof of Theorem \ref{theorem: LoSAM runtime}}\label{appendix: LoSAM runtime}
\rootidruntime*
\begin{proof}
We first find the runtime of Algorithm \ref{algo: observed root id}, then find the runtime of Algorithm \ref{algo: observed sort id}.
\\
\textbf{Part 1: Root Finder}
\\
 In Stage 1, Root ID first runs at most $O(d^2)$ marginal independence tests that each have $O(n^2)$ complexity. Then, to identify all VPs and prune MRDs, every triplet of vertices is checked, which has worst case $O(d^3)$ complexity. In Stage 2, at most, $O(d^2)$ pairwise nonparametric regressions are run, each with $O(dn\log(n))$ complexity. Therefore, Root ID has $O(d^3n\log(n) +d^2n^2)$ time complexity.
\\
\textbf{Part 2: Sort Finder}
\\
    In the worst case of a fully connected graph, there are $O(d)$ iterations of the Sort ID algorithm. In each iteration LoSAM runs at most $O(d^2)$ nonparametric mutual information tests, each of which has $O(n^2)$ time complexity. Therefore, Sort ID has worst case $O(d^3n^2)$ time complexity.
\\
\textbf{Overall Time Complexity}
\\
LoSAM (Algorithm \ref{algo: LoSAM}) has an overall time complexity of $O(d^3n^2)$, due mainly to the time complexity of Sort ID.
\end{proof}

\subsection{Proof of Theorem \ref{theorem: losam_time_red}}\label{appendix: LoSAM time red}
\losam*
\begin{proof}
    Results regarding NHTS, RESIT and NoGAM follow from Theorem 4.7 in \cite{suj_2024}. In the case of a fully directed graph, the Root Finder stage in LoSAM runs no multivariate regressions, as only the true root vertex is identified as a potential root. In the Sort Finder stage, LoSAM regresses each unsorted vertex onto all sorted vertices, finding vertices with independent residuals. Therefore, the number of regressions run is equal to $d$ minus the size of the covruate set. THerefore, when the covariate set is $k > \frac{d}{2}$, there are $d-k$ regressions run.
\end{proof}

\section{EXPERIMENTS}\label{appendix: exp}
\subsection{SYNTHETIC DATA GENERATION}\label{appendix: data generation}

% Similar to recent related literature 
% \citep{phillip_neural_cyclic, prosemary_learning_structure}, the nonlinear functions are parameterized by a neural network with random weights. See supplemental code for more details.

Details on the synthetic DGP:
\begin{itemize}
\item We first generate a $d$-dimensional DAG according to a Erdos-Renyi model, with the average number of edges is either $d$ or $2d$.
\item Each variable $x_i$ in the DAG is generated according to an ANM, i.e. $x_i = f_i(Pa(x_i)) + \varepsilon_i$. Each mechanism $f_i$ is picked independently. Function $f_i$ is linear with probability $p$ and nonlinear with probability $1-p$ ($p = 0, 0.25, 0.5, 0.75, 1$). 
\item If $f_i$ is linear, then the coefficients of each variable in $Pa(x_i)$ are drawn from $\text{Uniform}[-1.5,-0.5]$ with probability $1/2$ or drawn from $\text{Uniform}[0.5,1.5]$ with probability $1/2$. 
\item If $f_i$ is nonlinear, we follow recent related literature \citep{phillip_neural_cyclic, prosemary_learning_structure} and parameterize $f_i$ as a single-hidden-layer feedforward neural network with $\tanh$ activation. Specifically, for a variable $x_i$ with $d$ parents, the neural network takes the parent data $x_{\mathrm{Pa}(i)} \in \mathbb{R}^d$ as input, applies a linear transformation with weights sampled from $\text{Uniform}[-5, 5]$ followed by a $\tanh$ nonlinearity, and outputs a scalar value using a second linear transformation with weights also sampled from $\text{Uniform}[-5, 5]$. The number of hidden units is fixed to 10. The neural network weights are initialized independently for each variable.
\item Each $\varepsilon_i$ is independently drawn from either a Uniform, Laplacian, or Gaussian distribution with mean zero and variance $1/12$.
\item We then process the data to ensure it is challenging enough (see Appendix G.2 for more details). We first standardize the data to mean $0$ and variance $1$. We then check to see if the $R^2$-sortability is not too high - if the $R^2$-sortability is above $0.75$, we resample the data.
\end{itemize}

\subsection{DATA PROCESSING}\label{appendix: data processing}
\paragraph{Synthetic Data}
Recent work have pointed out that simulated ANM often have statistical artifacts missing from real-world data \citep{reisach_beware_2021, reisach_scale-invariant_2023}, leaving their real-world applicability open to question; for example, \cite{reisach_beware_2021} develop Var-Sort, which sorts variables according to increasing variance and is performant on synthetic data that is not standardized. Additionally, \cite{reisach_scale-invariant_2023} develop a scale-invariant heuristic method $R^2$-Sort by sorting variables according to increasing coefficient of determination $R^2$; they find that when synthetic data is naively generated it tends to have high $R^2$-sortability, while the prevalence of high $R^2$-sortability in real data is unknown.

To alleviate concern that our data is gameable by Var-Sort, we standardize all data to zero mean and unit variance. However, processing data to be challenging for $R^2$-sort is more difficult: 
\cite{reisach_scale-invariant_2023} explain how the $R^2$ coefficients depend on the graph structure, noise variances, and edge weights of the underlying DAG in a complex manner, concluding that "one cannot isolate the effect of individual parameter choices on $R^2$-sortability, nor easily obtain the expected $R^2$-sortability when sampling ANMs given some parameter distributions, because the $R^2$ values are determined by a complex interplay between the different parameters." Therefore, we are unable to directly generate data with low $R^2$; instead, to generate each instance of data we randomly sample DAGs up to 100 times until a DAG with $R^2$-sortability less than $0.75$ is generated. We were able to achieve lowered $R^2$-sortability in all experiments except linear ANMs on dense graphs. These processing steps prevent the heuristic algorithms Var-Sort \citep{reisach_beware_2021} and $R^2$-Sort \citep{reisach_scale-invariant_2023} from leveraging arbitrary features of simulated data to accurately recover a topological sort, and ensure that the data is sufficiently challenging for our discovery algorithm and the baselines.

\paragraph{Sachs Data}
To generate the results in Table \ref{fig: sachs_data}, we ran the methods on 30 different random samples from the Sachs Dataset of size $n=300$.

\subsection{ATOP METRIC}\label{appendix: atop def}
\cite{montagna_scalable_2023} introduce topological order divergence as a measure of the discrepancy between the estimated
topological ordering $\pi$ and the adjacency matrix of the true causal graph $A$, expressed as:

\begin{equation}
    D_{top}(\pi, A) = \sum_{i=1}^{d} \sum_{j:\pi_i > \pi_j} A_{i,j}
    \label{eq:order_divergence}.
\end{equation}

They note that "if $\pi$ is a correct topological order for $A$, then $D_{top}(\pi, A) = 0$. Otherwise, $D_{top}(\pi, A)$ counts the number of edges that cannot be recovered due to the choice of topological order."

\cite{suj_2024} introduce a normalized version of $D_{top}$, $A_{top}$, expressed as:
\begin{equation}
    A_{top} = \frac{D_{top}}{|A|},
\end{equation}
where $|A|$ is the number of edges in graph $A$. If $\pi$ is a correct topological order for $A$, then $A_{top}(\pi, A) = 1$. Otherwise, $A_{top}(\pi, A)$ equals the percentage of edges that cannot be recovered due to the choice of topological order.

\subsection{DETAILS}\label{appendix: exp details}
% We provide additional details on our experimental approach. Note that, to enable a fair comparison to NHTS, similar to \cite{suj_2024} we use the version of NHTS they provide that returns a linear topological sort, rather than a hierarchical topological sort, by adding only one vertex to the sort in each iteration of its sorting procedure. See supplement file for more details.

%Cutoff values were set to $\alpha=0.01$ for independence tests and $\alpha = 0.05$ for residual independence tests for NHTS, while 

% For the regression step in LoSAM, we use a RandomForestRegressor with the following settings: n\_estimators = $100$, max\_depth = $10$, min\_sample\_split = $10$, min\_sample\_leaf = $5$, max\_features = "sqrt".
% All cutoffs for LoSAM were set at $0.01$; for all other possible hyperparameters in other baseline methods, default settings were used. 
Implementation details of LoSAM and all considered baselines.
  \begin{itemize}
    \item For regression tasks, LoSAM uses a RandomForestRegressor from the \texttt{sklearn.ensemble} package with the following settings: n\_estimators = 100, max\_depth = 10, min\_sample\_split = 10, min\_sample\_leaf = 5, max\_features = "sqrt." LoSAM uses the kernel independence test (KCI) developed in \citep{zhang_kernel-based_2011}, with the estimator KCI from the \texttt{causal-learn} package for independence tests, and the mutual information test developed in \citep{kraskov_estimating_2004}, with the estimator from the \texttt{npeet} package (both with default settings). The cutoff for all independence tests is 0.01.
    \item All baselines were used with default hyperparameters encoded by the packages they were imported from. CaPS and NHTS were imported from the github repositories associated with those papers (https://github.com/E2real/CaPS/tree/main, https://github.com/Sujai1/hybrid-discovery). DirectLiNGAM and RESIT were imported from the \texttt{lingam} package. SCORE, NoGAM, and CAM were imported from the \texttt{DoDiscover} package. VarSort and Randsort were implemented by hand. The $R^2$-sortability metric was imported from the \texttt{CausalDisco} package. Please see the supplemental code to find the exact code used to run each baseline.
    \end{itemize}

All experiments were conducted in python; experiments involving CaPS were conducted on an Apple M2 Pro Chip, 16 Gb
of RAM, with no parallelization, while experiments with all other methods were conducted on a r6a.metal EC2 instance with 192vCPUs, 1536 GiB memory.

\newpage
\subsection{ADDITIONAL EXPERIMENTAL RESULTS}\label{appendix: add_exps}
\subsubsection{DENSE GRAPHS}\label{appendix: dense graphs}
Experimental results for dense graphs (ER2).

\begin{figure*}[h!]%{0.8\textwidth}
\centering
     \begin{subfigure}[t]{0.8\textwidth}  % 4th row, 1st column
        \includegraphics[width=\textwidth]{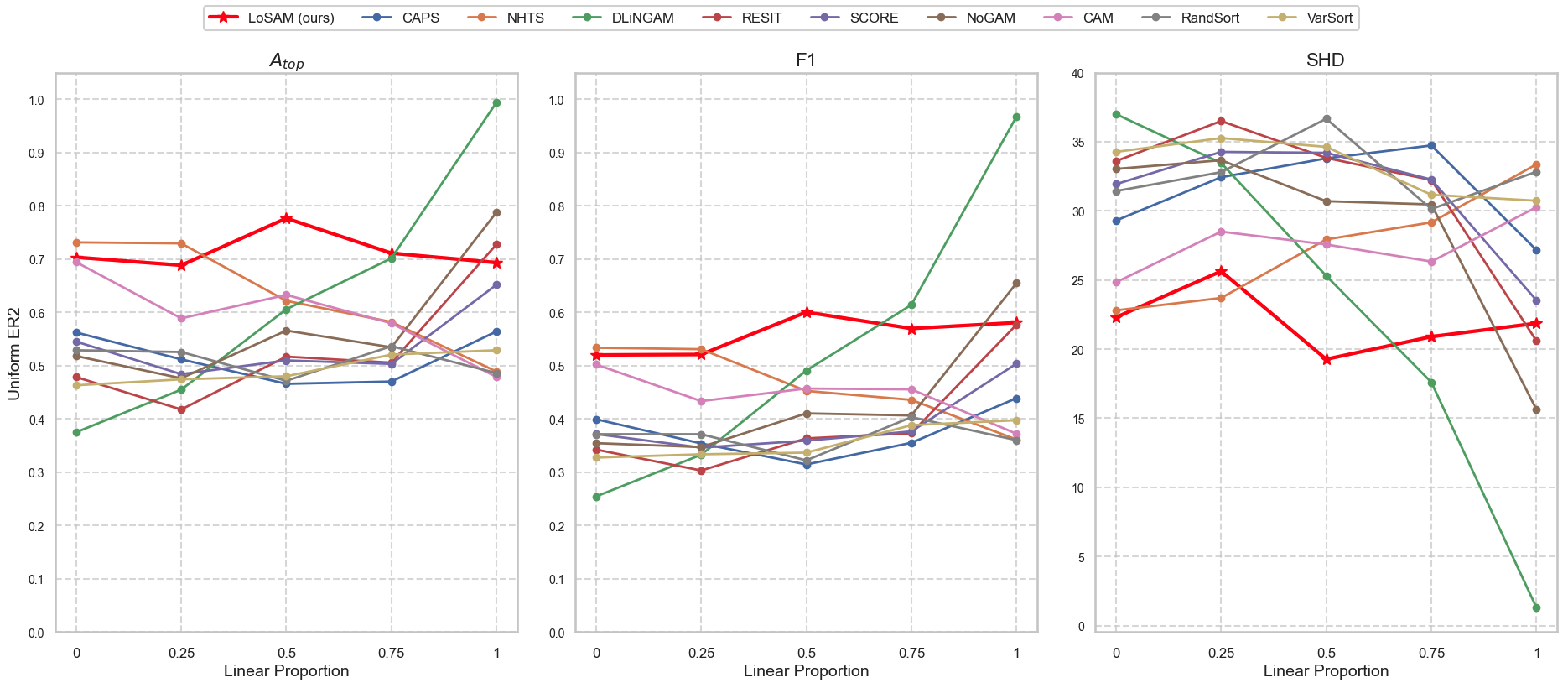}
    \end{subfigure}
    
     \begin{subfigure}[t]{0.8\textwidth}  % 4th row, 1st column
        \includegraphics[width=\textwidth]{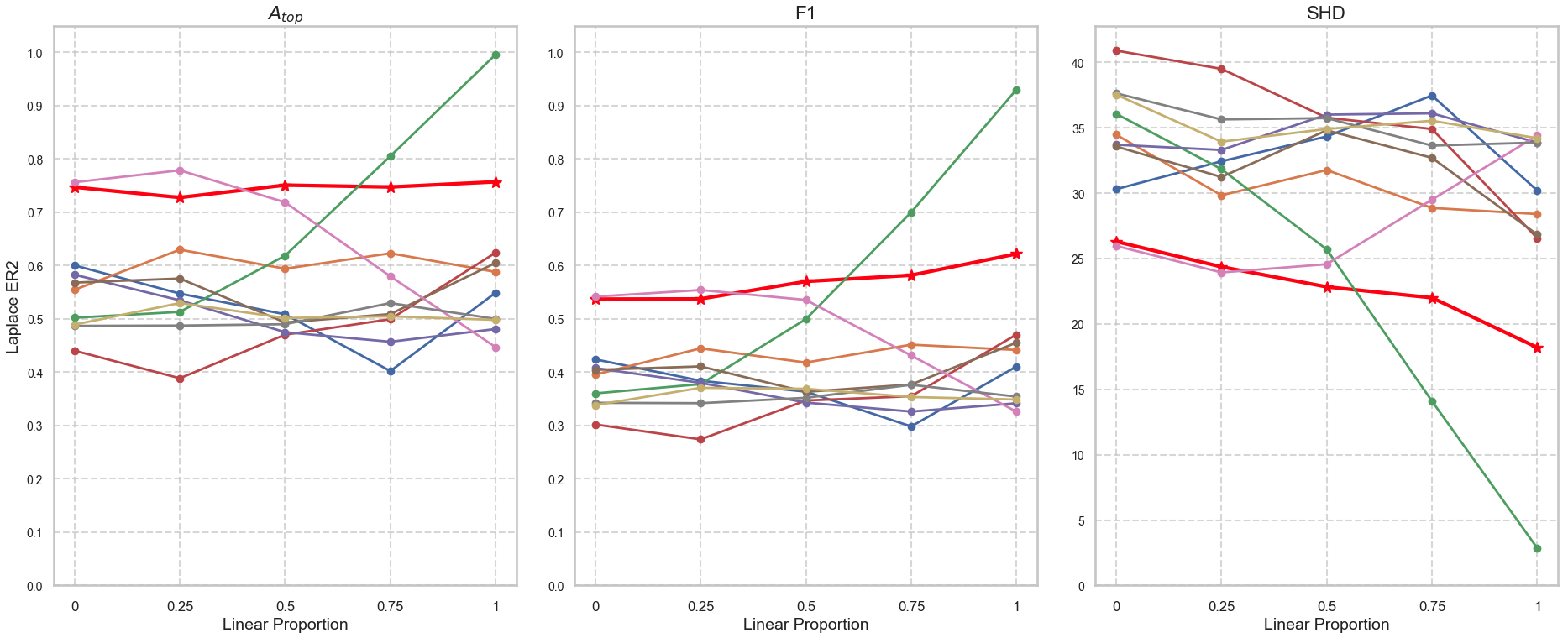}
    \end{subfigure}
    \caption{Performance of LoSAM across diff. proportion of linear mechanism on dense ER2 data; uniform (top) and laplacian noise (bottom). } 
    \label{fig: dense results breakdown}
\end{figure*}

\begin{figure*}[h!]%{0.8\textwidth}
\centering
     \begin{subfigure}[t]{0.8\textwidth}  % 4th row, 1st column
        \includegraphics[width=\textwidth]{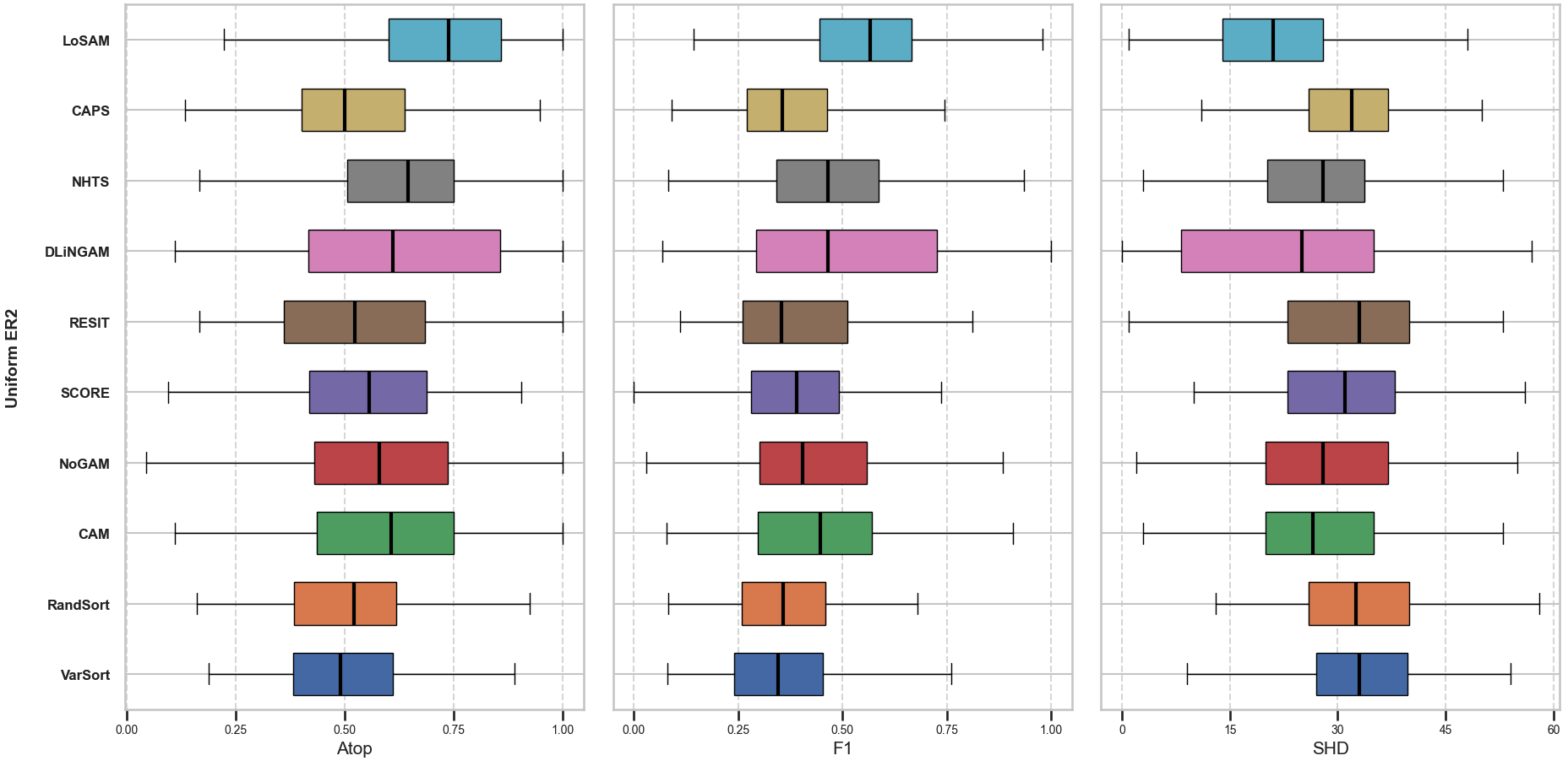}
    \end{subfigure}
    
     \begin{subfigure}[t]{0.8\textwidth}  % 4th row, 1st column
        \includegraphics[width=\textwidth]{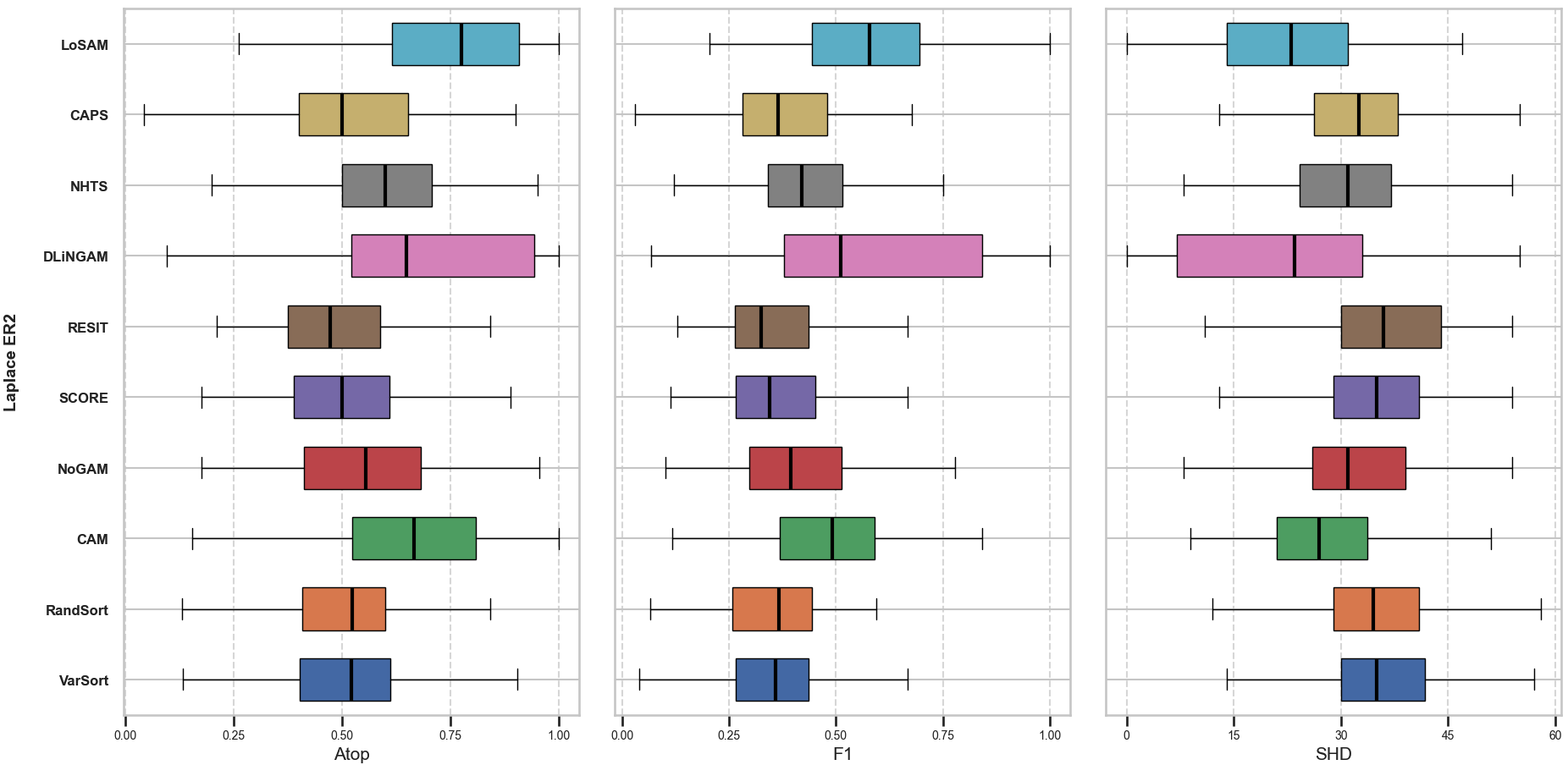}
    \end{subfigure}
    \caption{ Performance of LoSAM on dense ER2 synthetic data with $50\%$ proportion of linear mechanisms. Top row: uniform noise. Bottom row: Laplace noise.
    }
    \label{fig: dense results overall}
\end{figure*}

\newpage
\subsubsection{HIGH-DIMENSIONAL GRAPHS}\label{appendix: high dim graphs}
% Experimental results for high-dimensional graphs. Note: due to extremely slow runtime and poor performance from other methods, we elect to compare LoSAM against baselines with fast runtime (Var-Sort, Rand-Sort) in higher dimensions.

\begin{figure*}[h!]%{0.8\textwidth}
\centering
     \begin{subfigure}[t]{.8\textwidth}  % 4th row, 1st column
        \includegraphics[width=\textwidth]{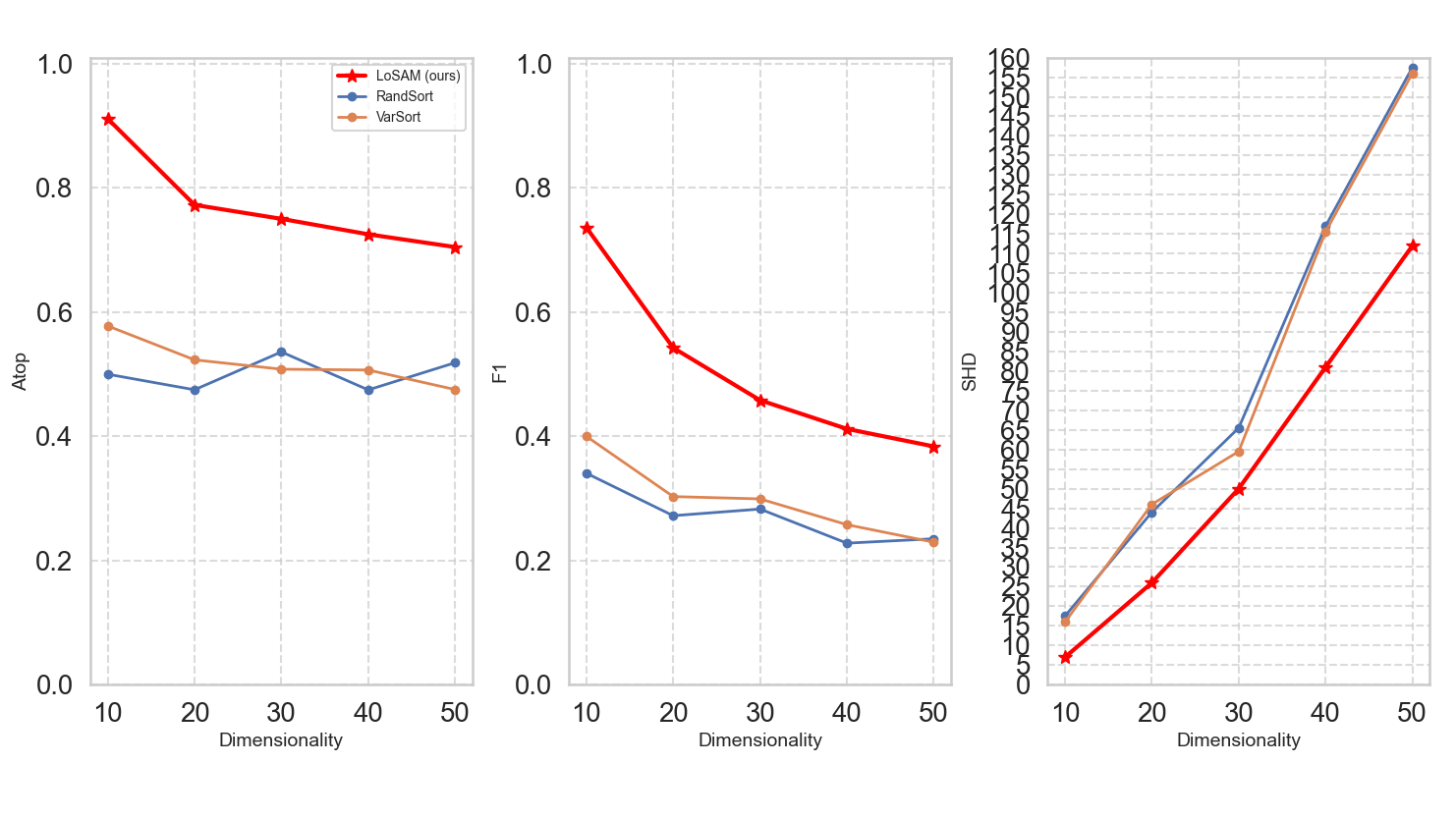}
    \end{subfigure}
    \caption{Performance of LoSAM on ER1 graph with uniform noise and linear proportion = $0.5$, across diff. dimensionalities.} 
    \label{fig: high dim results}
\end{figure*}
We evaluate how LoSAM's performance metrics evolve in high-dimensional settings (we omit other baselines due to computational cost). Results show LoSAM remains effective as the complexity of the data increases, though performance declines as $d$ increases with fixed $n$ -- an expected consequence of fixed $n$ in high dimensions. This degradation occurs because the same $n=2000$ samples become increasingly sparse as $d$ grows, which will reduce estimation accuracy for any method.

\newpage
\subsubsection{NONLINEAR GAUSSIAN}\label{appendix: nonlin gaussian graphs}

\begin{figure*}[h!]%{0.8\textwidth}
\centering
     \begin{subfigure}[t]{0.8\textwidth}  % 4th row, 1st column
        \includegraphics[width=\textwidth]{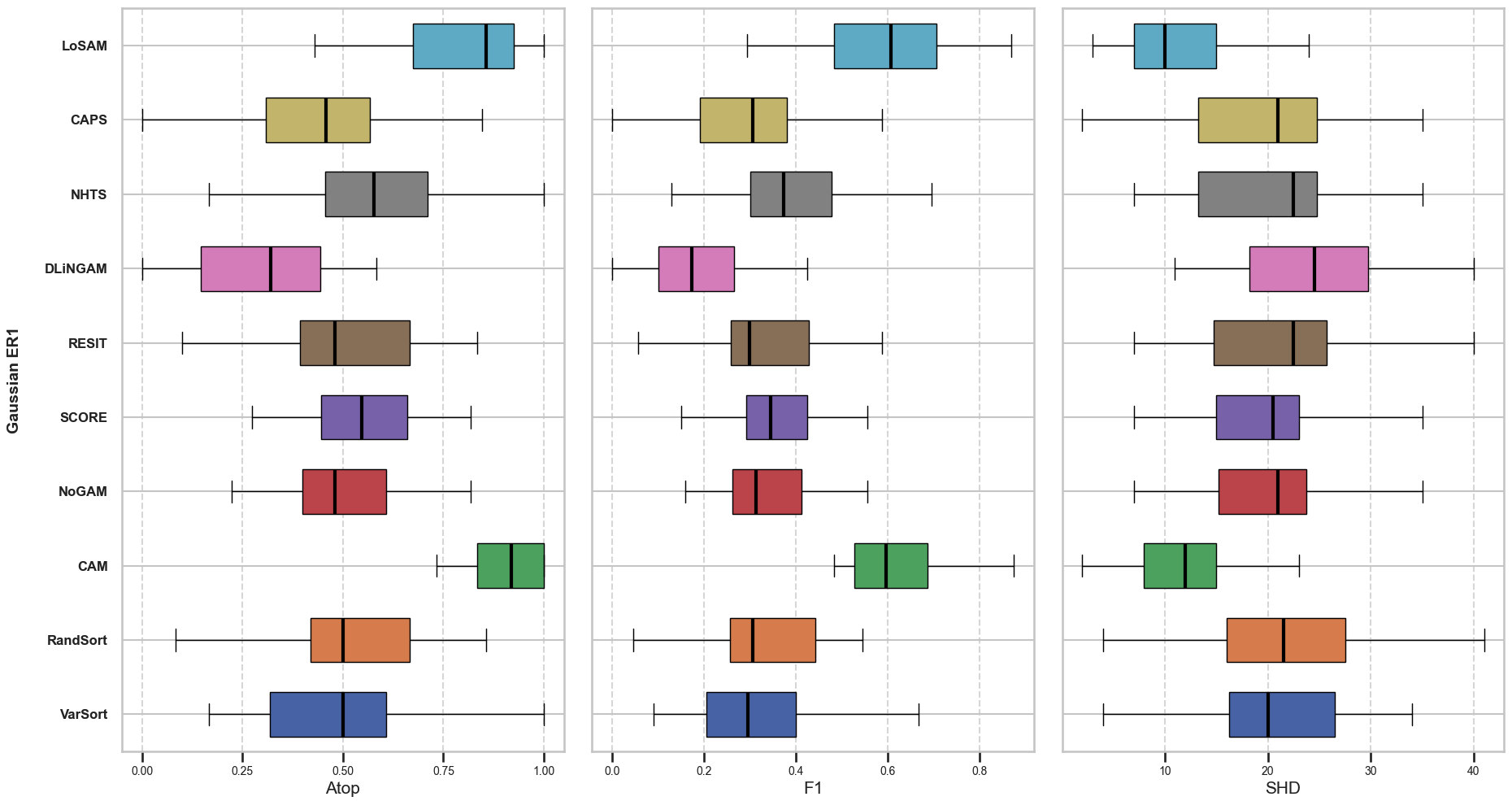}
    \end{subfigure}
    \caption{Performance of  LoSAM across all DGMs; uniform noise (left) and laplacian noise (right). %this is unclear} 
    }
    \label{fig: nonlinear gauss results}
\end{figure*}

Experimental results for nonlinear gaussian ANMs.
\newpage
\subsubsection{SENSITIVITY TO ESTIMATION ERROR}\label{appendix: estim error sensit}
\begin{figure*}[h!]%{0.8\textwidth}
\centering
     \begin{subfigure}[t]{0.8\textwidth}  % 4th row, 1st column
        \includegraphics[width=\textwidth]{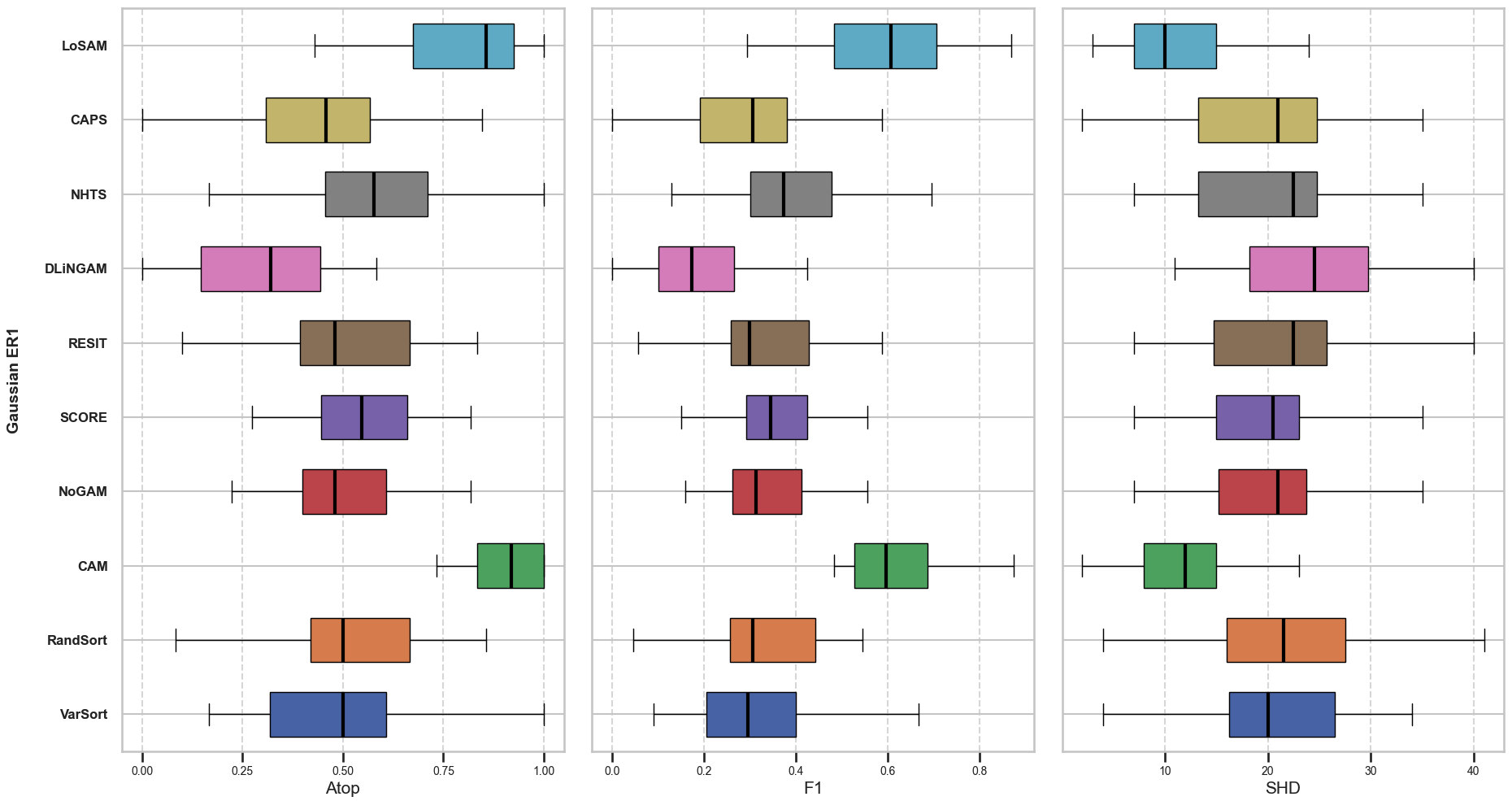}
    \end{subfigure}
    \caption{Performance of LoSAM on ER1 graph with uniform noise, $d=10,n=300$ across all DGMs, with different regression estimators. %this is unclear} 
    }
    \label{fig: esim error sensit}
\end{figure*}

We evaluate LoSAM variants with different regression estimators (all taken from the \texttt{Sklearn} package):
\begin{enumerate}
    \item LoSAM: RandomForestRegressor.
\item LoSAM\_KRR\_Poly4: Kernel Ridge Regression (polynomial kernel, degree 4).
\item LoSAM\_KRR\_Poly8: Kernel Ridge Regression (polynomial kernel, degree 8).
\item LoSAM\_KRR\_RBF: Kernel Ridge Regression (RBF kernel).
\end{enumerate}

Hyperparameters for each estimator:
\begin{enumerate}
    \item 
LoSAM: n\_estimators = 100, max\_depth = 10, min\_sample\_split = 10, min\_sample\_leaf = 5, max\_features = "sqrt".
\item LoSAM\_KRR\_Poly4: kernel = "poly", degree = 4, alpha =0.1, coef0=1.
\item LoSAM\_KRR\_Poly8: kernel = "poly", degree = 8, alpha =0.1, coef0=1.
\item LoSAM\_KRR\_RBF: kernel = "rbf", alpha =0.1, gamma = 0.01.
\end{enumerate}
% GES \citep{chickering_learning_nodate}, and GRaSP \citep{grasp_liam_2021} as baseline comparators that are agnostic to the noise distribution. Note that the scoring-based approaches (GES, GRaSP) return a MEC, rather than a unique DAG; similar to prior work \citep{montagna_assumption_2023}, we randomly select one topological ordering permitted by the returned MEC for evaluation, to enable fair comparison.

\newpage
\subsubsection{COMAPARISON TO OPTIMIZATION-BASED ANM METHOD}\label{appendix: notearsish compare}
\begin{figure*}[h!]%{0.8\textwidth}
\centering
     \begin{subfigure}[t]{0.8\textwidth}  % 4th row, 1st column
        \includegraphics[width=\textwidth]{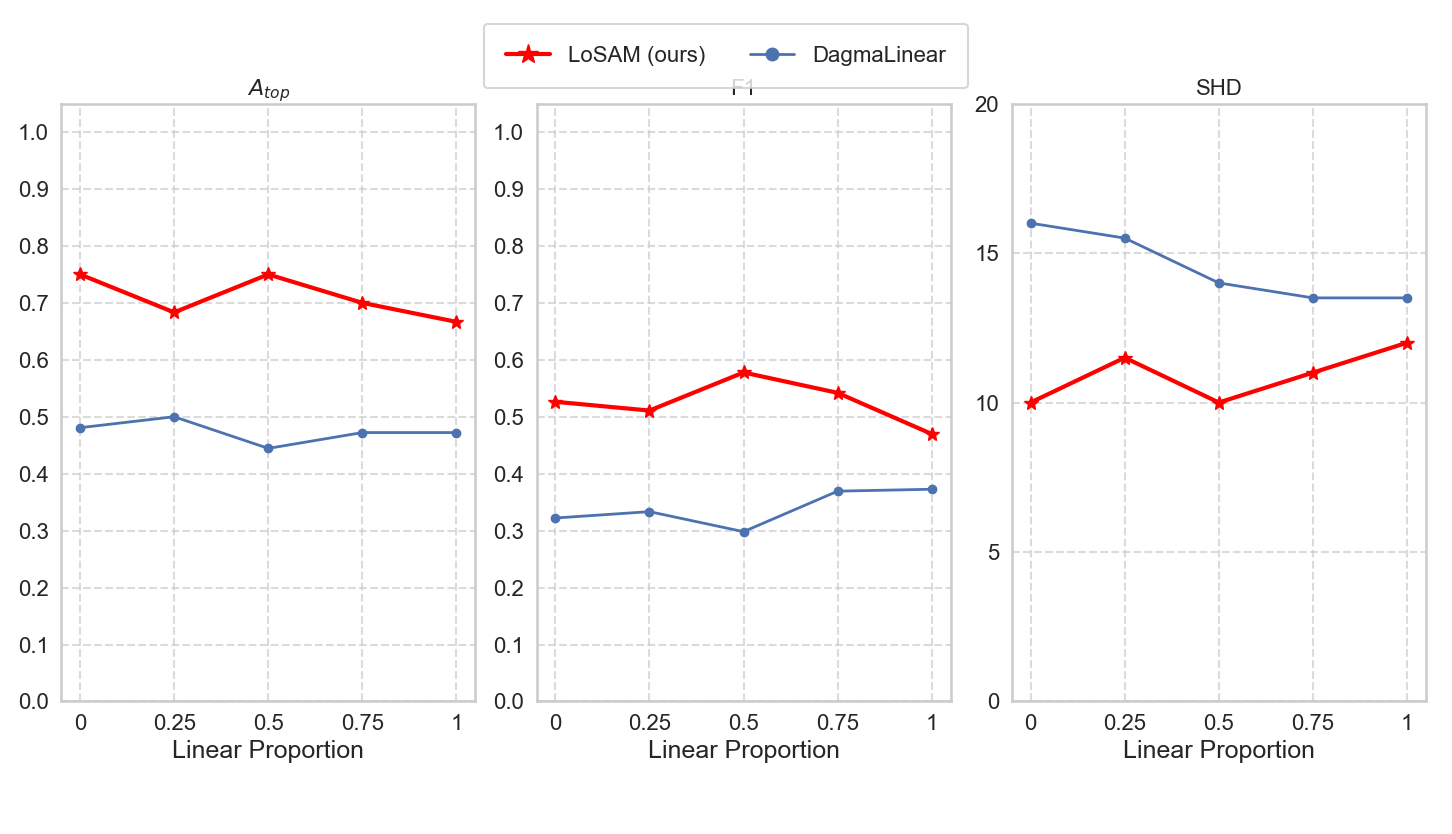}
    \end{subfigure}
    \caption{Performance of LoSAM and DagmaLinear on ER1 graph with uniform noise, $d=10,n=300$ across all DGMs. %this is unclear} 
    }
    \label{fig: notearish compare}
\end{figure*}
We evaluate LoSAM against a linear optimization-based method (implementation from \texttt{DAGMA} package, hyperparameters taken from Appendix C.1.1. of \citealt{bello2022dagma}). Results show that LoSAM's performance is superior to this optimization-based ANM method.
\newpage
\subsection{RUNTIME RESULTS}\label{appendix: runtime}

\subsubsection{SPARSE}
The runtime results for Figure \ref{fig: sparse results overall}

\begin{table}[h!]
    \centering
    \renewcommand{\arraystretch}{1.2}
    \setlength{\tabcolsep}{6pt}
    \begin{tabular}{lccccc}
        \toprule
         & \multicolumn{5}{c}{\textbf{Linear Proportion}} \\
        \cmidrule(lr){2-6}
        \textbf{Methods} & \textbf{0} & \textbf{0.25} & \textbf{0.5} & \textbf{0.75} & \textbf{1} \\
        \midrule
        LoSAM    & $8.495 \pm 1.998$   & $8.853 \pm 2.022$   & $9.606 \pm 2.034$   & $9.818 \pm 2.214$   & $8.634 \pm 1.529$   \\
        CAPS     & $8.753 \pm 0.228$   & $8.589 \pm 0.295$   & $8.849 \pm 0.281$   & $8.877 \pm 0.486$   & $9.273 \pm 0.414$   \\
        NHTS     & $38.962 \pm 12.904$ & $42.804 \pm 15.097$ & $45.760 \pm 19.363$ & $57.838 \pm 12.157$ & $39.070 \pm 11.562$ \\
        DLiNGAM  & $3.301 \pm 0.983$   & $3.251 \pm 1.095$   & $3.737 \pm 1.160$   & $4.004 \pm 1.592$   & $4.152 \pm 1.432$   \\
        RESIT    & $36.666 \pm 1.848$  & $35.099 \pm 1.676$  & $37.113 \pm 2.322$  & $35.920 \pm 2.623$  & $33.576 \pm 2.350$  \\
        SCORE    & $9.717 \pm 8.122$   & $7.766 \pm 26.275$  & $11.759 \pm 16.300$ & $13.039 \pm 27.730$ & $17.037 \pm 33.070$ \\
        NoGAM    & $14.205 \pm 12.660$ & $14.176 \pm 9.061$  & $21.131 \pm 14.918$ & $34.412 \pm 35.567$ & $42.827 \pm 49.857$ \\
        CAM      & $37.760 \pm 5.892$  & $39.825 \pm 5.133$  & $39.192 \pm 6.005$  & $43.425 \pm 6.491$  & $38.608 \pm 9.230$  \\
        RandSort & $2.503 \pm 0.812$   & $2.556 \pm 1.103$   & $2.499 \pm 0.896$   & $3.150 \pm 1.360$   & $3.102 \pm 1.373$   \\
        VarSort  & $2.923 \pm 0.806$   & $2.528 \pm 0.961$   & $2.254 \pm 0.695$   & $2.455 \pm 1.077$   & $2.841 \pm 1.116$   \\
        \bottomrule
    \end{tabular}
    \caption{Runtime results for ER1 graphs and Uniform noise.}
    \label{tab:runtime uniform}
\end{table}

\begin{table}[h!]
    \centering
    \renewcommand{\arraystretch}{1.2}
    \setlength{\tabcolsep}{6pt}
    \begin{tabular}{lccccc}
        \toprule
         & \multicolumn{5}{c}{\textbf{Linear Proportion}} \\
        \cmidrule(lr){2-6}
        \textbf{Methods} & \textbf{0} & \textbf{0.25} & \textbf{0.5} & \textbf{0.75} & \textbf{1} \\
        \midrule
        LoSAM    & $9.049 \pm 1.802$   & $9.564 \pm 2.945$   & $9.963 \pm 1.965$   & $9.708 \pm 2.601$   & $10.650 \pm 2.108$ \\
        CAPS     & $8.851 \pm 0.275$   & $8.915 \pm 0.534$   & $9.434 \pm 0.363$   & $9.522 \pm 0.682$   & $9.290 \pm 0.408$  \\
        NHTS     & $31.039 \pm 6.179$  & $29.860 \pm 9.266$  & $39.188 \pm 12.381$ & $38.780 \pm 13.329$ & $44.217 \pm 16.933$ \\
        DLiNGAM  & $2.910 \pm 0.880$   & $3.263 \pm 1.177$   & $3.320 \pm 1.106$   & $4.450 \pm 1.212$   & $4.673 \pm 1.320$  \\
        RESIT    & $36.100 \pm 1.195$  & $36.212 \pm 2.022$  & $34.985 \pm 1.477$  & $36.036 \pm 2.705$  & $36.697 \pm 2.106$ \\
        SCORE    & $9.117 \pm 25.930$  & $10.183 \pm 19.615$ & $8.483 \pm 30.603$  & $12.368 \pm 18.493$ & $34.701 \pm 24.086$ \\
        NoGAM    & $16.238 \pm 24.006$ & $9.448 \pm 14.069$  & $12.992 \pm 27.836$ & $13.347 \pm 13.847$ & $26.202 \pm 19.485$ \\
        CAM      & $41.502 \pm 13.920$ & $40.390 \pm 9.508$  & $38.531 \pm 11.565$ & $37.154 \pm 4.340$  & $39.573 \pm 13.534$ \\
        RandSort & $2.632 \pm 2.123$   & $2.535 \pm 1.557$   & $2.306 \pm 1.526$   & $2.649 \pm 0.796$   & $2.757 \pm 1.124$  \\
        VarSort  & $2.777 \pm 1.965$   & $2.859 \pm 1.317$   & $2.225 \pm 0.643$   & $2.593 \pm 0.689$   & $2.532 \pm 1.167$  \\
        \bottomrule
    \end{tabular}
    \caption{Runtime results for ER1 graphs and Laplace noise.}
    \label{tab:runtime laplace}
\end{table}

\newpage
\subsubsection{DENSE}
The runtime results for Figure \ref{fig: dense results breakdown}.

\begin{table}[h!]
    \centering
    \renewcommand{\arraystretch}{1.2}
    \setlength{\tabcolsep}{6pt}
    \begin{tabular}{lccccc}
        \toprule
         & \multicolumn{5}{c}{\textbf{Linear Proportion}} \\
        \cmidrule(lr){2-6}
        \textbf{Methods} & \textbf{0} & \textbf{0.25} & \textbf{0.5} & \textbf{0.75} & \textbf{1} \\
        \midrule
        LoSAM    & $12.203 \pm 2.274$  & $13.083 \pm 8.218$  & $12.951 \pm 5.856$  & $14.761 \pm 4.445$  & $14.646 \pm 6.186$  \\
        CAPS     & $8.345 \pm 0.195$   & $8.551 \pm 0.256$   & $8.714 \pm 0.244$   & $8.177 \pm 0.145$   & $8.196 \pm 0.260$   \\
        NHTS     & $68.842 \pm 10.248$ & $72.542 \pm 8.703$  & $88.527 \pm 10.292$ & $84.519 \pm 10.007$ & $67.167 \pm 6.453$  \\
        DLiNGAM  & $4.461 \pm 1.220$   & $4.080 \pm 1.903$   & $5.185 \pm 1.172$   & $5.599 \pm 1.518$   & $4.602 \pm 1.371$   \\
        RESIT    & $35.534 \pm 3.293$  & $35.021 \pm 7.270$  & $36.113 \pm 4.412$  & $36.214 \pm 5.124$  & $35.061 \pm 5.673$  \\
        SCORE    & $103.090 \pm 66.855$& $138.517 \pm 45.652$& $113.100 \pm 64.391$& $115.136 \pm 61.357$& $151.977 \pm 15.391$\\
        NoGAM    & $83.755 \pm 25.638$ & $93.860 \pm 18.106$ & $57.771 \pm 24.034$ & $60.016 \pm 35.062$ & $108.342 \pm 14.460$\\
        CAM      & $38.067 \pm 37.486$ & $36.290 \pm 26.036$ & $36.541 \pm 32.798$ & $37.179 \pm 39.518$ & $40.331 \pm 6.242$  \\
        RandSort & $3.012 \pm 1.442$   & $2.831 \pm 0.785$   & $3.410 \pm 3.181$   & $2.618 \pm 3.355$   & $3.579 \pm 0.945$   \\
        VarSort  & $2.524 \pm 1.168$   & $2.644 \pm 0.904$   & $2.996 \pm 3.431$   & $2.495 \pm 2.434$   & $3.163 \pm 1.066$   \\
        \bottomrule
    \end{tabular}
    \caption{Results for ER1 graphs and Uniform noise.}
    \label{tab:linear_proportiondense}
\end{table}

\begin{table}[h!]
    \centering
    \renewcommand{\arraystretch}{1.2}
    \setlength{\tabcolsep}{6pt}
    \begin{tabular}{lccccc}
        \toprule
         & \multicolumn{5}{c}{\textbf{Linear Proportion}} \\
        \cmidrule(lr){2-6}
        \textbf{Methods} & \textbf{0} & \textbf{0.25} & \textbf{0.5} & \textbf{0.75} & \textbf{1} \\
        \midrule
        LoSAM    & $19.513 \pm 10.874$ & $15.102 \pm 7.256$  & $17.479 \pm 11.748$ & $15.625 \pm 10.218$ & $17.261 \pm 3.968$  \\
        CAPS     & $8.554 \pm 0.212$   & $8.281 \pm 0.263$   & $8.137 \pm 0.254$   & $8.392 \pm 0.224$   & $8.296 \pm 0.187$   \\
        NHTS     & $41.759 \pm 4.190$  & $54.444 \pm 11.174$ & $67.633 \pm 14.289$ & $75.535 \pm 11.840$ & $94.058 \pm 12.138$ \\
        DLiNGAM  & $4.238 \pm 1.499$   & $4.866 \pm 1.486$   & $5.100 \pm 1.208$   & $4.828 \pm 1.667$   & $5.094 \pm 1.194$   \\
        RESIT    & $36.790 \pm 2.213$  & $36.638 \pm 6.152$  & $39.732 \pm 2.838$  & $36.152 \pm 4.840$  & $39.040 \pm 4.453$  \\
        SCORE    & $136.591 \pm 66.444$& $142.725 \pm 63.160$& $23.886 \pm 42.393$ & $41.719 \pm 56.895$ & $82.003 \pm 67.079$ \\
        NoGAM    & $85.983 \pm 30.039$ & $70.836 \pm 28.574$ & $31.275 \pm 17.726$ & $62.495 \pm 32.673$ & $68.000 \pm 45.448$ \\
        CAM      & $34.128 \pm 38.388$ & $37.716 \pm 24.377$ & $50.529 \pm 30.624$ & $48.106 \pm 32.871$ & $35.543 \pm 22.036$ \\
        RandSort & $3.241 \pm 2.999$   & $3.687 \pm 1.932$   & $3.160 \pm 3.058$   & $2.868 \pm 3.216$   & $3.082 \pm 1.783$   \\
        VarSort  & $2.849 \pm 2.216$   & $3.393 \pm 1.501$   & $2.393 \pm 3.510$   & $2.910 \pm 1.249$   & $2.836 \pm 1.105$   \\
        \bottomrule
    \end{tabular}
    \caption{Runtime results for ER1 graphs and Laplace noise.}
    \label{tab:linear_proportion}
\end{table}

\newpage
\subsubsection{HIGH DIMENSIONAL}
The runtime results for Figure \ref{fig: high dim results}.

\begin{table}[h!]
    \centering
    \renewcommand{\arraystretch}{1.2}
    \setlength{\tabcolsep}{6pt}
    \begin{tabular}{lcccc}
        \toprule
         & \multicolumn{4}{c}{\textbf{\(d\) Value}} \\
        \cmidrule(lr){2-5}
        \textbf{Methods} & \textbf{10} & \textbf{15} & \textbf{20} & \textbf{25} \\
        \midrule
        LoSAM    & \(9.606 \pm 2.034\)   & \(34.217 \pm 12.832\)  & \(70.256 \pm 28.749\)  & \(94.054 \pm 25.865\) \\
        DLiNGAM  & \(3.737 \pm 1.160\)   & \(19.336 \pm 4.231\)   & \(48.557 \pm 24.781\)  & \(55.616 \pm 7.221\)  \\
        RandSort & \(2.499 \pm 0.896\)   & \(18.547 \pm 2.984\)   & \(39.321 \pm 11.026\)  & \(53.576 \pm 3.136\)  \\
        VarSort  & \(2.254 \pm 0.695\)   & \(17.907 \pm 4.758\)   & \(32.088 \pm 12.803\)  & \(52.063 \pm 7.968\)  \\
        \bottomrule
    \end{tabular}
    \caption{Runtime results for different \(d\) values.}
    \label{tab:d_values}
\end{table}

\subsubsection{NONLINEAR GAUSSIAN}
The runtime results for Figure \ref{fig: nonlinear gauss results}.

\begin{table}[h!]
    \centering
    \renewcommand{\arraystretch}{1.2}
    \setlength{\tabcolsep}{6pt}
    \begin{tabular}{l c}
        \toprule
        \textbf{Methods} & \textbf{0} \\
        \midrule
        LoSAM    & $8.228 \pm 2.200$ \\
        CAPS     & $9.149 \pm 0.337$ \\
        NHTS     & $44.242 \pm 13.680$ \\
        DLiNGAM  & $3.526 \pm 0.923$ \\
        RESIT    & $35.377 \pm 1.334$ \\
        SCORE    & $16.902 \pm 30.607$ \\
        NoGAM    & $12.432 \pm 12.605$ \\
        CAM      & $39.349 \pm 10.694$ \\
        RandSort & $3.330 \pm 2.406$ \\
        VarSort  & $3.085 \pm 2.322$ \\
        \bottomrule
    \end{tabular}
    \caption{Results for ER1 graphs and Gaussian noise with linear proportion $0$.}
    \label{tab:results_0}
\end{table}

\newpage~
\section{ALGORITHM WALKTHROUGH}
In this section we walk-through LoSAM on two different examplary DAGs.
\subsection{LoSAM}\label{appendix: LoSAM walkthrough}

\begin{comment}
\begin{figure}[h!]
    \centering
        \includegraphics[width=0.4\textwidth]{figures/LoSAM.png}
  
    \caption{LoSAM walkthrough on example DAG.}\label{fig: losam walkthrough}
\end{figure}
\end{comment}
\textbf{Example 1:}
\begin{figure}[h!]
    \centering
    \begin{subfigure}[t]{0.45\textwidth}  % 1st row, 1st column
        \includegraphics[width=\textwidth]{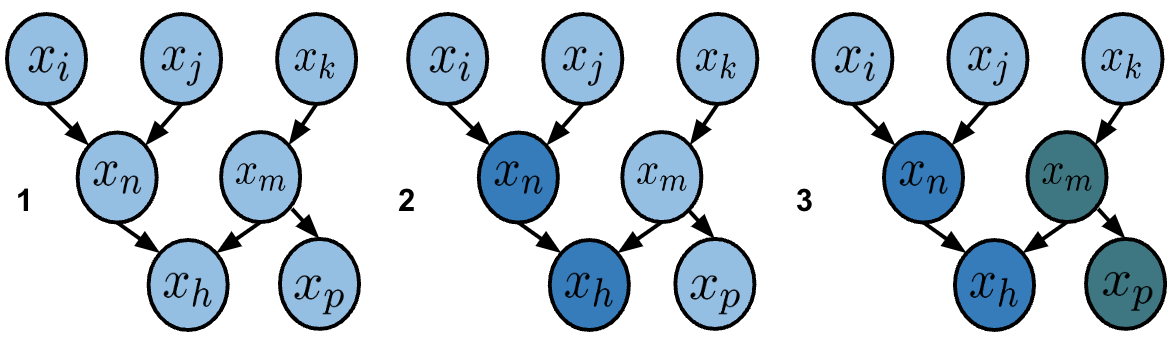}
    \end{subfigure}
    \begin{subfigure}[t]{0.45\textwidth}  % 1st row, 2nd column
        \includegraphics[width=\textwidth]{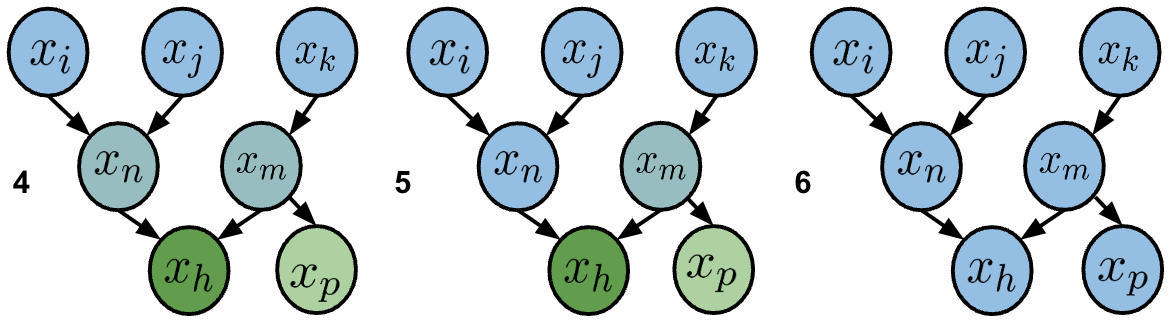}
    \end{subfigure}
    \caption{LoSAM walkthrough on example DAG.}\label{fig: losam walkthrough}
\end{figure}

Subfigure 1 of Figure \ref{fig: losam walkthrough} illustrates the example causal graph from which data is generated. Now, we walk through how LoSAM would obtain a topological sort of this DAG. In Stage 1 Root Finder, $x_n, x_h$ are identified as MRDs and are pruned (subfigure 2). Then, $x_i,x_j$ are recovered as roots, as they are independent of $x_k, x_m, x_p$. Then, in Stage 2, $x_m,x_p$ are pruned as they are not identified as ancestors of any vertices, while $x_k$ is (subfigure 3). We therefore recover $x_i,x_j,x_k$ as roots. Subfigure 4 illustrates the decomposition of the graph into roots ($x_i,x_j,x_k$), VLCs ($x_n,x_m$), a ND $x_h$ and an LD ($x_p$). In Stage 1 of Sort Finder, $x_p$ is pruned from $U = \{x_n,x_m,x_p,x_h\}$ as it as an LD. Then, $x_n$ is determined as a VLC, as a minimizer of the test statistic $t^*$ (subfigure 5), and is sorted. This is again repeated 3 more times to sort all of $U$ (subfigure 6); therefore, LoSAM correctly obtains a valid sort $\pi$.

\textbf{Example 2:}\\
Consider a DAG $G$ with five vertices $x_1,x_2,x_3,x_4,x_5$, where $x_1\rightarrow x_3, x_2\rightarrow x_3, x_1\rightarrow x_4, x_4\rightarrow x_5$. Suppose the functional relationships are given by $x_1=\varepsilon_1,x_2=\varepsilon_2,x_3=f_3(x_1,x_2)+\varepsilon_3,x_4=f_4(x_1)+ \varepsilon_4, x_5=f_5(x_4)+ \varepsilon_5$, where all $f_i$ are nonlinear.

In stage 1 of the root-finding subroutine of LoSAM (Algorithm 1), we run pairwise independence tests between all vertices. We find that $x_3$ induces a VP between $x_1,x_2$; this means that $x_3$ is an MRD and so we prune it. We then note that $x_2$ is independent of all non-pruned vertices, and so we classify it as a root. In stage 2, we run pairwise nonparametric regression between the remaining vertices $x_1,x_4,x_5$. We note that regressing $x_4$ onto $x_1$ yields an independent residual, and regressing $x_5$ onto $x_4$ yields an independent residual. Therefore, by Definition 3.4, $x_1$ is identified $\in An(x_4)$, and $x_4$ is identified $\in An(x_5)$. As $x_5$ is not identified as the ancestor of any variable, it cannot be a root and is pruned. Note that, although $x_4$ is identified $\in An(x_5)$, it is identified as a descendant of $x_1$ -- therefore, $x_4$ cannot be a root and is pruned. We are left with $x_1$ being the other root in this DAG.

% \begin{comment}
% \subsection{LoSAM-UC}\label{appendix: LoSAM-UC walkthrough}
\begin{comment}
\begin{figure}[h!]
    \centering
        \includegraphics[width=0.4\textwidth]{figures/LoSAM-UC1.png}
  
    \caption{LoSAM-UC Walkthrough}\label{fig: losam-uc walkthrough}
\end{figure}
\end{comment}
\begin{comment}
    
\begin{figure}[h!]
    \centering
    \begin{subfigure}[t]{0.45\textwidth}  % 1st row, 1st column
        \includegraphics[width=\textwidth]{figures/losamuc1.png}
    \end{subfigure}
    \begin{subfigure}[t]{0.45\textwidth}  % 1st row, 2nd column
        \includegraphics[width=\textwidth]{figures/losamuc2.png}
    \end{subfigure}
    \caption{LoSAM walkthrough on example DAG.}\label{fig: losam-uc walkthrough}
\end{figure}

Subfigure 1 of Figure \ref{fig: losam-uc walkthrough} illustrates the example causal graph from which data is generated, where $x_u$ is unobserved. In Stage 1 of LoSAM-UV, $x_j$ is identified as a MRD-I and is pruned (subfigure 2). Then, $x_m$ is found as a proxy of $x_k,x_h$, and therefore we identify that $x_k\ind x_h$ given some unobserved confounder, and therefore $x_i$ induces a VP between $x_k,x_h$ as a MRD-N. After pruning $x_i,x_j$, we identify $x_n$ as a root, independent of $x_m,x_j,x_h$. In Stage 2, we confirm that $x_m,x_k,x_h$ are roots (subfigure 4) by recovering independent residuals from every bivariate regression. Using the discovered roots, in Stage 3 Sort Finder recovers a valid topological sort $\pi$.
\end{comment}
\section{CLARIFYING EXPLANATIONS}\label{appendix: DETAILS}

\subsection{NHTS FAILURE MODE}\label{appendix: nonlinear anc-des pair}

% This example demonstrates that the residual independence relations differ between the same ancestor-descendant pairs when the underlying functions are linear or nonlinear, and thus NHTS fails to accurately recover the correct topological sort.

% Consider a DAG G with three vertices $x_1, x_2, x_3$, where $x_1 \rightarrow x_2, x_2 \rightarrow x_3$. Suppose NHTS has correctly identified the root vertex $x_1$.

% Suppose the functional causal relationships are nonlinear, given by $x_1 = \varepsilon_1, x_2 = x_1^2 + \varepsilon_2, x_3 = x_1x_2+ \varepsilon_3$. Then, when NHTS tries to determine the next variable in the sort, it will regress both $x_3$ and $x_2$ onto $x_1$, and find that $x_1$ is independent of the residual only in one regression; thus, NHTS can distinguish between the two and will accurately sort $x_2$ before $x_3$.

% Suppose the functional causal relationships are linear, given by $x_1 = \varepsilon_1, x_2 = x_1 + \varepsilon_2, x_3 = x_2 + \varepsilon_3$, where the $\varepsilon_i$s are mutually independent noise variables. Then, when NHTS tries to determine the next variable in the sort, it will regress both $x_3$ and $x_2$ onto $x_1$, and find that $x_1$ is independent of the residual in both regressions. Thus, NHTS cannot distinguish between the two, and will fail to accurately sort them. Residual independence from naively running regressions is insufficient to handle both linear and nonlinear mechanisms.

LoSAM shares some similiarities with NHTS, as they both utilize a local search approach and regression-based tests in their topological ordering procedure. However, LoSAM differs fundamentally from NHTS on a conceptual basis.

NHTS defines a set of local substructures (PP1, PP2, PP3, PP4 relations) that characterize the space of possible parent-child relationships. In particular it finds that, under nonlinear relationships, only the roots ( + a small class of nonroots) satisfy PP2 relations and can be identified through regression.%It then leverages conditional independence tests to prune the remaining non-roots. 
 To find the rest of the ordering, NHTS exploits the fact that, under nonlinear relationships, only vertices in the next unknown topological layer (which are only a children of the sorted vertices) will yield an independent residual after regression onto the sorted vertices.

In contrast, LoSAM defines a set of local causal substructures (SRDs, MRDs, VP, VLC, ND, LD) that are defined in terms of ancestor-descendant relationships. This difference results in LoSAM being able to handle mixed mechanism ANM with fewer high dimensional regressions.

\textbf{Identifiability Issues}
NHTS's reliance on local structures that are defined in terms of parent-child relations limits the method in terms of identifiability: as shown by \citealt{shimizu2011directlingam}, linear mechanisms can mean that even ancestor-descendant relationships can yield independent residuals. We provide Example 1 below to demonstrate that the residual independence relations differ between the same ancestor-descendant pairs when the underlying functions are linear or nonlinear, and thus NHTS fails to accurately recover the correct topological sort. In contrast, it follows from Theorem 4.7 that LoSAM accurately recovers the topological sort in Example 1.

\textbf{Sample Efficiency Issues}
NHTS's reliance on the PP2 framework limits the method in terms of sample efficiency - by leveraging parent-child relations, this requires that all parents are included as covariates in a regression to recover an independent residual. This can lead to high-dimensional regressions in the root finding stage. We provide Example 2 below to demonstrate that, even when the underlying mechanism is nonlinear, the PP2-framework approach of NHTS requires high-dimensional regressions to recover root vertics. In contrast, it follows from Theorem 3.6 that LoSAM recovers the roots with no high-dimensional regressions.

\textbf{Example 1}
Consider a DAG G with three vertices $x_1, x_2, x_3$, where $x_1 \rightarrow x_2, x_2 \rightarrow x_3$. Suppose NHTS has correctly identified the root vertex $x_1$.

Suppose the functional causal relationships are nonlinear, given by $x_1 = \varepsilon_1, x_2 = x_1^2 + \varepsilon_2, x_3 = x_1x_2+ \varepsilon_3$. Then, when NHTS tries to determine the next variable in the sort, it will regress both $x_3$ and $x_2$ onto $x_1$, and find that $x_1$ is independent of the residual only in one regression; thus, NHTS can distinguish between the two and will accurately sort $x_2$ before $x_3$.

Suppose the functional causal relationships are linear, given by $x_1 = \varepsilon_1, x_2 = x_1 + \varepsilon_2, x_3 = x_2 + \varepsilon_3$, where the $\varepsilon_i$s are mutually independent noise variables. Then, when NHTS tries to determine the next variable in the sort, it will regress both $x_3$ and $x_2$ onto $x_1$, and find that $x_1$ is independent of the residual in both regressions. Thus, NHTS cannot distinguish between the two, and will fail to accurately sort them. Residual independence from naively running regressions is insufficient to handle both linear and nonlinear mechanisms.

\textbf{Example 2} Consider a DAG $G$ with 100 vertices $x_1, x_2, \ldots, x_{99}, x_{100}$, where $x_1, \ldots, x_{99}$ are all roots, and are all parents of $x_{100}$ ($x_1 \rightarrow x_{100}, x_2 \rightarrow x_{100},\ldots, x_{99}\rightarrow x_{100}$).

Suppose the functional causal relationships are nonlinear, given by $x_1 = \varepsilon_1, x_2 = \varepsilon_2, \ldots, x_{99}= \varepsilon_{99},  x_{100} = \text{tanh}(x_1\times x_2\ldots\times x_{99})+ \varepsilon_{100}$. Then, when NHTS tries to determine that $x_{100}$ is in PP2 relation any of other vertices, to obtain an independent residual it will need to regress $x_{100}$ onto \textit{all} $99$ other vertices! Therefore, the PP2 framework approach is incredibly sample inefficient, tending towards using high-dimensional regressions as the number of roots grows large. 

% Then, $x_1$ is independent of the residuals produced by regressing either $x_2$ or $x_3$ onto $x_1$, allowing for its identification as a root. However, suppose the functional causal relationships are nonlinear, given by $x_1 = \varepsilon_1, x_2 = \varepsilon_2, x_3 = x_1x_2 + \varepsilon_3$. Now, $x_1$ is independent of the residual produced by the regression of $x_2$ onto $x_1$, but not independent of the residual produced by the regression of $x_3$ onto $x_1$ - this prevents the identification of $x_1$ as a root."

\subsection{COMPARISON OF V-PATTERNS AND V-STRUCTURES}\label{appendix: vpattern compar}
We note that v-structures do not always correspond to the statistical constraints required to be a VP, in the sense that $x_i,x_j$ may be conditionally independent, rather than marginally independent as required in our definition of VP. For example, in a DAG where $x_i \rightarrow x_j, x_i\rightarrow x_k, x_j\rightarrow x_p, x_k\rightarrow x_p, x_p \rightarrow x_m$, $x_j,x_k,x_p$ form a v-structure but not a VP. However, v-structures are also not always more general than VPs, as in a DAG where $x_j\rightarrow x_p, x_k\rightarrow x_p, x_p \rightarrow x_m$, $x_j,x_k,x_m$ form a VP but not a v-structure.
Therefore, it is accurate to characterize the statistical constraints of v-structures and VPs as distinct but related concepts that overlap when either a v-structure has marginally independent parent vertices or when a triplet pair is a VP and has only parent-child relations.

\section{LIMITATIONS}\label{appendix: limit}
\raggedbottom
%\subsection{Real Data Issues}
 We note that the standard approach in the causal discovery literature to assess a method's finite sample performance is to evaluate the method on synthetically generated data \citep{reisach_beware_2021}. This is because causal ground truth is incredibly rare, as it requires real-world experiments \citep{causal_ground}; these experiments can expensive, potentially unethical, but most importantly are often infeasible or ill defined \citep{ambig_manip_spirtes_2004}. This lack of substantial real-world benchmark datasets is important because key assumptions that are used to generate synthetic data, which discovery methods rely on (such as additive noise, faithfulness, and causal sufficiency), may be violated in practice \citep{causal_ground}. Therefore we caution that experimental results on synthetic data should be interpreted as a demonstration of a method's theoretical performance on somewhat idealized data, which may not reflect measurements from real-world settings where assumptions critical to the method are not met.

\section{ASSET INFORMATION}\label{sec: asset info}

DirectLiNGAM and RESIT were imported from the \texttt{lingam} package. $R^2-$sort was imported from the \texttt{CausalDisco} package. NHTS and LoSAM were implemented using the kernel ridge regression function from the \texttt{Sklearn} package, used kernel-based independence tests from the \texttt{causal-learn} package, and a mutual information estimator from the \texttt{npeet} package.
 %implemented using the kernel ridge regression function from the \texttt{Sklearn} package and used kernel-based independence tests from the \texttt{causal-learn} package. LoSAM was implemented using the kernel ridge regression function from the \texttt{Sklearn} package, used kernel-based independence tests from the \texttt{causal-learn} package, and a mutual information estimator from the \texttt{npeet} package.
 All assets used have a CC-BY 4.0 license. %See file named `aistats2025\_topological\_sort\_experiments' for further details.

\end{document}